\documentclass[12pt,oneside]{book}
\usepackage[table,x11names,svgnames]{xcolor}
\usepackage[sectionbib]{natbib}

\usepackage[ED=MITT-IT, Ets=UT3]{tlsflyleaf}
\usepackage[english]{babel}
\usepackage[utf8]{inputenc}
\usepackage[T1]{fontenc}
\usepackage{stmaryrd}
\usepackage{amssymb}
\usepackage{amsmath}
\usepackage{libertine}
\usepackage{caption}
\usepackage{subcaption}
\usepackage{float}
\usepackage{multirow}
\usepackage{apalike}
\usepackage{minitoc}
\usepackage{tikz}
\usepackage{enumitem}
\usepackage{epigraph}
\usepackage{lscape}
\usepackage{url}  
\usepackage{booktabs}
\usepackage[nonumberlist]{glossaries}
\usepackage{calc}
\usepackage[]{titlesec} 
\definecolor{linkColor}{HTML}{32a852}
\usepackage[pagebackref,breaklinks,linktocpage,colorlinks=true,citecolor=linkColor,linkcolor=cyan]{hyperref}
\usepackage{wrapfig}

\renewcommand*\backref[1]{\ifx#1\relax \else (cit. on pp. #1) \fi}

\usepackage{fancyhdr}
\usepackage{lipsum}

\usepackage{svg}
\usepackage{subcaption}
\usepackage{float}
\usepackage{multirow}
\usepackage{times}
\usepackage{color}
\usepackage{lscape}
\usepackage{tabularx}
\usepackage{arydshln}
\usepackage{graphicx}
\usepackage{adjustbox}
\usepackage{mwe}
\usepackage{subfiles}

\usepackage[numbib,notlof,notlot,nottoc]{tocbibind}

\usepackage{xr}
\makeatletter
\newcommand*{\addFileDependency}[1]{
  \typeout{(#1)}
  \@addtofilelist{#1}
  \IfFileExists{#1}{}{\typeout{No file #1.}}
}
\makeatother

\usepackage{titletoc}   

\usepackage[capitalize,noabbrev]{cleveref}
\usepackage{algorithm}
\usepackage{algpseudocode}
\usepackage{wrapfig}

\usepackage{amsfonts}
\usepackage{bm}

\definecolor{yourcolor}{HTML}{8a0e19}

\titleformat{\chapter}[display]
{\normalfont\color{yourcolor}}
{\filleft\huge\color{black}\textsc\chaptertitlename\hspace*{2mm}%
	\begin{tikzpicture}[baseline={([yshift=-.6ex]current bounding box.center)}]
	\node[fill=yourcolor,circle,text=white] {\thechapter};
	\end{tikzpicture}
}
{1ex}
{\titlerule[1.5pt]\vspace*{1.5ex}\Huge\color{black}\textsc}
[]

\titleformat{name=\chapter,numberless}[display]
{\normalfont\color{black}}
{}
{1ex}
{\vspace*{-5cm}\Huge\textsc}
[]

\newcommand{\printmyminitoc}[1]{%
	\noindent\hspace{1cm}%
	\colorlet{chpnumbercolor}{black}%
	\begin{tikzpicture}
	\node(s){
		\begin{minipage}{.9\linewidth}
		\printcontents[chapters]{}{1}{}
		\end{minipage}
	};
	{
		\color{yourcolor}
		\draw(s.north west)--(s.north east) (s.south west)--(s.south east);
	}
	\end{tikzpicture}
	\vspace*{3ex}
	
	#1
	\vfill
	\pagebreak
}

\title{\textbf{\large Exploring the role of (self-)attention in cognitive and computer vision architecture}}
\author{Mohit VAISHNAV}
\defencedate{13/04/2023}
\lab{D\'epartement Sciences et technologies de l'information et de la communication}

\nboss{2}
\makesomeone{boss}{1}{Thomas SERRE}{}{} 
\makesomeone{boss}{2}{Nicholas ASHER}{}{}  
\nreferee{2}
\makesomeone{referee}{1}{Jonathan D. Cohen (Princeton University, USA)}{}{}
\makesomeone{referee}{2}{Hugues Talbot (CentraleSup\`elec, France)}{}{}
\njudge{6}
\makesomeone{judge}{1}{Timoth\`ee Masquelier}{CerCo, France}{Pr\'esident du Jury}
\makesomeone{judge}{2}{Jonathan D. Cohen}{Princeton University, USA}{Membre du Jury}
\makesomeone{judge}{3}{Hugues Talbot}{CentraleSup\`elec, France}{Membre du Jury}
\makesomeone{judge}{4}{Jessica B. Hamrick}{DeepMind, UK}{Membre du Jury}
\makesomeone{judge}{5}{Thomas Serre}{Brown University, USA}{Thesis Director}
\makesomeone{judge}{6}{Nicholas Asher}{ANITI, France}{Thesis Co-Director}

\begin{document}

    \makeflyleaf

\pagestyle{fancy}

\fancyhead{}

\renewcommand{\chaptermark}[1]{\markboth{\textsc{#1}}{}}

\setlength{\headheight}{14.49998pt}
\addtolength{\topmargin}{-2.49998pt}

\frontmatter

\dominitoc

\chapter*{R\'esum\'e}
\addcontentsline{toc}{chapter}{R\'esum\'e }

Un m\'ecanisme fondamental de la cognition, n\'ecessaire \`a l'ex\'ecution de tâches de raisonnement complexes, est la capacit\'e de traiter s\'electivement les informations (attention) et de les conserver dans un \'etat accessible (m\'emoire). Nous analysons syst\'ematiquement le rôle de ces deux composantes, en commençant par l'auto-attention bas\'ee sur le modèle d'attention le plus populaire: Transformer, et en \'etendant ensuite l'architecture \`a la m\'emoire. Transformer est aujourd'hui la dernière classe d'architecture neuronale et est au coeur des demonstrations les plus fascinante du Deep Learning, il a apport\'e un changement de paradigme dans le domaine de l'intelligence artificielle. Il a remplac\'e les r\'eseaux de r\'ecurrence et de convolution par l'auto-attention comme choix architectural de facto pour la plupart des applications de l'IA.

Nous \'etudions d'abord les m\'ecanismes de calcul impliqu\'es dans un test de raisonnement visuel synth\'etique (SVRT), en analysant la capacit\'e d'une architecture de vision par ordinateur populaire (ResNet) de diff\'erentes profondeurs et entraîn\'ee sur des ensembles de donn\'ees de diff\'erentes tailles. Cela a conduit \`a une nouvelle taxonomie plus fine pour les vingt-trois tâches de SVRT, coh\'erente avec les classes de tâches de raisonnement - identiques-diff\'erentes (SD) et de relations spatiales (SR) - largement accept\'ees dans la litt\'erature. Ensuite, nous \'etudions le rôle de l'auto-attention incorpor\'ee \`a ResNet50 dans la r\'esolution du d\'efi SVRT. Inspir\'es par les deux types de systèmes d'attention visuelle, nous avons mod\'elis\'e l'auto-attention pour qu'elle soit utilis\'ee comme une attention bas\'ee sur les caract\'eristiques et sur une attention spatiale pour enrichir les cartes de caract\'eristiques d'un r\'eseau feedforward. Nous avons \'evalu\'e la capacit\'e de ces r\'eseaux d'attention \`a r\'esoudre le d\'efi SVRT et avons constat\'e que les architectures r\'esultantes \'etaient beaucoup plus efficaces pour r\'esoudre la plus difficile de ces tâches de raisonnement visuel. La nouvelle taxonomie obtenue pr\'ec\'edemment s'explique aussi partiellement par l'am\'elioration relative des deux r\'eseaux d'attention et conduit \`a des pr\'edictions testables concernant les besoins attentionnels des tâches SVRT. 

Enfin, nous d\'eveloppons une nouvelle architecture cognitive int\'egrant l'auto-attention et la m\'emoire. Nous proposons GAMR: \textbf{G}uided \textbf{A}ttention \textbf{M}odel for visual \textbf{R}easoning, motiv\'e par la th\'eorie de la vision active. Le GAMR a des m\'ecanismes de fonctionnement similaires \`a ceux du cerveau qui r\'esout des tâches complexes de raisonnement visuel par des s\'equences de changements d'attention pour s\'electionner et acheminer en m\'emoire les informations visuelles pertinentes pour la tâche. Ce changement d'attention est mis en œuvre \`a l'aide d'un module d'auto-attention guid\'e par une requête g\'en\'er\'ee en interne. Nous d\'emontrons que \textit{GAMR} est efficace, robuste et compositionnel par rapport \`a l'une ou l'autre des architectures bas\'ees sur le feedforward, l'attention ou la m\'emoire. De plus, GAMR est capable de g\'en\'eraliser \`a des tâches de raisonnement complètement nouvelles. Dans l'ensemble, notre travail analyse le rôle de l'auto-attention dans l'architecture cognitive et de vision par ordinateur par leur capacit\'e \'a r\'esoudre des tâches complexes de raisonnement visuel n\'ecessitant de l'attention comme \'el\'ement cl\'e pour r\'esoudre efficacement les tâches de raisonnement.

\chapter*{Abstract}
\addcontentsline{toc}{chapter}{Abstract}

A fundamental mechanism of cognition needed to perform complex reasoning tasks is the ability to selectively process information (attention) and retain information in an accessible state (memory). We systematically analyze the role of both these components, starting with Transformer-based self-attention as a model of attention and later extending the architecture with memory. The Transformer is the latest and seemingly most powerful class of neural architecture, and it has brought a paradigm shift in the field of artificial intelligence. It has replaced recurrence and convolution networks with self-attention as the de-facto architectural choice for most AI applications. 

We first study the computational mechanisms involved in a synthetic visual reasoning test (SVRT) challenge, analyzing the ability of popular computer vision architecture (ResNet) of different depths trained on different dataset sizes. It led to a novel, finer taxonomy for the twenty-three SVRT tasks consistent with the broadly accepted same-different (SD) and spatial-relation (SR) classes of reasoning tasks in literature. Next, we study the role of self-attention incorporated with ResNet50 in solving the SVRT challenge. Inspired by the two types of visual attention systems, we modeled self-attention to be used as feature-based and spatial attention to enrich the feature maps of a feedforward network. We evaluated the ability of these attention networks to solve the SVRT challenge and found the resulting architectures to be much more efficient at solving the hardest of these visual reasoning tasks. The novel taxonomy obtained earlier is also partially explained by the relative improvement of the two attention networks and leads to testable predictions regarding the attentional needs of SVRT tasks. 

At last, we develop a novel cognitive architecture integrating attention and memory. We propose GAMR: \textbf{G}uided \textbf{A}ttention \textbf{M}odel for (visual) \textbf{R}easoning, motivated by the theory of active vision. GAMR has similar working mechanisms as that of the brain that solves complex visual reasoning tasks via sequences of attention shifts to select and route the task-relevant visual information into memory. This shift of attention is implemented with the help of a attention module guided by an internally generated query. We demonstrate that \textit{GAMR} is sample-efficient, robust, and compositional compared to either of the feedforward, attention or memory-based architectures. In addition, GAMR is shown to be capable of zero-shot generalization on completely novel reasoning tasks. Overall, our work analyzes the role of self-attention in cognitive and computer vision architecture by their ability to solve complex visual reasoning tasks needing  attention as a key component to efficiently solve reasoning tasks.

\clearpage
\begin{center}
    \thispagestyle{empty}
    \vspace*{\fill}
    To the former president, missile man of India, nuclear scientist, writer, poet, and educator \\
    Dr. A. P. J. Abdul Kalam
    \vspace*{\fill}
\end{center}
\clearpage

\sloppy

\chapter*{Acknowledgments}
\addcontentsline{toc}{chapter}{Acknowledgments}

Sailing through the past three years has been an unforgettable experience filled with countless challenges. I would like to use this opportunity to show how grateful I am to all the people who have helped me throughout this exciting journey toward fulfilling my Ph.D.

First and foremost, I thank my academic advisor, Thomas Serre (Brown University, USA) and Nicholas Asher (ANITI, France), for accepting me at ANITI. Words cannot express my gratitude to them for their invaluable guidance and patience and for providing me with the intellectual freedom to work. I am particularly thankful to Thomas Serre for his unwavering support, assistance in bridging neuroscience and AI, and encouragement of my ideas. All the conversations we had helped me to develop scientific thinking and to comprehend the ability to filter out the most exciting approaches to be followed. Our conversations have shaped my scientific thinking and helped me filter out the most exciting approaches. His one-to-one review meetings and critical judgment have expanded my boundaries and enlightened me with countless ideas to progress. 

I acknowledge the members of my thesis committee, Timoth\`ee Masquelier (Senior Research Scientist, CerCo, France), Jonathan D. Cohen (Princeton Neuroscience Institute, Princeton University, USA), Hugues Talbot (CentraleSup\`elec, France) and Jessica Hamrick (Senior Research Scientist, DeepMind, UK). Their expertise and insights have enriched my research and contributed to its quality.

Additionally, this endeavor would not have been possible without the Agence Nationale de la Recherche (ANR) support for generously financing my research. I sincerely thank Corinne Joffre, Secr\`etaire g\`en\`erale, ANITI, and her colleagues for supporting my multiple mobilizations between the USA and France.

This journey would have been unfinished without the help of the computing staff at High-Performance Cluster (HPC) \textit{Oscar}, Brown University, USA and \textit{CALMIP}, Universit\`e F\`ed\`erale de Toulouse Midi-Pyr\`en\`ees (UFTMiP), France. They provided their expertise to help me handle computationally intensive jobs.

Special thanks go to Rufin VanRullen, Research Director at CerCo, for his reliable and practical scientific mentoring and for offering me office space alongside his team. I am grateful to my office mates Andrea Alamia and Aimen Zerroug, who became my friends and collaborators. Aimen has been my travel companion, and together we visited Brown University and brainstormed numerous ideas. I also want to acknowledge the NeuroAI team members at CerCo, including Milad Mozafari, Romain Bielawaski, Bhavin Choksi, Javier Cuadrado, Mathieu Chalvidal, Benjamin Devillers, Colin Decourt, Ismail Khalfaoui, and Sabine Muzellec. Our lab meetings provided a platform for exchanging scientific insights and fostering collaboration.

I want to thank the members of the \textit{Serre Lab}, Aimen Zerroug, Thomas Fel, Mathieu Chalvidal, Lakshmi N. Govindarajan, Jacob Rose, Pachaya Sailamul, Ivan Rodriguez, Rex Liu, Lore Goetschalckx, Victor Boutin, and Drew Linsley, for their warm welcome, collaborative spirit, and sharing of knowledge. Their valuable feedback and insightful analyses have played a crucial role in refining my work.

I  also had the opportunity to collaborate with Remi Cadene (Senior Scientist, Tesla), who encouraged me to organize my thoughts before working on them; Drew Linsley (Asst. Professor Research, Brown University), who inspired me with his choice of words in scientific writing and positive attitude for any new idea. I greatly benefited from the conversation with Jonathan D. Cohen (Professor, Princeton University) and his group members, especially Taylor W. Webb (University of California, Los Angeles). Their extensive discussions on the GAMR architecture helped me to comprehend better. I thank Peter Wilf (Professor, Pennsylvania State University) for having me on board with him to execute practical aspects of my understanding of Paleobotany. Lastly, our ongoing collaboration with Experimental Neurosurgery and Neuroanatomy at KU Leuven introduced me to another aspiring scientist, Jesus G. Ramirez. This opportunity allowed me to understand neural visual reasoning mechanisms at the anatomical level.

My mind and heart owe Ashwani Sharma (Asst. Professor, IIT Ropar), K. R. Ramakrishnan (Emeritus Prof, IISc Bangalore) and Anil Kumar Tiwari (Assoc. Professor, IIT Jodhpur), Ranjan Gangopadhyay (Emeritus Prof, IIT Kharagpur). They fueled my scientific curiosity during various stages of my life and supported me in becoming a keen researcher.

I would be remiss not to mention my family, including my parents Smt. Vijaylakshmi Vaishnav and Shri. Bharat Kumar Vaishnav (Commandant, CRPF), my sister Dr. Divya Vaishnav (Assistant Professor, Chandigarh University), my brother-in-law Mr. Sunil Sharma (Senior Scientist, ISRO), and my younger brother Mr. Gaurav Vaishnav (Provincial Civil Service, Govt. of Bihar). Their unwavering belief in me and constant moral support have been the pillars of my motivation throughout this process. 

In addition to my family, I owe a debt of gratitude to my friends, scattered across different parts of the world, whose support has been invaluable. Special thanks to Dr. Dinesh K. Chobey, Mohit Ahuja, Parita Verma, Himanshu Vaishnav, Malav Bateriwala, Pragnya Paramita, and many others whose names I regrettably cannot mention individually.

\pagenumbering{roman}
\setcounter{page}{6}

\phantomsection
\addcontentsline{toc}{chapter}{Contents}
\tableofcontents
\newpage
\cleardoublepage

\pagenumbering{roman}
\setcounter{page}{9}

\phantomsection
\addcontentsline{toc}{chapter}{\listfigurename}
\listoffigures
\newpage
\cleardoublepage

\pagenumbering{roman}
\setcounter{page}{16}

\phantomsection
\addcontentsline{toc}{chapter}{\listtablename}
\listoftables
\newpage
\cleardoublepage

\mainmatter

\setlength{\parskip}{.7em}

\titlespacing*{\section}{0pt}{.9em}{.8em}
\renewcommand{\baselinestretch}{1.1}

\fancyhead[RO]{\leftmark}
\fancyhead[LE]{\textsc{\chaptername~\thechapter}}

\part*{Chapter 1}
\chapter{Introduction}

\startcontents[chapters]
\printmyminitoc

\epigraph{\itshape  Everyone knows what attention is. It is the taking possession by the mind, in clear and vivid form, of one out of what seems several simultaneously possible objects or trains of thought.}{-- William James}

Attention is a field widely discussed and studied in neuroscience, psychology, cognitive science and machine learning \citep{chun2011taxonomy,cho2015describing}. Attention is the process of selectively focusing on a discrete aspect of information while ignoring other perceivable information. A widely accepted feature of attention is that it facilitates efficient use of the available computational resources.

Over time, the understanding of attention has progressed significantly, leading to the development of various models aimed at elucidating its mechanisms. The hypothesis put forth by \citet{descartes2010passions}, suggesting that the pineal body controls attention, is now considered invalid, and alternative models have been developed to offer more accurate explanations. Early investigations focused on psychophysics, particularly exploring the range of visual attention.  For instance, \citet{Helmholtz1989} introduced the concept of covert attention, which proposes that visual attention can operate independently of eye movements. While initial approaches viewed attention as a top-down guided internal state,  Gestalt theory \citep{kohler1947gestalt} presented attention as a bottom-up computation, emphasizing the role of the focus of attention. These early models challenged Descartes' (1649) outdated hypothesis that the pineal body controls attention.

Neurophysiological findings by \citet{Sechenov1863} suggested that inhibition from the central nervous system plays a crucial role in attention control. This discovery spurred the development of computational models by \citet{tsotsos1995modeling} and theoretical frameworks by \citet{Pavlov1927} that incorporated inhibition as a core process in attention. According to Pavlov, unexpected stimuli capture attention (facilitation), while stimuli in nearby cortical areas are suppressed through inhibitory processes. Further exploration of the role of inhibition as an essential attention process was conducted by \citet{itti2009bayesian}. Notably, attenuation theory by \citet{treisman1980feature} proposed that unattended signals are attenuated rather than filtered, reconciling opposing viewpoints. These models and theories contributed to the understanding of attention as a mechanism for coping with limited information processing capacity \citep{broadbent1958perception}.

Broadbent's comprehensive attention model, grounded in cognitive psychology, viewed attention as a biological mechanism to cope with the limited capacity of information processing. This model sparked a debate regarding the stage at which selection occurs: early in the information pipeline, as proposed by Broadbent, or later in the selection process, as suggested by \citet{norman1968toward}. Treisman introduced the concept of attenuating unattended signals as a middle-ground perspective \citep{treisman1980feature}. \citet{milner1974model} proposed that attention not only selects relevant features but also provides feedback to early stages of information processing. This framework was incorporated into Adaptive Resonance Theory by \citet{grossberg1975neural}. Subsequent neurophysiological evidence demonstrated the top-down influence of the attentional state on the activation of perceptual circuitry, indicating that feedback can occur at any stage of the information pipeline.

Many computational models of visual saliency originate from Treisman et al. Feature Integration Theory (FIT) of spatial visual attention proposed in 1980. The FIT aimed to explain the performance difference between pop-out stimuli and conjunction search. Texton theory by \citet{bergen1983parallel} demonstrated that certain features allow for rapid discrimination of a target and surrounding outliers (pop-out), while others do not. According to the FIT, a saliency map, often referred to as a ``master map,'' integrates information from separate feature maps to identify salient locations. Treisman and Gelade also addressed the binding issue, which involves the cohesive representation of an object by binding its features (color, shape, location, etc.) together. \citet{koch1987shifts} proposed a computational implementation of the FIT, where the saliency map is a weighted sum.

Visual attention is a specific form of attention that operates within the visual modality. It involves the selective processing and allocation of attentional resources to visual stimuli. Visual attention allows us to prioritize and focus on specific visual features, objects, or regions within our visual field while suppressing or inhibiting the processing of other visual inputs. It can operate at different levels. At the early perceptual level, it involves the selection and processing of basic visual features such as color, shape, and motion. This early selection process helps filter out irrelevant visual information and enhance the salience of relevant visual stimuli. At a higher cognitive level, visual attention enables us to selectively attend to objects or regions of interest, guiding our gaze and directing our focus within the visual scene.

The mechanisms of visual attention include both bottom-up and top-down processes. Bottom-up attention is driven by salient or physically distinctive features of stimuli that automatically capture our attention, such as a bright color or sudden movement. Top-down attention, on the other hand, is driven by our goals, expectations, and prior knowledge. It allows us to voluntarily direct our attention to specific stimuli based on their relevance or importance in a given context. Visual attention plays a crucial role in various cognitive processes, such as visual perception, object recognition, scene understanding, and visual search tasks. It helps us efficiently process and interpret visual information, guiding our interactions with the visual world.

The cognitive science literature depicts several aspects of attention, such as it can be concentrated, it can focus on a particular modality, it can be divided, it can be selective, and it can have a finite capacity. However, selectivity is its most characteristic feature. Selective attention is necessary because of the limited availability of resources. Visual attention and selective attention are closely interconnected concepts that involve the cognitive process of allocating attentional resources to specific stimuli while ignoring or suppressing others. Visual attention refers to the selective processing and prioritization of visual information, while selective attention encompasses the broader ability to attend to stimuli across different sensory modalities.

Selective attention is the cognitive process of focusing on one or a limited number of sensory stimuli while disregarding irrelevant inputs. Various theories have been proposed to explain selective attention, including bottleneck theories and load theories. Bottleneck theories, such as Filter Theory \citep{broadbent1958perception}, Late Selection Theory \citep{deutsch1963attention}, and Attenuation Theory \citep{treisman1964selective}, focus on the flow and filtering of information. Load theories, like Perceptual Load Theory by \citet{lavie1994perceptual}, Dilution Theory \citep{tsal2010diluting} address the allocation of perceptual and cognitive resources. However, operationalizing these constructs and validating the theories can be challenging. Selective attention is essential for daily functioning, preventing overload of the information processing system. 

Early theories of attention, such as Donald Broadbent's Filter Theory, proposed a "bottleneck" model of selective attention, likened to a bottle with a narrow opening. According to Broadbent, stimuli enter a sensory buffer where their physical characteristics are assessed, allowing only a few to pass through the selective filter. Unselected stimuli decay in the buffer, while the selected ones proceed to be processed for meaning and determine how we respond. Broadbent used the dichotic listening task to study selective attention, finding that participants performed better when attending to one ear at a time. However, criticisms arose regarding where stimuli gain meaning within the attention process, with the cocktail party effect suggesting that analysis occurs before filtering.

Deutsch and Deutsch proposed the late selection theory as an alternative to address limitations of Broadbent's theory. They suggested that all stimuli are analyzed for meaning, but only selected ones pass the filter based on their physical characteristics and relevance. Anne Treisman introduced the attenuation theory, suggesting that stimuli are not filtered but attenuated or enter the sensory register at a lower intensity, gaining meaning early on. Her theory addresses the limitation of Broadbent's theory regarding the cocktail party effect. Treisman used the dichotic listening task with complete words and found that people often combine messages from both ears, implying that the unattended message still holds meaning regardless of retention. These early theories laid the groundwork for understanding selective attention and the processing of stimuli based on their relevance and physical characteristics.

Visual attention involves the mechanisms and processes that enable us to focus on relevant visual stimuli and filter out irrelevant or distracting visual inputs. It allows us to direct our attention to specific regions or objects within the visual field, selectively process their features, and integrate them into our perceptual experience. Visual attention plays a crucial role in various tasks, such as visual search, where we actively scan our environment to find a specific target among distractors. Selective attention, on the other hand, extends beyond the visual domain and encompasses attentional processes across different sensory modalities, including auditory, tactile, and cognitive inputs. It involves the ability to prioritize and allocate attentional resources to relevant stimuli or information while disregarding or suppressing irrelevant or less important stimuli from all sensory channels.

While visual attention is a specific subset of selective attention that focuses on the processing and filtering of visual information, it is interconnected with other modalities of attention. For example, during a complex task that requires both visual and auditory processing, selective attention allows us to prioritize the relevant visual stimuli while simultaneously attending to relevant auditory cues or instructions.

Recently visual attention has gained tremendous attention in the field of artificial intelligence. 
Visual attention helps in answering \textit{what} to look and \textit{where} to look. It has been vastly studied in psychology and neuroscience \citep{posner1990attention,cohen1990control,phaf1990slam,bundesen1990theory,desimone1995neural,mozer1998computational,corbetta2002control,o2006making,petersen2012attention,moore2017neural} and more recently by \citet{flesch2022orthogonal,dekker2022curriculum}. These studies have acted as a source of inspiration for several artificial intelligence models \citep{khosla2007bio,lindsay2018biological} including the ones proposed in this thesis. 

There are three categories of selectivity in a visual attention system: by spatial location \textit{(space-based)} \citep{posner1980orienting,posner1982neural}, by object membership \textit{(object-based)}~\citep{duncan1984selective,egly1994shifting,vecera1994does,kramer1997object} and by particular features of the input \textit{(feature-based)}~\citep{harms1983color,driver1989movement,kramer1991perceptual,baylis1992visual,duncan1996objects}. 

\paragraph{Visual Spatial Attention}

Every second, our eye makes small and rapid movements several times, known as saccades. These eye movements change the locus of attention. Visible shifts of attention, such as saccades, are known as \textit{overt} visual attention. One more method used to emphasize a spatial location without any over-the-shift of the fovea location is \textit{covert} attention. An example is the subject's fixation on a particular region throughout a task where the stimulus is likely to appear. This region is also referred to as the ``\textit{spotlight}'' of attention. Certain visual patterns that involve edges, contrast, or motion automatically attract attention. These patterns are known as ``\textit{salient}'' \citep{itti2001computational}. In the presence of task-specific information, these saccadic movements are controlled in a top-down fashion around the particular visual target instead of the salient regions. Eye movements are one of the possible ways to control visual attention.

\paragraph{Visual Feature Attention} 
When the focus of attention is on features like color, shape or orientation instead of location, it is known as feature-based attention. It is an example of covert visual attention. Cueing the right features enhances the system's performance. It is used in tasks such as visual search combining covert feature-based attention with overt attention. Feature-based attention is global as opposed to spatial attention, i.e., when attention is focused on a particular feature, neurons representing that particular feature in the visual space are also modulated~\citep{saenz2002global}. It is related to object attention, i.e., instead of attending to an abstract feature, the attention is deployed at a specific object in a visual scene~\citep{chen2012object}. A single feedforward pass in the visual hierarchy can segregate the objects of a visual scene if there is a distinct salient difference between them as opposed to a complex scene where recurrent and serial processing might be required~\citep{lamme2000distinct}. 

In addition to feature-based or spatial attention, another widely accepted classification is characterized by the type of data processing~\citep{connor2004visual,buschman2007top}. There are two types of data processing, \textit{bottom-up} and \textit{top-down}. In a bottom-up attention process, external factors guide the attentional process because of their inherent properties, like their color or sudden motion in the scene. It is fast and primitive sensory driven. In top-down attention, there is an internal attentional guidance mechanism based on prior knowledge and current goals, like searching for food if one is hungry. It can ignore the salient stimuli and focus on the target object or event. 

\begin{figure}[htbp]
\centering
  \includegraphics[width=.7\linewidth]{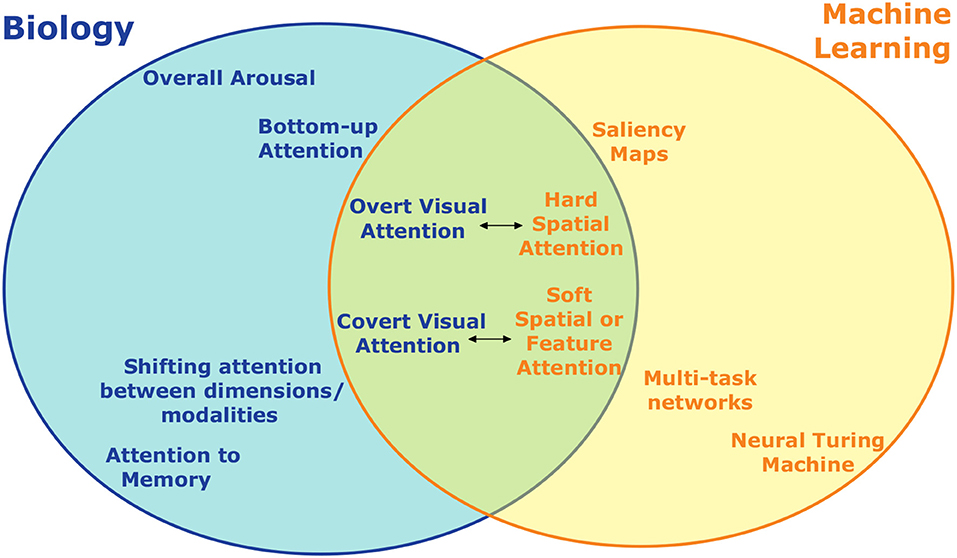}
  \caption{A summary of attention in Cognitive science and machine learning (\href{https://www.frontiersin.org/articles/10.3389/fncom.2020.00029/full\#h5}{source})} \label{fig:atnml}
\end{figure}

Attention is also involved while performing tasks requiring multiple sensory signals. In the presence of multiple tasks or sensory signals, the central executive controller helps to route the focus of attention. The Central executive controller is responsible for coordinating activity with the cognitive system for directing attention, decision making and maintaining task goals. Context and history are deemed helpful to executing tasks optimally -- making it highly related to the working memory. Attention is furthermore seen as the output of the central controller. The controller selects the targets of attention and passes them to the system responsible for its implementation. There is a three-way relationship between executive control, working memory and attention in such a way that the focus of attention is selected by the executive controller based on the contents of the working memory ~\citep{soto2008automatic}. Although all the objects in the working memory can influence attention, the executive controller helps decide which one should affect the most~\citep{olivers2011difference}. These vast and extensive cognitive studies related to attention have inspired the field of AI and helped to boost its performance (Figure~\ref{fig:atnml}).

The first attempt to adapt the attention mechanism in a neural network was made in the 1980s when the improved version of \textit{Neocognitron} \citep{Fukushima1980} incorporated selective attention \citep{fukushima1987neural} to decompose the image into elementary features. Later, \citet{Fukushima1993} modified the network to recognize and segment characters in cursive handwriting. \citet{Postma1997} proposed an attentional scanning model, \textit{SCAN}, to attend to and identify object patterns without decomposing the scene into elementary features. As an alternative to these static neural approaches, \citet{Schmidhuber1991LearningTG} proposed a sequential model inspired by the sequential eye movements for object detection. In this model, a neural controller learns sequential generation of fovea trajectories to reach the target. Furthermore, data processing types inspired the development around the same time, thereby leading to a model extracting the region of interest using bottom-up and top-down processing~\citet{MilaneseIntegration1994}.

By the early 2000s, the influence of attention on the evolution of neural networks increased. \citet{miau2001neural} proposed a model of primate vision integrating both, \textit{what} and \textit{where} pathways. The model has a fast visual attention-based front-end to select the most salient image areas and a slow back-end to recognize objects in those selected areas. Another model based on the primate selective mechanism is presented in \citet{salah2002selective} with the idea of selectively attending to relevant parts of the input image. In this model, a neural network analyzes the input image and generates posterior probabilities for the Markov models. Attention has also been used for object recognition \citep{walther2002attentional} and scene analysis \citep{schill2001scene}. 

The year 2015 marks the new beginnings of attention-based architectures with the introduction of the attentional model for Neural Machine Translation (NMT) \citep{Bahdanau2014-vr,luong-etal-2015-effective} and image captioning \citep{xu2015show}. In NMT, the expectation is to learn continuous representations of variable-length sequences. \textit{Recurrent neural networks} (RNNs) like LSTMs \citep{10.1162/neco.1997.9.8.1735}, GRUs \citep{cho-etal-2014-properties} and Quasi-RNNs \citep{DBLP:conf/iclr/0002MXS17} were some of the popular sequence models for representation learning at that time. While these RNNs' output depends on the previous elements in a sequence, traditional feedforward neural networks assume that inputs and outputs are independent of each other. Nonetheless, their limitation includes their inability to parallelize computations -- making them slow during training and their fixed-size memory -- bottleneck for long-range interactions \citep{vaswani2017attention}. 

Models used for NMT typically consist of encoder-decoder architecture \citep{cho-etal-2014-learning}. Typically, both encoder and decoder are RNN, where the encoder takes an input sequence of fixed-length vector and represents it again to another fixed-length vector. A decoder then takes this encoded vector to generate the output sequence token by token. However, this method has two challenges; first, the encoder compresses the input sequence into a fixed vector length which may lead to the loss of information \citep{cho-etal-2014-properties}. Second, the model is incapable of aligning between input and output sequences which is essential for tasks such as translation or summarization \citep{Young2018trends}. While generating the output sequence, the decoder also lacked the mechanism to selectively focus on relevant input tokens. Later, \citet{bahdanau2014neural} proposed a sequence-to-sequence modeling task with the help of soft attention, emphasizing the parts of the sentence relevant to predicting the target word. \citet{bahdanau2014neural} extended the basic encoder-decoder by letting the model search a set of input words while generating target words. It allowed the model to focus on information needed to generate the subsequent target sequence. 

In the following two years, the adoption of attentional mechanisms in neural networks diversified. Content-based soft attention mechanism \citep{goodfellow2016deep} is used in Neural Turing Machine (NTM) \citep{graves2014neural} with end-to-end training. Around the same time, \citet{cheng2016long} used a form of attention called intra-attention in the Long Short-Term Memory (LSTM) \citep{hochreiter1997long} architecture. \citep{hochreiter1997long} embedded a memory network inside the LSTM architecture to store the contextual representation of the input. This memory network has a set of key and value vectors in the hidden state to represent what is stored in the memory. These vectors are used to estimate the intra-attention with the previously stored tokens in the memory as opposed to the self-attention mechanism used by \citet{vaswani2017attention} where interaction between the whole input sequence is estimated. One of the first uses of the self-attention mechanism in NLP is done by \cite{parikh-etal-2016-decomposable}. 

Since then, self-attention mechanisms have become an integral part of sequence modeling allowing the network to model dependencies between input and output sequences irrespective of their distances. A self-attention layer calculates a single-shot interaction between all pairs of words in a sequence. 

\section{Self-attention-based Transformer architecture}
\begin{figure}[ht]
\centering
  \includegraphics[width=1\linewidth]{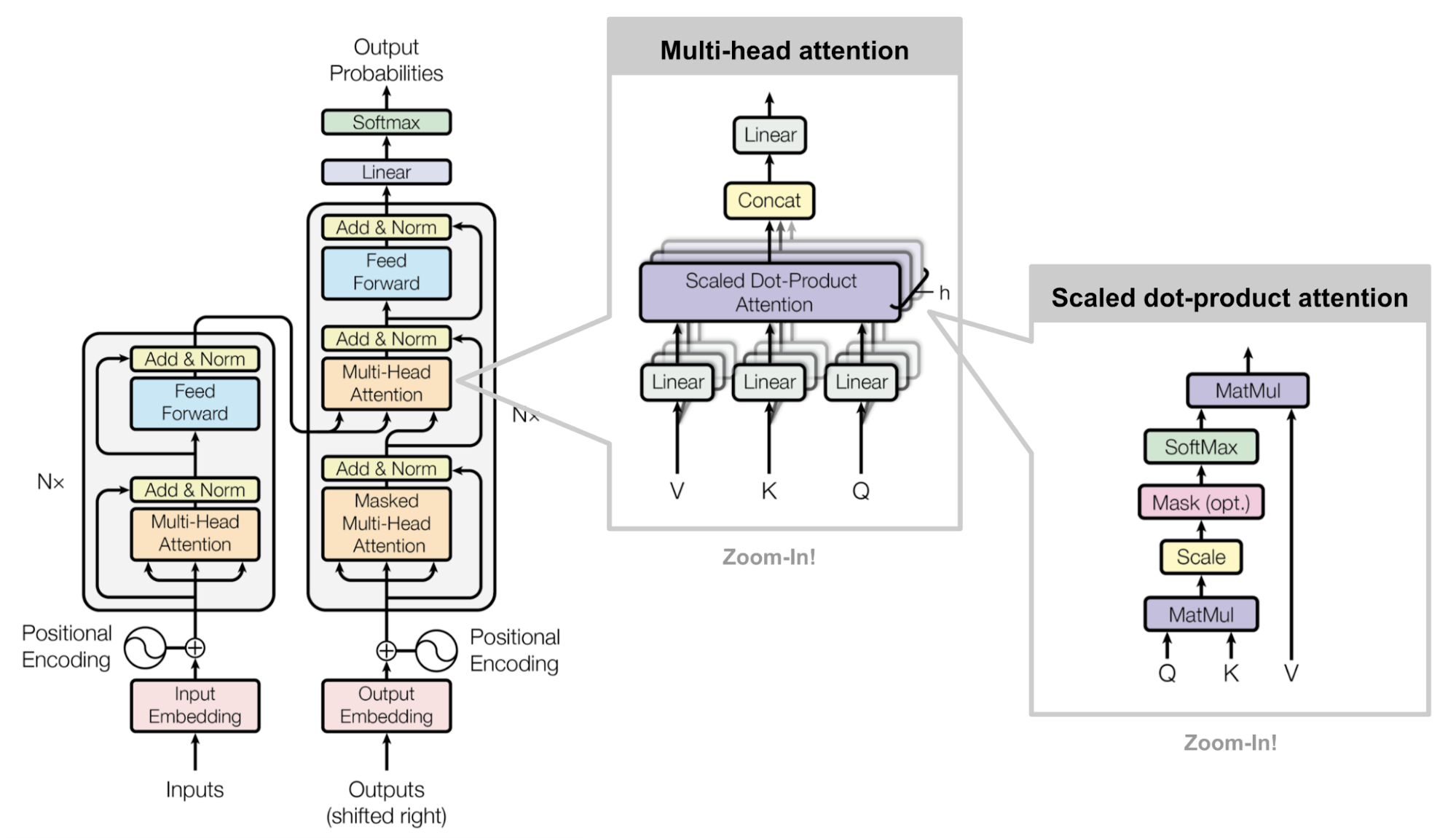}
\caption{Transformer architecture proposed by \citet{vaswani2017attention} (\href{https://deepfrench.gitlab.io/deep-learning-project/}{source})} 
\label{fig:vaswani}
\end{figure}

In 2017 \citet{vaswani2017attention} proposed a novel architecture, \textit{Transformer} for NLP. It is predominantly a self-attention network driving the waves of advances in AI. A Transformer architecture (Figure~\ref{fig:vaswani}) includes a stack of encoder and decoder blocks. Each encoder block is identical and contains a self-attention layer and a feedforward layer. The encoder's input flows through the self-attention layer helping the encoder to look at other words while encoding the current word. Its output is then fed to the feedforward layer. The same feedforward network is applied independently to each word. While a decoder consists of an encoder-decoder attention block in addition to the self-attention layer and a feedforward layer helping the decoder to focus on relevant parts of the input sequence.

In the NLP task, each word of the input sequence is first converted into an embedding vector. They are provided as input to the encoder block, passing through a self-attention layer and feedforward network. The obtained output vector is fed to the next encoder block. Using the self-attention layer, the Transformer models the relationship between the current word with other relevant words of a sequence. 

In a self-attention layer, its input vector is transformed into a key ($K$), query ($Q$) and value ($V$) vectors of dimension $d_q$ = $d_k$ = $d_v$ = 512 using a learnable matrix transformation. At first, the score ($S$) is calculated to determine the amount of focus to place on the other words in a sequence while encoding the current word. This score is calculated using the dot product between the query and key vectors ($S~=~Q~.~K^T$). It is normalized ($S=S/\sqrt{d_k}$) to stabilize the gradients, and later, using a $softmax$, converted into probabilities. The extent of the probability score shows the relevancy of the current word with other words in the sequence. This score is multiplied by the value vector ($V$) so that relevant words are given additional focus while irrelevant words are neglected in the subsequent layers. 

\[ Attention~(Q,K,V)~ = softmax(\frac{Q~.~K^T}{\sqrt{d_k}})~.~V\]

\begin{figure}[htbp]
\centering
  \includegraphics[width=1\linewidth]{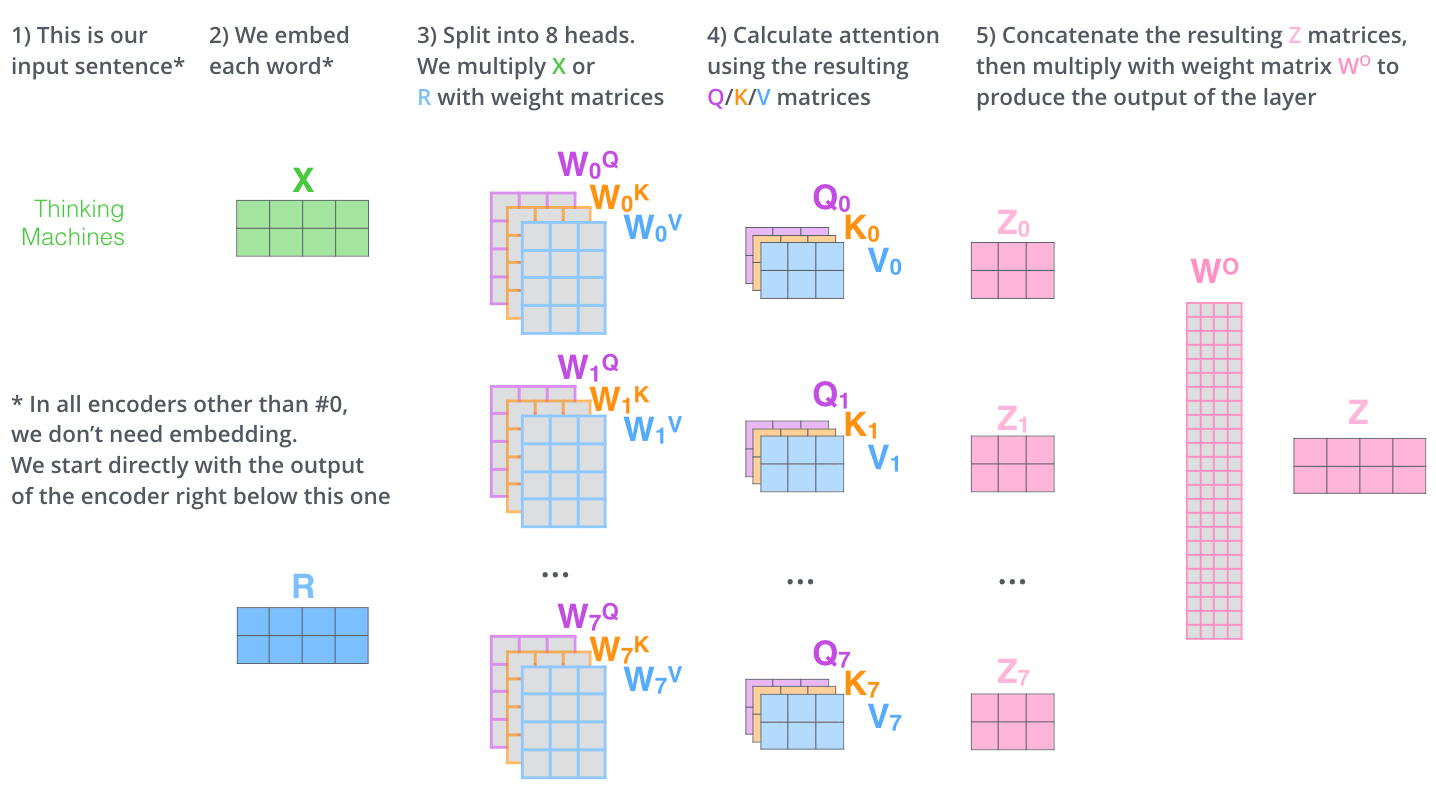}
\caption{Illustration of Multi-head attention mechanism in a Transformer network (\href{https://jalammar.github.io/images/t/transformer_multi-headed_self-attention-recap.png}{source})} 
\label{fig:intromha}
\end{figure}

The self-attention mechanism proposed by \cite{vaswani2017attention} has an additional feature called \textit{multi-head} attention (MHA). It helps to improve the performance in two ways: by augmenting the network's ability to focus on multiple positions and by giving distinct representational subspaces to each word. For example, if there are eight heads, eight sets of K, Q and V matrices exist, each representing a unique representational subspace. They are concatenated before passing through the feedforward network (Figure~\ref{fig:intromha}). 

The key characteristic of NLP tasks is the order of the words in a sequence. However, the operations we have discussed till now are permutation invariant. Positional embedding vectors are added to each input embedding vector to address this issue. These vectors help the model estimate the position of each word in a sequence in the projection space (i.e., $K/Q/V$).

Positional encoding in the Transformer is an active and vibrant research area. Vanilla Transformer \citep{vaswani2017attention} uses absolute positional encoding; however, more recent work \citep{DBLP:journals/corr/abs-1810-04805,Dosovitskiy2020-iq} prefers a learned \citep{gehring2017convolutional} or relative positional encoding \citep{shaw2018self}. The absolute coordinate system does not encode translational equivariance, while relative geometry could. \cite{ramachandran2019stand,bello2019attention} studied different positional encoding techniques and established that relative positional encoding offers the best results while providing additional advantages like encoding for an unseen length of sequences (refer to \cite{wu2021rethinking} for a review). An overview of different positional encoding strategies used in NLP is discussed by \citet{dufter2021position}.

The residual connection around the self-attention layer and a feedforward network is an essential module in an encoder block. It is followed by the layer normalization step~\citep{baevski2018adaptive,wang2019learning,Dosovitskiy2020-iq}. A residual connection is added to each sub-layer in the encoder (and decoder) to strengthen the flow of information and achieve higher performance.

At the end of the encoding steps, decoding begins. The decoder uses the key ($K$) and value ($V$) vectors from the top-most encoder block for its encoder-decoder attention layer. It helps to focus on the appropriate locations of the input sequence. At each step, the decoder layer provides an element of the final output sequence. This output vector is again fed to the subsequent decoder layer in the next time step. This process continues till the end of the sequence. An independent set of positional encoding is applied on the decoder side. 

The encoder-decoder attention module is similar to the multi-head self-attention mechanism described earlier. The only difference is that the key $K$ and value $V$ vectors are obtained from the top-most encoder block, and the query vector $Q$ is derived from the previous self-attention layer of the decoder. Unlike the encoder, self-attention layers in the decoder are only allowed to access previously obtained output by masking the future words of a sequence. Masking future positions is done to prevent the decoder from cheating during the training phase -- otherwise, it will already know what is coming next. The linear layer at the end of the decoder block is a fully connected neural network. It projects the vector obtained from the decoder layers into a logit vector. This logit vector represents the complete vocabulary of the language where translation has to be performed. A softmax converts this logit to the probabilities and represents the concerned word from the available vocabulary. 

In terms of computational complexity, for a sequence of length $n$ and dimensionality $d$, self-attention layers are faster than recursive or convolutional layers when $n$ is smaller than $d$, which is typically the case.

\begin{table}[ht]
\begin{center}
\vspace{.1in}
\begin{tabular}{cccc}
\hline
\textbf{Layer Type}     & \textbf{Complexity} & \textbf{Sequential } & \textbf{Maximum} \\
        & \textbf{per Layer} & \textbf{operations} & \textbf{ path length} \\\hline
Self-Attention & $\mathcal{O}$($n^2.d$)  & $\mathcal{O}$(1)    & $\mathcal{O}$(1)      \\
Recurrent      & $\mathcal{O}$($n.d^2$)  & $\mathcal{O}$(n)    & $\mathcal{O}$(n)      \\
Convolutional  & $\mathcal{O}$($k.n.d^2$) & $\mathcal{O}$(1)   & $\mathcal{O}$($log_k$(n))  \\\hline                      
\end{tabular}
\caption{Complexity comparison of different networks for a sequence of length $n$, kernal size $k$ and dimensionality $d$ \citep{vaswani2017attention} }
\label{tab:complexity}

\end{center}
\end{table}

\section{Self attention in vision tasks}

\paragraph{Why self-attention for vision?}

In a self-attention mechanism, each word of a sequence is correlated with all the others. Thus containing information about the rest of the sequence -- increasing the receptive field size equivalent to the length of a sequence. In some sense, images are no different from NLP sequences. Computer vision can take inspiration from the NLP domain to model long-range interactions between pixels with the added benefit of multi-head attention helping to parallelize these interactions. With the help of the multi-head attention method, different heads can focus on modeling different relations between pixels. For example, in a visual reasoning task where the objective is to count the number of pairs of shapes in an image, one head can focus on finding a pair while the other can focus on counting them. It helps the network to model self-similarity within an image. Images such as natural scenes and paintings display a great amount of self-similarity. Such non-local self similarity property was earlier explored for applications such as texture synthesis \citep{efros1999texture}, object detection and segmentation \citep{Wang_2018_CVPR}, bilateral filtering \citep{tomasi1998bilateral} and image classification \citep{pmlr-v80-parmar18a}. Hereafter, the main focus of this thesis will be computer vision.  

\paragraph{Self-attention with CNN}

In a computer vision task, the resolution of the images could reach around 1000$\times$1000 px. Applying a self-attention mechanism to all these pixels ($~10^6$ in number) is computationally expensive because of the quadratic complexity associated with the length of the sequence. Convolutional layers, on the other hand, do not have this bottleneck. However, they face trouble capturing long-range interactions because of their inability to scale up with the large receptive fields. 

To address this problem, there are predominantly two approaches. The first is to reduce the self-attention operation cost to a linear scale. Aligned to this line of work, \citet{ramachandran2019stand} proposed a pure stand-alone attention model for vision tasks by replacing the convolution operations with self-attention operations. Nonetheless, the self-attention operation used in this approach is local. Another similar linear attention variant Halo~\citep{vaswani2021scaling} uses block-wise local attention to improve speed and accuracy. 

The second approach is to build hybrid CNN-Transformer architectures where the convolutions operations are used to encode the input image, and attention is applied to those encoded features. \citet{srinivas2021bottleneck} explored a hybrid combination of CNNs and multi-head self-attention (MHSA) models and showed that replacing the $3\times 3$ kernal size convolutional layer in the bottleneck blocks of ResNet \citep{he2016deep} with MHSA layers improved several CNN baselines. Interestingly, DETR \citep{carion2020end} showed that concatenating the Transformer model at the end of the feature-extraction network is helpful for tasks like detection, localization, and segmentation.

There are four broad categories of research to incorporate self-attention mechanism with CNN, which are as follows:

\textbf{Inserting few attention modules in between residual blocks}: Along this line of work, \citet{Wang_2018_CVPR,chen20182} proposed a non-local block similar to \citet{ramachandran2019stand} and used them for video-based applications. In this network, features are gathered and propagated motivated by the squeeze and excite \citep{hu2018squeeze} network. As mentioned earlier, these methods only focus on the spatial dimension for calculating the non-local interaction, so \citet{yue2018compact} added a correlation factor between the channels to improve the model effectiveness. Similarly, \citet{Shen_2021_WACV} proposed a method to bring down the quadratic complexity of the self-attention mechanism to a linear scale. We demonstrate a unique way to incorporate self-attention with a feedforward network in \hyperref[chapter1b]{Chapter 3} where the intermediate features of the network are passed through the self-attention layer to find the global association. This attention is applied directly over the feature space in contrast to the previously used methods of squeezing the feature vector dimensions to save computations. 

\textbf{Inserting attention modules at the end}: Usually, such models have a front-end of convolutional block acting as a feature extraction module for self-attention block as back-end. These models are used for tasks like object detection and semantic segmentation. \citet{huang2019ccnet} designed criss-cross attention that learns the complete image dependency recurrently for semantic segmentation tasks using dot-product attention. Moving away from this trend of using self-attention operations, \citet{carion2020end} proposed DETR architecture by placing a Transformer model as the back-end. 

\textbf{Replacing convolution layers by self-attention layers}: Self-attention mechanism used in this line of research is primarily local in nature to decrease the computational demand associated with the increasing sequence length in an image which is directly proportional to total pixel count. \citet{bello2019attention} made a unique attempt to augment the feature maps of convolutional layers with the self-attention modules. Feature maps obtained with the help of the self-attention module are concatenated with the feature maps of CNNs. They discovered that replacing all the feature maps of CNN with the feature maps of self-attention layers degrades the system's performance. Contrary to their finding, \citet{ramachandran2019stand} came up with the architecture replacing all the convolution layers with a local self-attention layer and achieved better performance than a fully convolutional network on the image classification task. 

In addition to these four categories of research, where the primary focus is on computer vision applications, cognitive studies also explore self-attention mechanisms. In one of the first studies by \citet{whittington2022relating}, neural representations of Hippocampal formation are related to the Transformer model. They did this correspondence with the help of the Tolman-Eichenbaum Machine \cite{whittington2020tolman}, a model for hippocampal formation. This work showed that when recurrent positional encodings are used in Transformer, they replicate spatial representations of hippocampal formations like place cells and grid cells. To analyze from an attentional point of view, we studied the role of a self-attention layer of the Transformer model in understanding visual reasoning tasks in \hyperref[chapter1b]{Chapter 3}. This layer is used as a feature-based or spatial attention layer. A multi-head self-attention layer is significantly different from the other existing self-attention models where the span of attention in the dot-product mechanisms is local. They proposed a self-attention mechanism that could be applied globally over a feedforward network's spatial or feature space. This method gives the network higher representation power because of its ability to use multi-head attention. We also built a cognitive architecture inspired by the active vision literature relating to the shifting of the spotlight of attention in \hyperref[chapter2b]{Chapter 4}. This attention routing is implemented with a controller module consisting of an attention module and an LSTM layer, which generates a query to guide the shifting. More studies in the NLP domain focus on relating language models to brain activations; however, a similar trend is yet to be seen in the computer vision domain. 

These developments exploring the self-attention mechanisms propelled toward building a fully self-based attention architecture for computer vision applications. Evolutions in the NLP domain were vital in inspiring the fully self-attention-based architecture for vision tasks.

\section{Transformer-based vision architecture}

\begin{figure}[htbp]
\centering
  \includegraphics[width=1\linewidth]{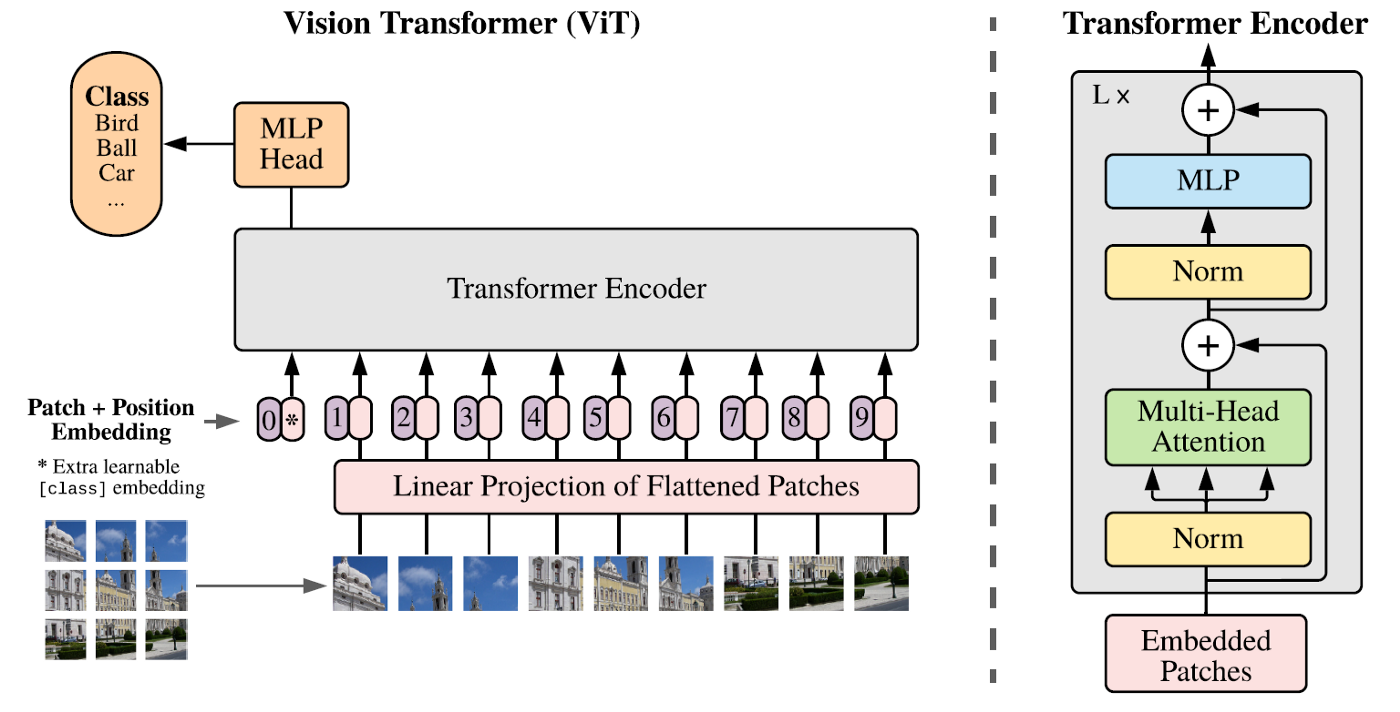}
\caption{Vision Transformer architecture \citep{Dosovitskiy2020-iq}} 
\label{fig:vit}
\end{figure}

The first fully self-attention-based Transformer architecture is presented by \citet{Dosovitskiy2020-iq}. It is known as Vision Transformer (ViT) (Figure~\ref{fig:vit}). In this architecture, an input image is divided into a sequence of image patches called visual tokens and transforms those patches before passing them to the network. The core idea is to treat each pixel as a token and pass it to the Transformer network. However, with the increasing size of the number of pixels, attention cost scales quadratically, so patches of 16$\times$16 pixels are used instead. Each patch is flattened and linearly projected to a vector of the desired dimension. As the network is agnostic to the positions of these patches w.r.t. the input image, position embeddings are added to learn the 2D structure. ViT learns this encoded structural information while training. A learnable class embedding token is also added at the beginning of the sequence. A class embedding token is inspired from \citet{devlin2018bert} that is learned along with other patches while training the network. This learnable token eventually helps to predict the classification label with the help of a multi-layer perceptron (MLP) head.

When the ViT is trained on a mid-sized dataset like ImageNet \citep{deng2009imagenet}, outcomes are not impressive because of their lack of inductive biases such as translational equivariance and locality. ViT experiences difficulty learning image-specific inductive biases like a CNN, as the model never sees the complete 2D image during training but only a sequence of transformed patches. Such CNN-like biases are compensated by training the model with massive databases like JFT-300M and fine-tuning it for downstream tasks. ViT learns the spatial relationship from scratch, which raises its demand for extra training data and longer training time. \cite{touvron2021training} proposed DeIT, equalizing this pre-training-related bottleneck using techniques like the teacher-student distillation approach and robust augmentation methods. DeIT, when trained on ImageNet by incorporating these methods, surpasses the performance of the ViT model. 

Vision transformers are a front-runner in capturing the long-range dependencies in an image, yet they fail to account for local features as CNNs do. A wide gap is perceived between ViT and CNN learnability. \cite{wu2021cvt,guo2022cmt,yuan2021incorporating,graham2021levit,dai2021coatnet,Peng_2021_ICCV} analysed the potential weaknesses in directly applying Transformer model from NLP domain and proposed a combination with convolutional network. \cite{wu2021cvt} proposed a Convolutional vision Transformer (CvT) and presented a convolutional-based patch projection of image tokens along with a hierarchical design. Another alternative, LocalViT \citep{li2021localvit} proposed depthwise convolution to capture local features.
Meanwhile, LeViT \cite{graham2021levit} enhanced the inference speed of ViT by designing multistage transformer architecture and downsampling the image using attention. In yet another network proposed by \cite{zhou2021elsa}, it incorporated locality without convolutions with the help of enhanced local self-attention using Hadamard attention and ghost head. Hadamard attention is more computational-friendly than dot-product attention, while ghost heads increase the channel capacity by combining attention maps. 

A striking network, ConViT, proposed by \cite{d2021convit} took a step further to incorporate the convolutional biases into the Transformer architecture. \cite{d2021convit} initialized self-attention layers with soft convolutions with the help of Gated-Positional-Self-Attention (GPSA). This self-attention block is characterized by locality strength and head-specific center of attention. The locality factor determines how much the head should focus around its center of attention. For any given query patch, which head should give attention to which position is decided by the head-specific center of attention. With suitable parameters setting, ConViT can have ViT-like expressive power and could be trained in low-data regimes like CNNs. In a collaborative project with paleobotanist, we proposed \textit{conviformer} \citet{vaishnav2022conviformers} to incorporate convolutional biases into any vision transformer with minimal architectural change. With the conviformer architecture, the network can also attend to higher-resolution images and provide compatibility with the base architecture used. 

\paragraph{Challenges}

\begin{figure}[htb]
\centering
  \includegraphics[width=.7\linewidth]{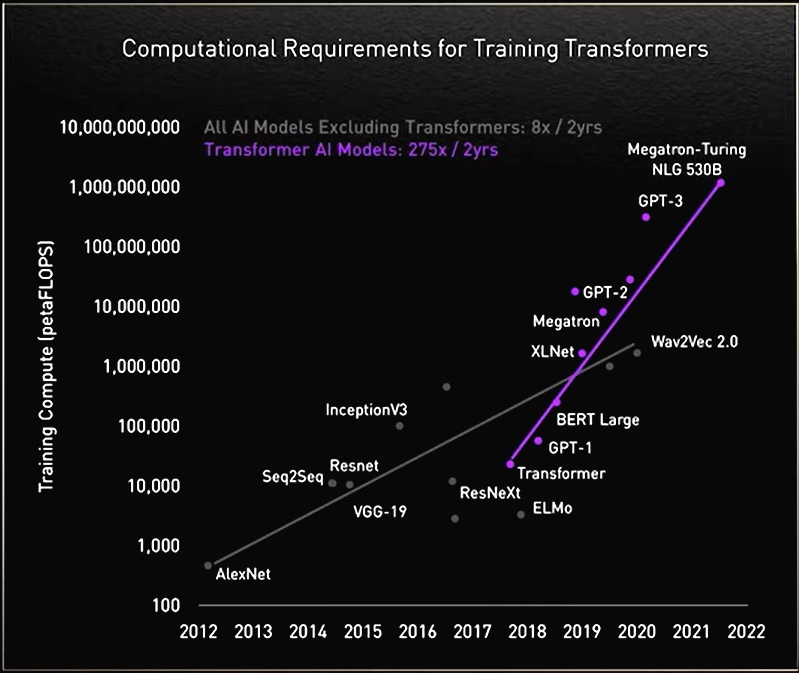}
  \caption{Computational demands for training Transformers vs. CNNs. Compute needed to train a Transformer network has increased by 275 times in the last two years. (\href{https://blogs.nvidia.com/blog/2022/03/25/what-is-a-transformer-model/}{source})} \label{fig:transcomp}
\end{figure}

Transformer architecture confronts a two-front challenge. It requires enormous data for training to learn the right inductive bias and the computational cost associated with the sequence length. Figure~\ref{fig:transcomp} compares the computational requirement of different Transformers and CNNs models. An empirical study is done by \citet{zhai2022scaling} on the scalability of the ViT. They report that scaling up training samples and parameters of the model scale up the overall performance of the model; nevertheless, this plateaus quickly for smaller models as they cannot leverage additional data. It indicates that larger models have the scope to improve their representation learning abilities. Training a Transformer model requires massive data to compete with inductive biases like translation invariance similar to CNN. The self-attention mechanisms in a Transformer learn such image-specific concepts during longer training times, thereby significantly increasing the compute requirements. Strong data augmentation techniques nowadays compensate for the vast dataset requirement. 

Transformer architecture furthermore lacks explicit mechanisms to attend to local neighborhoods. A commonly accepted solution to this issue is to restrict the attention mechanism to the local area \cite{parmar2018image} or to incorporate structural priors on attention like Sparsity \citep{child2019generating}. It makes a dense attention matrix into a sparse matrix limiting the computations. Regardless, the approach has some limitations. Sparse matrix multiplication operations are uncommon for hardware accelerators. 

An additional computational bottleneck is calculating the dot product operation in the self-attention layer. Existing techniques to handle this situation are half-precision, gradient accumulation and gradient checkpointing. Tensor computations on modern hardware architectures are effectively done with 16-bit float tensors. Sometimes higher precision is required while calculating the loss, which doubles the required memory. This precision handling is carried out with the help of \textit{apex} library\footnote{\url{https://nvidia.github.io/apex/}}. On a fixed GPU/TPU machine, a large model may only fit a single-digit batch size, ultimately leading to unstable learning. A multivariate chain rule is used to incorporate the dynamics related to bigger batch sizes. It sums the gradients for a larger batch and computes the gradient descent at the end. For more bigger models, the trade-off is to separate the model into different chunks and compute the gradient in a forward/backward pass for each chunk.   

Our proposed \textit{conviformer}  \cite{vaishnav2022conviformers} addressed ViT's inability to process longer sequences which restricted ViT to smaller resolution images. In the conviformer, the input image is passed through a convolutional backbone, down-sampling the image to 224$\times$224 (a commonly accepted input resolution). With the help of a convolutional frontend, the network makes sure to introduce the inductive biases of CNN into the network. The feature vectors obtained by the CNN modules are later passed to the base architecture of the vision transformer. This technique holds the compatibility of the network with the base model and provides a performance boost with insignificant additional computational cost.

Finally, training a huge Transformer model has negatively impacted the environment. Compute cost and the complexity associated with the Transformer are directly related to environmental factors such as $CO_2$ emission \citep{Strubell2020} and high energy consumption \citep{You2020Drawing}. There is also a cost associated with mining rare metals for manufacturing these hardware accelerators. 

\section{Original Contributions}
Our contributions are as follows:
\begin{itemize}
    \item We present a novel fine-grained taxonomy for the SVRT tasks by systematically analyzing the ability of feedforward neural networks.
    \item We first propose a self-attention-augmented feedforward network modeled as spatial or feature-based attention. 
    \item Our attentional networks analysis on SVRT tasks provides a granular computational account of visual reasoning and yields testable neuroscience predictions regarding the differential need for feature-based versus spatial attention depending on the type of visual reasoning problem.
    \item Next, we present a novel end-to-end trainable guided-attention module to learn to solve visual reasoning challenges in a data-efficient manner.
    \item We show that our guided-attention module learns to shift attention to task-relevant locations and gate relevant visual elements into a memory bank; 
    \item We show that our architecture demonstrate zero-shot generalization ability and learns compositionally. GAMR is capable of learning efficiently by re-arranging previously-learned elementary operations stored within a reasoning module. 
    \item Our architecture sets new benchmarks on two visual reasoning challenges, SVRT~\citep{fleuret2011comparing} and ART~\citep{Webb2021EmergentST}. 
\end{itemize}

The work presented in \hyperref[chapter1a]{Chapter 2} and \hyperref[chapter1b]{Chapter 3} are taken from our following publication:
\begin{itemize}
    \item \textbf{Mohit Vaishnav}, Remi Cadene, Andrea Alamia, Drew Linsley, Rufin VanRullen, Thomas Serre; ``Understanding the Computational Demands Underlying Visual Reasoning.'' \textit{Neural Computation} 2022; 34 (5): 1075–1099. doi: \url{https://doi.org/10.1162/neco_a_01485}
\end{itemize}

The work presented in \hyperref[chapter2b]{Chapter 4} is taken from our following publication:
\begin{itemize}
    \item \textbf{Mohit Vaishnav}, Thomas Serre. ``GAMR: A Guided Attention model for (visual) Reasoning.'' \textit{International Conference on Learning Representations (ICLR)} 2023, \href{https://openreview.net/forum?id=iLMgk2IGNyv}{https://openreview.net/forum?id=iLMgk2IGNyv}
\end{itemize}

\part*{Chapter 2}
\chapter[Computational Demands of Visual Reasoning]{Understanding the computational demand underlying visual reasoning}
\label{chapter1a}

\startcontents[chapters]
\printmyminitoc

\section{Introduction}

Humans can effortlessly reason about the visual world and provide rich and detailed descriptions of briefly presented real-life photographs \citep{Fei-Fei2007-zb}, vastly outperforming the best current computer vision systems \citep{Geman2015-jm, Kreiman2020-zd}. 
For the most part, studies of visual reasoning in humans have sought to characterize the neural computations underlying the judgment of individual relations between objects, such as their spatial relations (e.g., \citet{logan1994ability}) or whether they are the same or different (up to a transformation, e.g., \citet{shepard1971mental}). It has also been shown that different visual reasoning problems have different attentional and working memory demands \citep{logan1994spatial,Moore1994,Rosielle2002,Holcombe2011,vanderham2012,kroger2002recruitment,Golde2010,clevenger2014working,brady2015contextual}. However, there is still little known about the neural computations that are engaged by different types of visual reasoning (see~\citet{ricci37same} for a recent review). 

\begin{figure}[t]
\centering
  \includegraphics[width=1\linewidth]{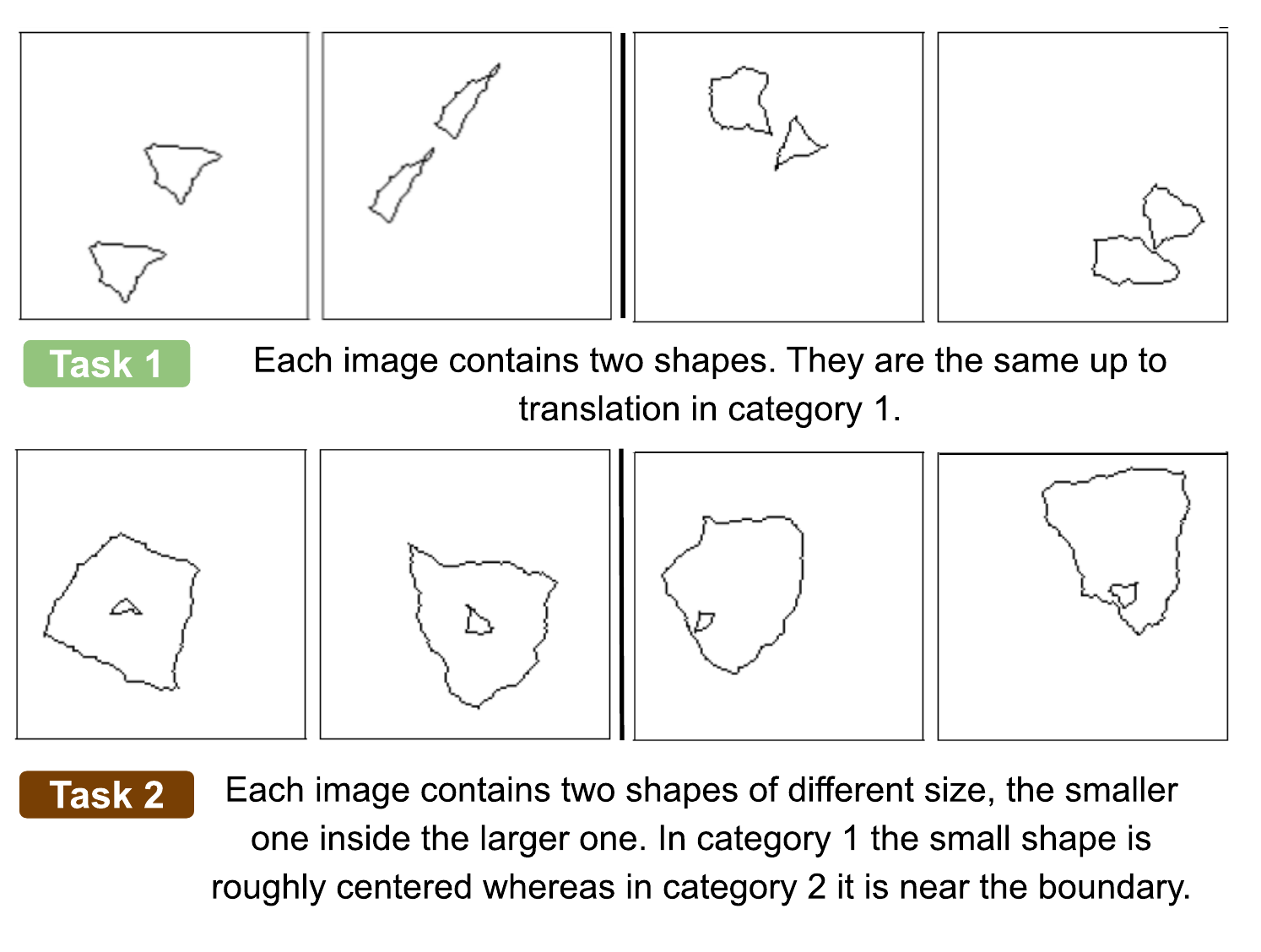}
 \caption{Two SVRT sample tasks from a set of twenty-three in total. For each task, the leftmost and rightmost two examples illustrate the two categories to be classified. Representative samples for the complete set of twenty-three tasks can be found in Figure~\ref{fig:exampleSD} and~\ref{fig:exampleSR}.} \label{fig:example}
\end{figure}

One benchmark that has been designed to probe abstract visual relational capabilities in humans and machines is the \textit{Synthetic Visual Reasoning Test} (SVRT) ~\citep{fleuret2011comparing}. The dataset consists of twenty-three hand-designed binary classification problems that test abstract relationships between objects posed on images of closed-contour shapes. Observers are never explicitly given the underlying rule for solving any given problem. Instead, they learn it while classifying positive and negative examples and receiving task feedback. Examples from two representative tasks are depicted in Figure~\ref{fig:example}: observers must learn to recognize whether two shapes are the same or different (Task \textit{1}) or whether or not the smaller of the two shapes are near the boundary (Task \textit{2}). Additional abstract relationships tested in the challenge include ``inside", ``in between”, ``forming a square”, ``aligned in a row" or ``finding symmetry" (see Figures \ref{fig:exampleSD} and \ref{fig:exampleSR} for examples).

Most SVRT tasks are rapidly learned by human observers within twenty or fewer training examples~\citep{fleuret2011comparing} (see Table~\ref{table:humans}; reproduced from the original study). On the other hand, modern deep neural network models require several orders of magnitude more training samples for some of the more challenging tasks~\citep{ellis2015unsupervised,kim2018not,MESSINA2021,stabinger2021evaluating,Stabinger2016,Puebla2021.04.06.438551} (see~\citet{ricci37same} for review; see also \citet{Funke2021} for an alternative perspective). 

It is now clear that some SVRT tasks are more difficult to learn than others. For instance, tasks that involve spatial-relation (SR) judgments can be learned much more easily by deep convolutional neural networks (CNNs) than tasks that involve same-different (SD) judgments~\citep{Stabinger2016, kim2018not,yihe2019program}. In contrast, a very recent study \citep{Puebla2021.04.06.438551} demonstrated that even when CNNs learn to detect whether objects are the same or different, they fail to generalize over small changes in appearance, meaning that they have only partially learned this abstract rule. The implication of the relative difficulty of learning SR versus SD tasks is that CNNs appear to need additional computations to solve SD tasks beyond standard filtering, non-linear rectification, and pooling. Indeed, recent human electrophysiology work~\citep{AlamiaENEURO.0267-20.2020} has shown that SD tasks recruit cortical mechanisms associated with attention and working memory processes to a greater extent than SR tasks. \textcolor{black}{Others have argued that SD tasks are central to human intelligence \citep{firestone2020performance, FORBUS202163, GENTNER202184}.} Beyond this basic dichotomy of SR and SD tasks, little is known about the neural computations necessary to learn to solve SVRT tasks as efficiently as human observers.

Here, we investigate the neural computations required for visual reasoning. In our experiment, we extend prior studies on the learnability of individual SVRT tasks by feedforward neural networks using a popular class of deep neural networks known as deep residual networks (``ResNets'') \citep{he2016deep}. We systematically analyze the ability of ResNets to learn all twenty-three SVRT tasks as a function of their expressiveness, parameterized by processing depth (number of layers), and their efficiency in learning a particular task. Through these experiments, we found that most of the performance variance in the space of SVRT tasks could be accounted for by two principal components, which reflected both the type of task (same-different vs. spatial-relation judgments) and the number of relations used to compose the underlying rules. 

\section{Systematic analysis of SVRT tasks' learnability}

All experiments were carried out with the \textit{Synthetic Visual Reasoning Test} (SVRT) dataset using code provided by the authors to generate images with dimension \textit{128 $\times$ 128} pixels (see \citet{fleuret2011comparing} for details). \textcolor{black}{ All images were normalized and resized to 256$\times$256 pixels for training and testing models. No image augmentations were used during training.}
In our first experiment, we wanted to measure how easy or difficult each task is for ResNets to learn. We did this by recording the SVRT performance of multiple ResNets, each with different numbers of layers and trained with different numbers of examples. By varying model complexity and the number of samples provided to a model to learn any given task, we obtained complementary measures of the learnability of every SVRT task for ResNet architectures. In total, we trained 18-, 50-, and 152-layer ResNets separately on each of the SVRT's twenty-three tasks. Each of these models was trained with .5k, 1k, 5k, 10k, 15k, and 120k class-balanced samples. We also generated two unique sets of 40k positive and negative samples for each task: one was used as a validation set to select a stopping criterion for training the networks (if validation accuracy reaches 100\%) and one was used as a test set to report model accuracy. In addition, we used three independent random initializations of the training weights for each configuration of architecture/task and selected the best model using the validation set. Models were trained for \textit{100} epochs using the $Adam$ optimizer \citep{kingma2014adam}  with a training schedule (we used an initial learning rate of 1$e$-3 and changing it to 1$e$-4  from the $70^{th}$ epoch \textcolor{black}{onward}). As a control, because these tasks are quite different from each other, we also tested two additional initial learning rates (\textit{1e-4, 1e-5}).

\begin{figure}[htbp]
\centering
  \includegraphics[width=.5\textwidth]{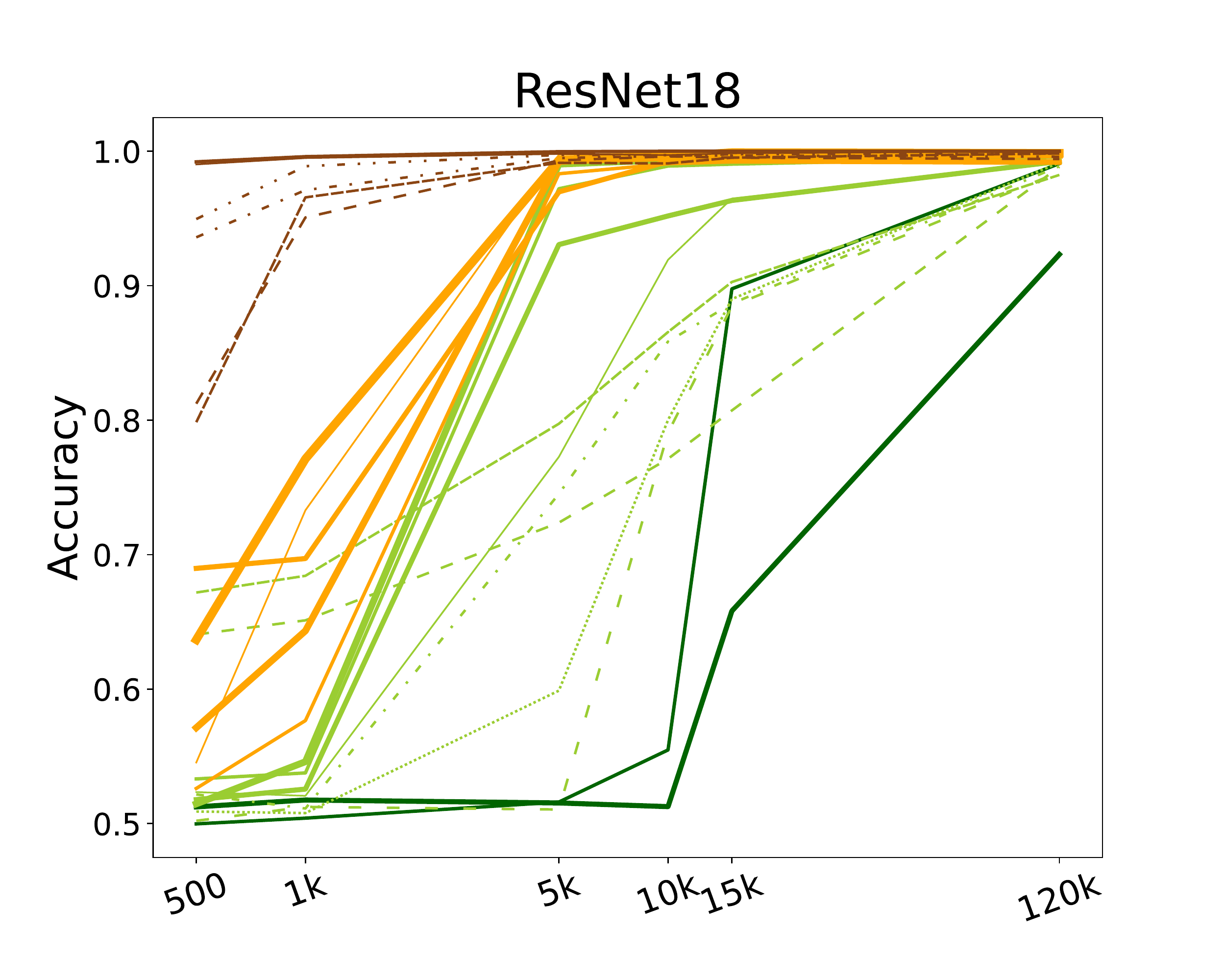} \includegraphics[width=.5\textwidth]{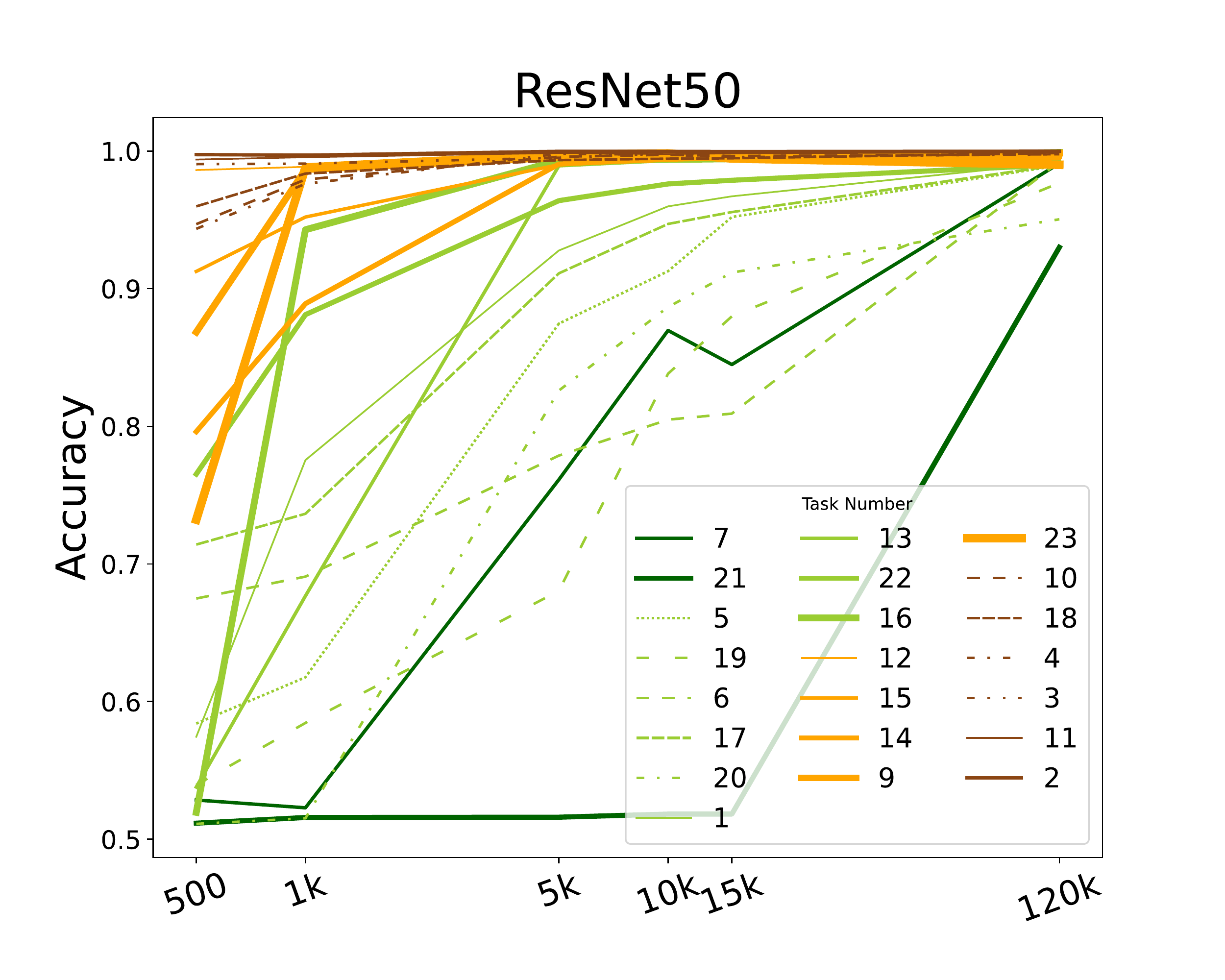} \\
  \includegraphics[width=.5\textwidth]{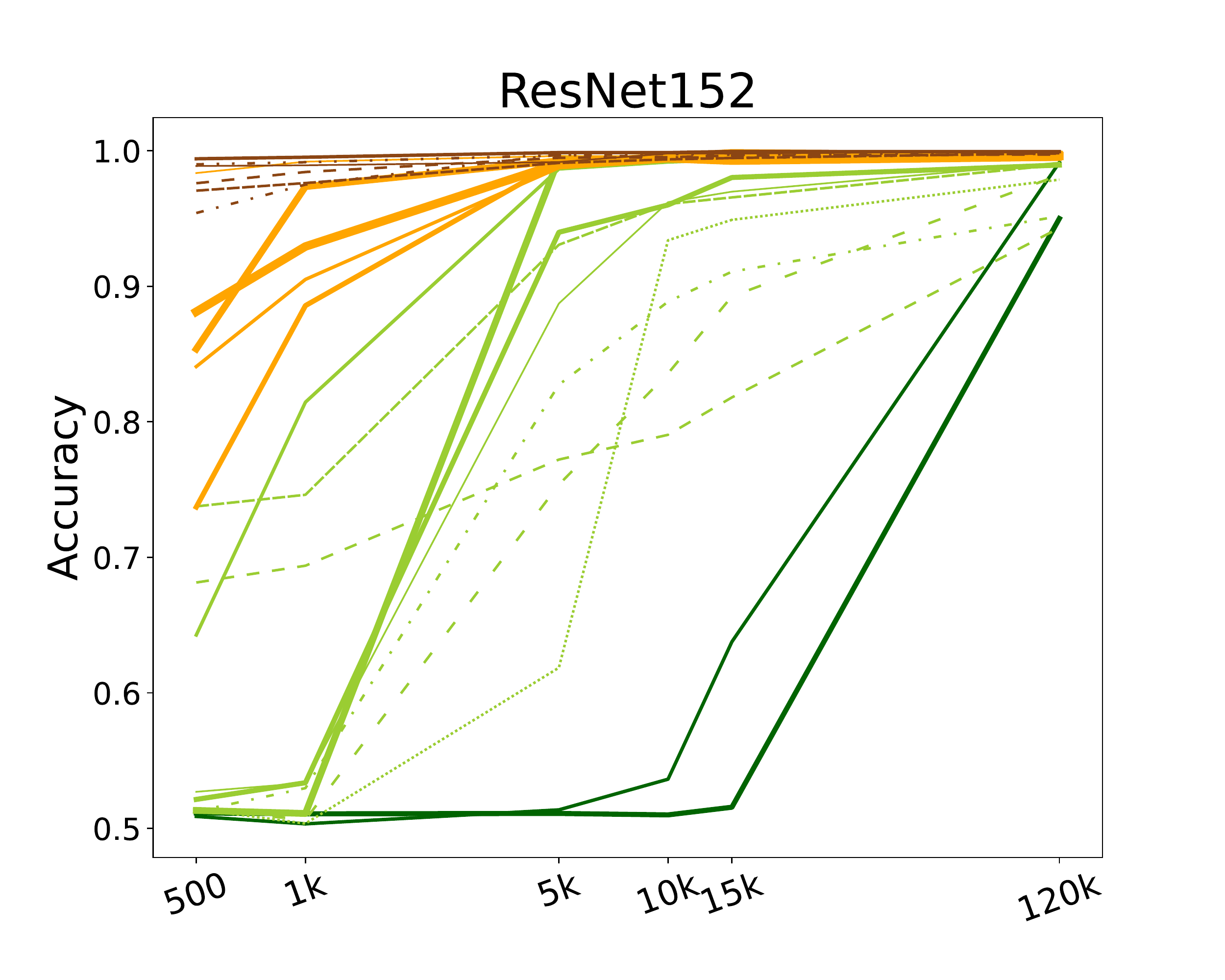}
 \caption{Test accuracy for each of the twenty-three SVRT tasks as a function of the number of training samples for ResNets with depths 18, 50 and 152, resp. The color scheme reflects the identified taxonomy of SVRT tasks (see Figure~\ref{fig:clustering} and text for details).}\label{fig:overall}
\end{figure}

Consistent with prior work \citep{kim2018not,Stabinger2016,yihe2019program}, we found that some SVRT tasks are much easier to learn than others for ResNets (Figure~\ref{fig:overall}). For instance, a ResNet50 needs only \textit{500} examples to perform well on tasks \textit{2, 3, 4, 8, 10, 11, 18} but the same network needs \textit{120k} samples to perform well on task \textit{21} (see Figures~\ref{fig:exampleSD} and \ref{fig:exampleSR} for examples of these tasks). Similarly, with \textit{500} training examples, task \textit{2, 3, 4 \& 11} can be learned well with only 18 layers while task \textit{9, 12, 15 \& 23} require as many as 152 layers. A key assumption of our work is that these differences in training set sizes and depth requirements between different SVRT tasks reflect different computational strategies that need to be discovered by the neural networks during training for different tasks. Our next goal is to characterize what these computational strategies are.

\section{An SVRT taxonomy}
\begin{figure}[t!]
\centering
  \includegraphics[width=.95\linewidth]{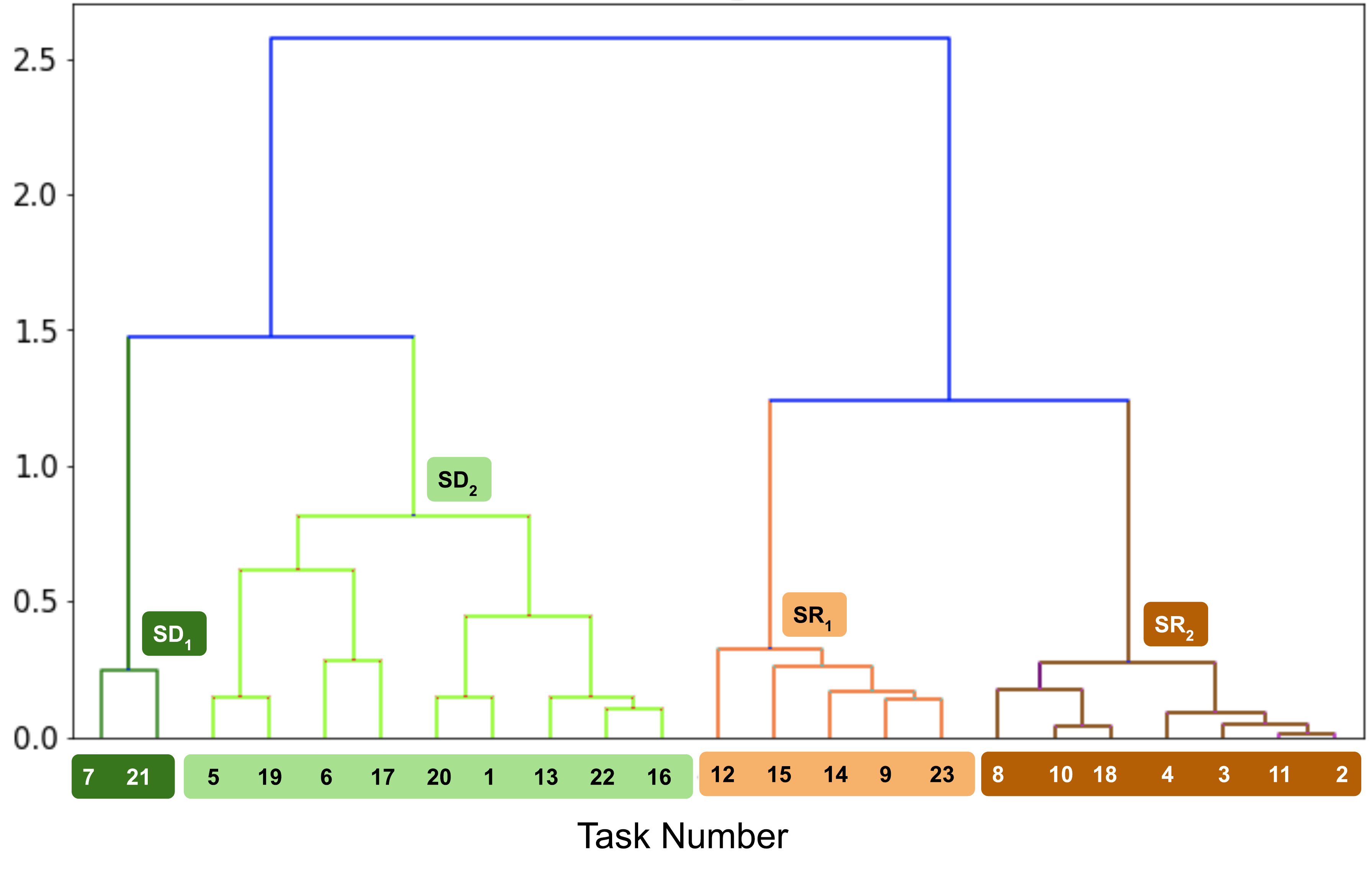}
\caption{Dendrogram derived from an N-dim hierarchical clustering analysis on the test accuracy of N=15 ResNets[18/50/152] trained to solve each task  over a range of training set sizes.} \label{fig:clustering}
\end{figure}

To better understand the computational strategies needed to solve the SVRT, we analyzed ResNet performance on the tasks with a multi-variate clustering analysis. For each individual task, we created an $N$-dimensional vector by concatenating the test accuracy of all ResNet architectures ($N = 3$ depths $\times$ 5 training set sizes = 15), which served as a signature of each task's computational requirements.  We then passed a matrix of these vectors to an agglomerative hierarchical clustering analysis (Figure~\ref{fig:clustering}) using the $Ward's$ method.

Our clustering analysis revealed a novel taxonomy for the SVRT.
At the coarsest level, it recapitulated the dichotomy between \textit{same-different} (SD; green branches) and \textit{spatial-relation} (SR; \textcolor{black}{brown} branches) categorization tasks originally identified by \citet{kim2018not} using shallow CNNs. Interestingly, two of the tasks which were classified as SR by \citet{kim2018not} (tasks \textit{6 \& 17}) were assigned to the SD cluster in our analysis. We examined the descriptions of these two tasks as given in~\citet{fleuret2011comparing} (see also Figures~\ref{fig:exampleSD} and~\ref{fig:exampleSR}) and found that these two tasks involve both SR and SD: they ask observers to tell whether shapes are the same or different and judge the distance between the shapes. Specifically, task \textit{6} involves two pairs of identical shapes with one category having the same distance in-between two identical shapes vs. not in the other. Similarly, in task \textit{17}, three of the four shapes are identical and their distance from the non-identical one is the same in one category vs. different in the other. Thus, our data-driven dichotomization of SR vs. SD refines the original proposal of \citet{kim2018not}. This could be due to our use of ResNets (as opposed to vanilla CNNs), deeper networks, and a greater variety of training set sizes (including much smaller training set sizes than those used by \citet{kim2018not}). \textcolor{black}{The analysis by \citet{fleuret2011comparing} also revealed that several SD tasks (\textit{6, 16, 17, 21}) are particularly challenging for human observers.}

Our clustering analysis also revealed a finer organization than the main SR vs. SD dichotomy. The SR cluster could be further subdivided into two sub-clusters. The $SR_2$ (dark-brown-coloured) branch in Figure~\ref{fig:clustering} captures tasks that involve relatively simple and basic relation rules such as shapes making close contact (\textit{3, 11}), or being close to one another (\textit{2}), one shape being inside the other (\textit{4}) or whether the shapes are arranged to form a symmetric pattern (\textit{8, 10, 18}). In contrast, tasks that fall in the $SR_1$ (light-brown-colored) branch involve the composition of more than two rules such as comparing the size of multiple shapes to identify a subgroup before identifying the relationship between the members of the sub-groups. This includes tasks such as finding a \textit{larger} object \textit{in between} two smaller ones (\textit{9}) or three shapes of which two are small and one large with two smaller (\textit{identification of large and small object}) ones either inside or outside in one category vs. one \textit{inside} and the other \textit{outside} in the second (\textit{23}), or \textit{two small} shapes \textit{equally close} to a bigger one (\textit{12}), etc. These tasks also tend to be comparatively harder to learn, requiring ResNets with greater processing depth and more training samples. For instance, tasks \textit{9, 12, 15, 23} were harder to learn than tasks \textit{2, 4, 11} requiring more samples and/or more depth to solve well (Figure~\ref{fig:overall}).

We found that task \textit{15} gets assigned to this latter sub-cluster because the task requires finding four shapes in an image that are identical vs. not. One would expect this task to fall in the SD cluster but we speculate that the deep networks are actually able to leverage a shortcut \citep{geirhos2020shortcut} by classifying the overall pattern as symmetric/square (when the four shapes are identical) vs. trapezoid (when the four shapes are different; see Figure~\ref{fig:exampleSR}) -- effectively turning an SD task into an SR task.

Our clustering analysis also reveals a further subdivision of the SD cluster. These tasks require recognizing shapes that are identical to at least one of the other shapes in the image. The first sub-cluster $SD_2$ (light green color branch) belongs to tasks that require identification of simple task rules, like answering whether or not two shapes are identical (even if it is along the perpendicular bisector) (tasks \textit{1, 20}; see Figure~\ref{fig:exampleSD}), determining if all the shapes on an image are the same (\textit{16, 22}), or detecting if two pairs of identical shapes can be translated to become identical to each other
(\textit{13}). Another set of tasks within this sub-cluster includes tasks that are defined by more complex rules that involve the composition of additional relational judgments. Sample tasks include identifying pairs/triplets of identical shapes and measuring the distance with the rest (\textit{6, 17}), determining if an image consists of pairs of identical shapes (\textit{5}), or detecting if one of the shapes is a scaled version of the other (\textit{19}). Finally, the second sub-cluster $SD_1$ shown in dark-green color involves two tasks that require an understanding of shape transformations. One task asks observers to say if one of the shapes is the scaled, translated, or rotated version of the other one (\textit{21}). The other task test asks observers to judge whether or not an image contains two pairs of three identical shapes or three pairs of two identical shapes in an image (\textit{7}).  

To summarize this first set of experiments, we have systematically evaluated the ability of ResNets spanning multiple depths to solve each of the twenty-three SVRT tasks for different training set sizes. This allowed us to represent SVRT tasks according to their learnability by ResNets of varying depth. By clustering these representations, we extracted a novel SVRT taxonomy that both recapitulated an already described SD-SR dichotomy \citep{kim2018not}, and also revealed a more granular task structure corresponding to the number of rules used to form each task. Tasks with more rules are harder for ResNets to learn. Our taxonomy also reveals an organization of tasks where easier $SR_1$ and $SR_2$ sub-clusters fall closer to each other than harder $SD_1$ and $SD_2$ sub-clusters.

\section{Conclusion}

The goal of the present study was to shed light on the computational mechanisms underlying visual reasoning using the Synthetic Visual Reasoning Test (SVRT)~\citep{fleuret2011comparing}. \textcolor{black}{There are} twenty-three binary classification problems in this challenge, which include a variety of same-different and spatial reasoning tasks.  

In our experiment, we systematically evaluated the ability of a battery of $N=15$ deep convolutional neural networks (ResNets) -- varying in depths and trained using different training set sizes -- to solve each of the SVRT problems. We found a range of accuracies across all twenty-three tasks. Shallower networks easily learned some tasks, and relatively small training sets and some tasks were hardly solved with much deeper networks and orders of magnitude more training examples.

Under the assumption that the computational complexity of individual tasks can be well characterized by the pattern of test accuracy across these $N=15$ neural networks, we formed N-dimensional accuracy vectors for each task and ran a hierarchical clustering algorithm. The resulting analysis \textcolor{black}{suggests} a taxonomy of visual reasoning tasks: beyond two primary clusters corresponding to same-different (SD) vs. spatial relation (SR) judgments, we also identified a finer organization with sub-clusters reflecting the nature and the number of relations used to compose the rules defining the task. Our results are consistent with previous work by \citet{kim2018not}, who first identified a dichotomy between SD and SR tasks. Our results also extend prior work \citep{fleuret2011comparing,kim2018not,yihe2019program} in \textcolor{black}{proposing} a finer-level taxonomy of visual reasoning tasks. The accuracy of neural networks is reflected in the number of relationships used to define the basic rules, which is expected, but it deserves closer examination. 

\citet{kim2018not} have previously suggested that SD tasks ``strain'' convolutional neural networks. That is, while it is possible to find a network architecture of sufficient depth (or the number of units) that can solve a version of the task up to a number of stimulus configurations (e.g., by forcing all stimuli to be contained within a $\Delta H \times \Delta W$ window), it is relatively easy to render the same task unlearnable by the same network \textcolor{black}{past} a certain number of stimulus configurations (e.g., by increasing the size of the window that contains all stimuli). It is as if these convolutional networks are \textcolor{black}{capable of learning} the task if the number of stimulus configurations remains below their memory capacity, and fails beyond that. \textcolor{black}{It remains an open question  whether non-convolution} alternatives to the CNNs tested here such as the now popular transformer networks  \citep{Dosovitskiy2020-iq,touvron2021training,tolstikhin2021mlp} would learn to solve some of the harder SVRT tasks more efficiently. As an initial experiment, we attempted to train and test a Vision Transformer \footnote[1]{\url{https://github.com/facebookresearch/dino}} (ViT) \citep{Dosovitskiy2020-iq}  constrained to have a similar number of parameters (21M) to the ResNet-50 used here. We were not able to get these architectures to do well on most of the tasks that are difficult for ResNets, even with 100k samples (also shown in \citet{messina2021recurrent}). It is worth noting that even 100k samples remain a relatively small dataset size by modern-day standards since ViT was trained from scratch.

Multi-layer perceptrons and convolutional neural networks including ResNets and other architectures can be formally shown to be universal approximators under certain architectural constraints. That is, they can learn arbitrary mappings between images to class labels. Depending on the complexity of the mapping, one might need an increasing number of hidden units to allow for enough expressiveness of the network; but provided enough units $/$ depth and a sufficient amount of  training examples, deep CNNs can learn arbitrary visual reasoning tasks. While we cannot make any strong claim for the specific ResNet architectures used in this study (currently, the proof is limited to a single layer without max pooling or batch normalization \citep{lin2018resnet}), we have indeed found empirically that all SVRT tasks could indeed be learned for networks of sufficient depth and provided a sufficient amount of training examples.
However, deep CNNs typically lack many of the human cognitive functions, such as attention and working memory. Such functions are likely to provide a critical advantage for a learner to solve some of these tasks \citep{10.7551/mitpress/1187.001.0001}. CNNs might have to rely instead on function approximation which could lead to a less general ``brute-force'' solution. 
Given this, an open question is whether the clustering of SVRT tasks derived from our CNN-based analyses will indeed hold for human studies. At the same time, the prediction by \citet{kim2018not} using CNNs that SD tasks are harder than SR tasks and hence that they may demand additional computations (through feedback processes) such as attention and/or working memory was successfully validated experimentally by \citet{AlamiaENEURO.0267-20.2020} using EEG. 

Additional evidence for the benefits of feedback mechanisms for visual reasoning was provided by  \citet{linsley2018learning} who showed that contour tracing tasks that can be solved efficiently with a single layer of a recurrent CNN may require several order of magnitudes more processing stages in a non-recurrent-CNN to solve the same task. This ultimately translates into much greater sample efficiency for recurrent-CNNs on natural image segmentation tasks \citep{linsley2020recurrent}. \textcolor{black}{The closely related task of ``insideness'' was also studied by \citet{villalobos2021neural} who demonstrated the inability of CNNs to learn a general solution for this class of problems.}
Universal approximators with minimal inductive biases such as multi-layer perceptrons, CNNs and other feedforward or \textcolor{black}{non}-attentive architectures can learn to solve visual reasoning tasks, but they might need a very large number of training examples to properly fit. Hence, beyond simply measuring the accuracy of very deep nets in high data regimes (such as when millions of training examples are available), systematically assessing the performance of neural nets of varying depths and for different training regimes may provide critical information about the complexity of different visual reasoning tasks.

\part*{Chapter 3}
\chapter{Role of self-attention in a computer vision architecture}
\label{chapter1b}

\startcontents[chapters]
\printmyminitoc

\section{Introduction}

Humans continue to outperform modern AI systems in their ability to flexibly parse and understand complex visual relations. Prior cognitive neuroscience work suggests that attention plays a key role in humans' visual reasoning ability. 

In the realm of artificial intelligence, attention mechanisms have become essential components of cutting-edge machine learning algorithms. Inspired by the principles observed in neuroscience, attention models in AI enable machines to focus on salient features or regions of interest within input data, allowing them to allocate computational resources effectively and improve performance on various tasks. By selectively attending to relevant information, AI systems can extract meaningful patterns, make informed decisions, and exhibit more human-like intelligence. The synergy between attention research in neuroscience and AI has led to significant advancements in both fields. Neuroscientists can validate their theories by testing their predictions on AI models, while AI researchers can leverage the findings from neuroscience to design more biologically plausible and efficient attention models. This bidirectional flow of knowledge and insights has the potential to revolutionize our understanding of attention, cognitive processes, and the development of intelligent systems. By bridging the gap between neuroscience and artificial intelligence, we can unlock new perspectives on attention, fostering a deeper understanding of how attention shapes our cognitive abilities and paving the way for more sophisticated and efficient intelligent systems. 

Attention, a fundamental cognitive process, plays a crucial role in shaping our perception, memory, and decision-making. It allows us to selectively focus on relevant information while filtering out distractions, enabling efficient and adaptive behavior in complex environments. By employing selective attention, we give priority to information that is behaviorally relevant while disregarding surrounding stimulation. In neuroscience, the study of attention has provided valuable insights into the workings of the human brain. Neuroscientists have identified distinct neural networks and mechanisms that govern attentional processes, shedding light on how the brain filters and processes sensory inputs, allocates cognitive resources, and guides behavior. Understanding the neural basis of attention has not only deepened our understanding of human cognition but has also provided inspiration for developing attention models in AI.

Attention can be consciously directed towards spatial or non-spatial properties, also known as feature-based attention. By selectively directing our attention, we possess the ability to intentionally focus on particular aspects of our environment. This may include directing our attention to a specific position in space (spatial attention) or highlighting a specific feature, such as a particular color (feature-based attention). When a location or feature is correctly indicated (valid cue; attentional focusing), it results in improved performance for the subsequent stimulus. Conversely, when the cue is incorrect (invalid cue), performance declines as it necessitates reorienting attention to the unexpected target stimulus \citep{alanallport1971parallel,posner1980orienting,posner1987selective}. The impact of cueing, as observed through the difference between valid and invalid trials, can be observed in the activity modulations of neurons in early visual areas \citep{moran1985selective,treue1996attentional,roelfsema1998object}. This effect is further supported by increased activity modulations in early visual areas as revealed by functional magnetic resonance imaging (fMRI) \citep{o1999fmri,ungerleider2000mechanisms}.

Numerous studies have provided evidence that both spatial and feature attention have a modulatory effect on neuronal responses, resulting in an improved signal-to-noise ratio during the encoding of the attended stimulus. Additionally, it has been reported that both types of attention lead to increased neuronal response magnitudes across multiple visual areas \citep{carrasco2011visual}. Moreover, both spatial and feature attention contribute to enhancing the representation of the attended stimulus by reducing neuronal response variability (often measured using the Fano factor) and pairwise noise correlation \citep{cohen2009attention,mitchell2009spatial}. 

While there are notable similarities between feature and spatial attention, several differences have been observed as well. One prominent distinction is that spatial attention is confined to a specific location within the retinotopic map \citep{womelsdorf2006dynamic}, whereas feature attention impacts processing across the entire visual field \citep{saenz2002global}. Furthermore, the temporal dynamics of sensory neuron modulation differ between the two types of attention \citep{hayden2005time}. In order to investigate the underlying neural mechanism that may account for both similarities and differences between spatial and feature attention, \citet{ni2019neuronal} conducted a study involving trained monkeys performing a direction-change detection task. During the task, neuronal activity was recorded from the medial temporal cortex (MT). By manipulating the direction of the Gabors and the attended location, the researchers created three distinct task variants aimed at measuring the neuronal modulation induced by normalization, spatial attention, and feature attention.

Attention plays a crucial role in addressing the binding problem, which involves integrating various features of a stimulus into a unified object representation. Extensive research in the cognitive and neuroscience fields underscores the significance of attention in this process. The Feature Integration Theory (FIT) proposed by \citet{treisman1980feature} emphasizes how attention facilitates the binding of features, enabling the formation of coherent object representations. \citet{desimone1995neural} work focuses on the neural mechanisms of selective visual attention, elucidating how attention contributes to feature binding and information integration within the visual system. \citet{reynolds2004attentional} explore attention's impact on visual processing, highlighting its role in feature binding and coordinating neural activity across different brain regions. \citet{treisman1998feature} delves into the intricate relationship between feature binding, attention, and object perception. Additionally, \citet{corbetta2002control} investigate the control of goal-directed and stimulus-driven attention, shedding light on their implications for solving the binding problem. Collectively, these studies underscore the indispensable role of attention in integrating features and effectively addressing the binding problem through both cognitive and neural processes.

Furthermore, other works contribute valuable insights into object vision and the temporal dynamics of attention during visual search tasks. \citet{tanaka1996inferotemporal} focuses on the inferotemporal cortex (IT) and its role in object vision, discussing the neural mechanisms involved in recognizing complex object features. \citet{riesenhuber1999cortical} address the ``binding problem'' by examining cortical models and proposing distributed representations and feature-based attention as potential solutions, challenging the notion that cortical models are limited by this problem. \citet{woodman2003serial} investigate the temporal dynamics of attention during visual search tasks and propose a two-stage model involving serial deployment of attention to select target locations, followed by parallel processing within those locations. Together, these works contribute to our understanding of object recognition, the integration of visual features, and the temporal aspects of attention during visual search tasks, complementing the broader literature on attention and the binding problem in cognitive and neuroscience research.

Addressing the binding problem has been a prominent focus in cognitive and neuroscientific research, with computational approaches providing valuable insights. \citet{hommel1998event} introduces the concept of event files, which propose the automatic integration of stimulus-response episodes in memory. This work presents evidence supporting the automatic binding of stimuli and responses, suggesting the creation of temporary associations to optimize processing efficiency. \citet{lisman2013theta} contribute to the field by discussing the theta-gamma neural code, highlighting the significance of synchronized theta and gamma oscillations in neural communication and information processing. Their review emphasizes the role of the theta phase as a temporal framework for precise encoding and integration of information in various cognitive processes. Additionally, \citet{verguts2017binding} presents a computational model known as "binding by random bursts" to elucidate cognitive control mechanisms. This model posits that cognitive control emerges from dynamic interactions between low-level sensory processing and high-level control processes, where random bursts of neural activity act as a binding mechanism, coordinating information flow and facilitating flexible cognitive control. Lastly, \citet{senoussi2022time} investigate time-based binding as a solution and limitation for flexible cognition. They propose that temporal associations between events are crucial for cognitive processing and the binding of information over time. The discussion explores how time-based binding can enhance cognitive flexibility but also introduces constraints in rapidly changing contexts. Together, these studies contribute to our understanding of the binding problem and offer computational models that shed light on cognitive control and the role of time-based binding in flexible cognition.

In the previous chapter, we discussed a benchmark used to evaluate the abilities of machines in solving visual reasoning tasks and compare them with humans. We did this by systematically assessing the ability of modern deep convolutional neural networks (CNNs) to learn to solve the synthetic visual reasoning test (SVRT) challenge, a collection of 23 visual reasoning problems. Our analysis revealed a novel taxonomy of visual reasoning tasks, which can be primarily explained by the type of relations (same-different (SD) versus spatial-relation (SR) judgments) and the number of relations used to compose the underlying rules.

Consistent with the speculated role of attention in solving the binding problem when reasoning about objects \citep{egly1994covert,roelfsema1998object}, prior work by \citet{kim2018not} has shown that combining CNNs with an oracle model of attention and feature binding (i.e., preprocessing images so that they are explicitly and readily organized into discrete object channels) renders SD tasks as easy to learn by CNNs as SR tasks. Here, we build on this work and introduce CNN extensions incorporating spatial or feature-based attention.  In the first set of experiments, we show that these attention networks learn difficult SVRT tasks with fewer training examples than their \textcolor{black}{non}-attentive (CNN) counterparts but that the different forms of attention help on different tasks.

This experiment raises the question: how do attention mechanisms help with learning different visual reasoning problems? There are at least two possible computational benefits: attention could improve model performance by simply increasing its capacity, or attention could help models learn the abstract rules governing object relationships more efficiently. To adjudicate between these two possibilities, we measured the sample efficiency of ResNets pre-trained on SVRT images so that they only had to learn the abstract rules for each SVRT task. We found that attention ResNets and ResNets pre-trained on the SVRT were similarly sample-efficient in learning new SVRT tasks, indicating that attention helps discover abstract rules instead of merely increasing model capacity.

\section{Experiment 1: Self-attention with ResNet50}
\label{sec:exp2}
We sought to identify computational mechanisms that could help ResNets learn the more challenging SVRT tasks revealed by our novel taxonomy. Attention has classically been implicated in visual reasoning in primates and humans \citep{egly1994covert,roelfsema1998object}. Attentional processes can be broadly divided into \textit{\textbf{spatial}} (e.g., attending to all features in a particular image location) vs. \textit{\textbf{feature-based}} (e.g., attending to a particular shape or color at all spatial positions) \citep{desimone1995neural}. The importance of attention for perceiving and reasoning about challenging visual stimuli has also been realized by the computer vision community. There are now a number of attention modules proposed to extend CNN's -- including spatial  (e.g., \citet{Sharma2015-ow,journals/corr/ChenWCGXN15, yang2016, journals/corr/XuS15a,journals/corr/RenZ16}), feature-based (e.g., \citet{stollenga2014deep, chen2017sca, hu2018squeeze}) and hybrid (e.g., \citet{linsley2018global,woo2018cbam}) approaches. Here, we adapt the increasingly popular Transformer architecture \citep{vaswani2017attention} to implement both forms of attention. These networks, which were originally developed for natural language processing, are now pushing the state of the art in computer vision \citep{Zhu2020-nc,carion2020end,Dosovitskiy2020-iq}. \textcolor{black}{Recent work \citep{Ding2020AttentionOL} has also shown the benefits of such architectures and especially attention mechanisms for solving higher-level reasoning problems}.

Transformers are neural network modules usually consisting of at least one ``self-attention'' module followed by a feedforward layer. Here, we introduced different versions of the self-attention module into ResNets to better understand the computational demands of each SVRT task. Transformers' self-attention is applied to and derived from the module's input. By reconfiguring standard Transformer self-attention, we developed versions capable of allocating either spatial or feature-based attention over the input. Specifically, we created these different forms of attention by reshaping the convolutional feature map input to a Transformer. For spatial attention, we reshaped the $ \mathcal{Z} \in \mathcal{R}^{H, W, C}$ (where $H$ is \textit{height}, $W$ is \textit{width} and $C$ is \textit{number of feature channels}) feature maps to $ \mathcal{Z} \in \mathcal{R}^{C, H * W}$ so that the Transformer's self-attention was allocated overall spatial locations. For feature-based attention, we reshaped the convolutional feature maps to $\mathcal{Z} \in \mathcal{R}^{H * W, C}$, enforcing attention to overall features instead of spatial locations. 

\paragraph{Spatial Attention Module (SAM)}

Our first attention module takes a features map $X \in \mathcal{R}^{d_C \times d_H \times d_W}$ as input, where $d_C$, $d_H$, and $d_W$ respectively  refer to the number of channels, height and width of the map, and outputs a features map $Y$ of the same dimensions.

We flatten the spatial dimensions to obtain $X' \in \mathcal{R}^{d_C \times d_N}$, where $d_N = d_H \times d_W$, and we apply the original multi-head self-attention module from \citet{vaswani2017attention} as follows.

We first apply independent linear mappings of the input $X'$ to obtain three feature maps of dimensions $\mathcal{R}^{d \times d_N}$ for each attention head from a total of $n_H$ heads. For the $i^{th}$ head, these maps are known as the query $Q_i$, the key $K_i$ and the value $V_i$, and are obtained such as:
\begin{align*}
Q_i = W^Q_i . X' \\
K_i = W^K_i . X' \\
V_i = W^V_i . X'
\end{align*}
The mappings are parametrized by three matrices $W^Q_i$, $W^K_i$ and $W^V_i$ of dimensions $\mathcal{R}^{d \times d_C}$ for each head. The symbol $.$ denotes a matrix multiplication.

Then, we apply the scaled dot-product attention \citep{vaswani2017attention} to obtain $n_H$ attention heads of dimensions $\mathcal{R}^{d \times d_N}$ such as:
\begin{align}
    H_i = SoftMax(\frac{Q_i . K_i^T}{\sqrt{d}}) V_i
\end{align}

After, we concatenate all attention heads along the first dimension and apply a linear mapping to obtain $Y' \in \mathcal{R}^{d_C \times d_N}$ such as:
\begin{align}
    Z = W^O . Concat(H_1, ..., H_{n_H}) 
\end{align}
The mapping is parametrized by the matrix $W^O \in \mathcal{R}^{d_C \times d}$.

As commonly done, we have a residual connection before applying a layer normalization \citep{ba2016layer} such as:
\begin{align}
    Y' = LayerNorm(Z + X')
\end{align}

Finally, we unflatten $Y'$ to obtain $Y \in \mathcal{R}^{d_C \times d_H \times d_W}$.

We obtain the best results with a representation space of 512 dimensions ($d=512$) and four attention heads ($n_H=4$).

\paragraph{Features-based Attention Module (FBAM)}

Our second attention module is simply obtained by transposing the channel dimension with the spatial dimensions before applying the same transformations.
In other words, we transpose the input $X'$ into $\mathcal{R}^{d_N \times d_C}$ and transpose the output $Y'$ back into $\mathcal{R}^{d_C \times d_N}$. While SAM models attention over the $d_H*d_W$ regions that compose the input features map, FBAM models attention over the $d_C$ features channels.

We obtain the best results with a representation space of 196 dimensions ($d=196$) and one attention head ($n_H=1$).

\begin{figure}[ht]
    \centering
    \includegraphics[width=.9\linewidth]{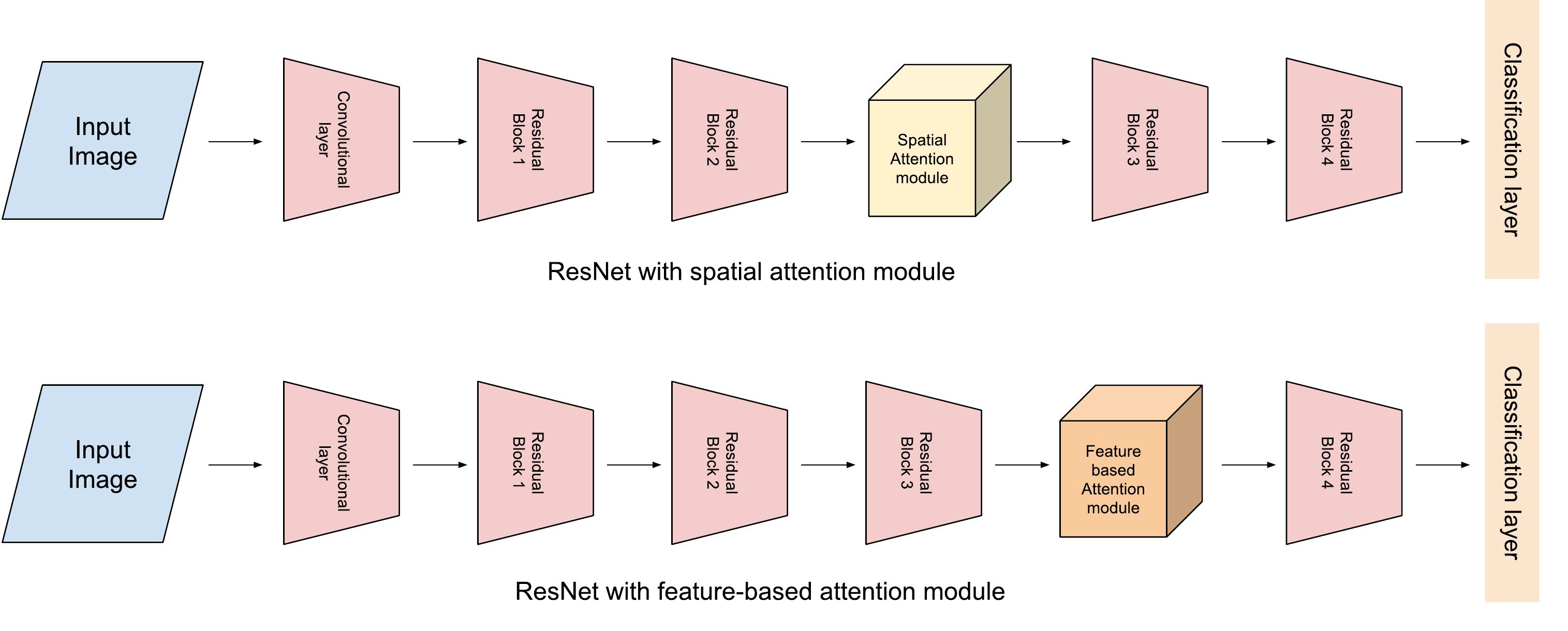} 
    \caption{\label{fig:architecture} Location of the Transformer self-attention modules in our ResNet extensions.}
    
\end{figure}

\textcolor{black}{ We added one spatial or feature-based attention after one of the four residual blocks in a ResNet-50. We placed either form of attention module to a ResNet-50 by choosing the location where the addition of attention yielded the best validation accuracy across the SVRT tasks. Through this procedure, we inserted a spatial attention module after the second residual block and a feature-based attention module after the third residual block (Figure~\ref{fig:architecture}).}

\begin{figure}[ht!]
\centering
  \begin{subfigure}{.8\textwidth}
    \centering
    \includegraphics[width=.9\linewidth]{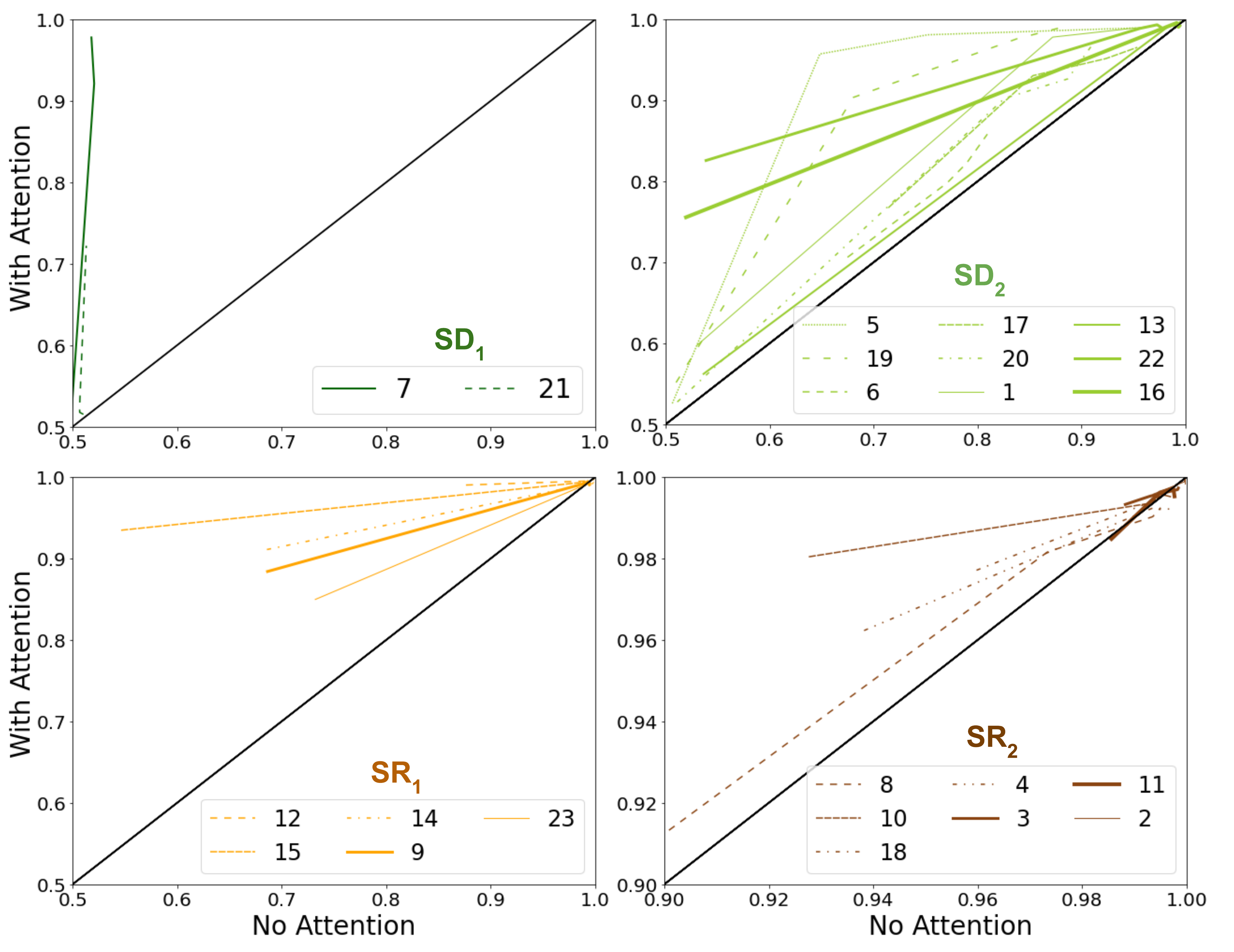} 
    \subcaption{Spatial attention}
  \end{subfigure}
  
  \begin{subfigure}{.8\textwidth}
    \centering
    \includegraphics[width=.9\linewidth]{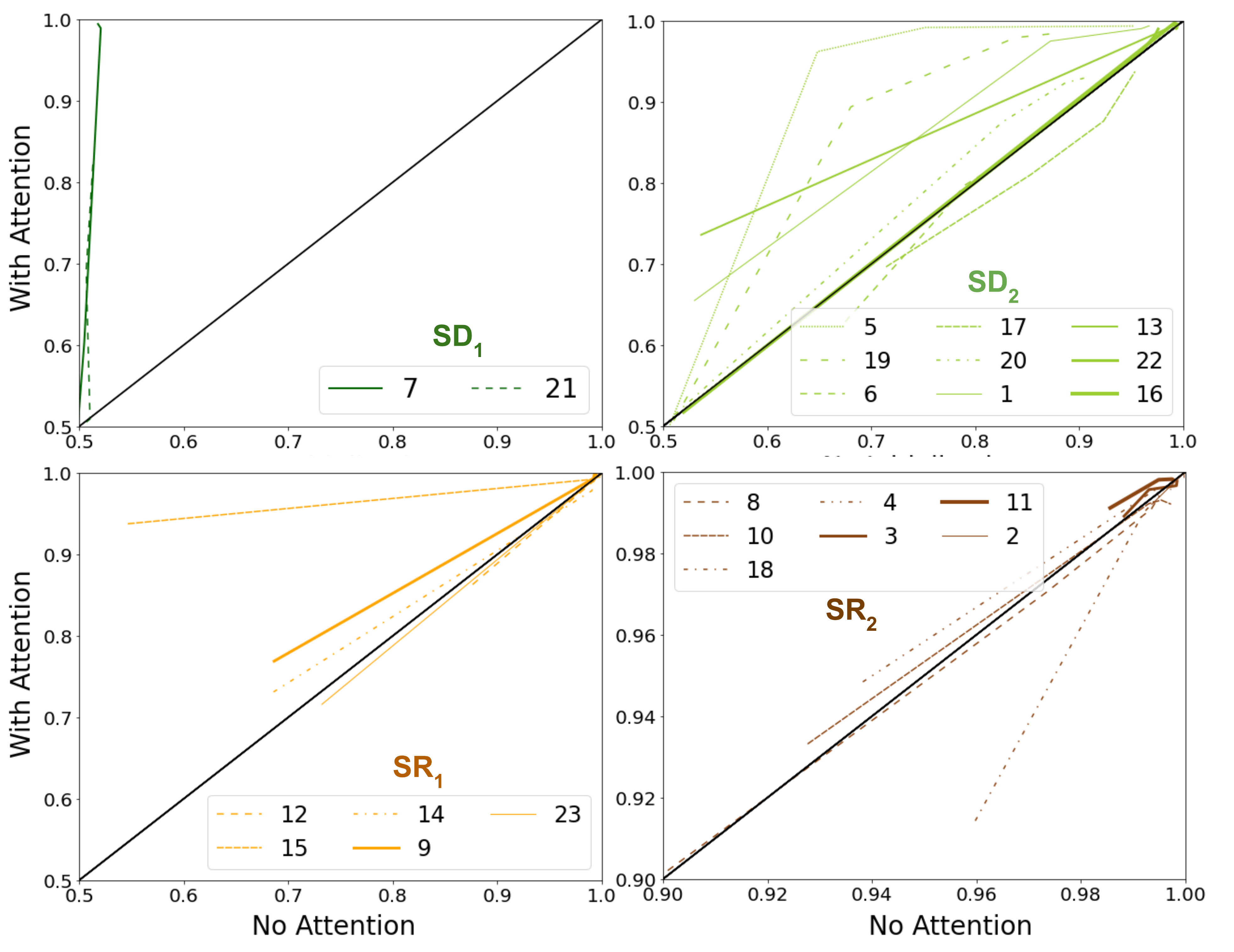}
  \subcaption{Feature-based attention}
 \end{subfigure}%
 \caption{Test accuracies for a baseline ResNet50 vs. the same architecture endowed with the two forms of attention for each of the twenty-three SVRT tasks when varying the number of training examples. A different axis scale is used for $SR_2$ to improve visibility. \textcolor{black}{These curves are constructed by joining task accuracy for five points representing dataset sizes.} }\label{fig:xy_sa}
\end{figure}

To measure the effectiveness of different forms of attention for solving the SVRT, we compared the accuracy of three ResNet-50 models: one capable of spatial attention, one capable of feature-based attention, and one that had no attention mechanisms (``vanilla'') (Figure~\ref{fig:xy_sa}). Spatial attention consistently improved model accuracy on all tasks across all five dataset sizes that models we used for training. The improvement in accuracy is particularly noticeable for the $SD_1$ cluster. Tasks in this sub-cluster are composed of two rules, which ResNets, without attention, struggled to learn. Attention helps ResNets learn these tasks more efficiently. The improvement is also evident for $SD_2$ and $SR_1$. However, the benefit of attention for $SR_2$ is marginal since ResNets without attention already perform well on these tasks.

We find that feature-based attention leads to the largest improvements for $SD_1$, especially when training on 5k or 10k examples 

\begin{figure}[t]
\centering
 \includegraphics[width=1\linewidth]{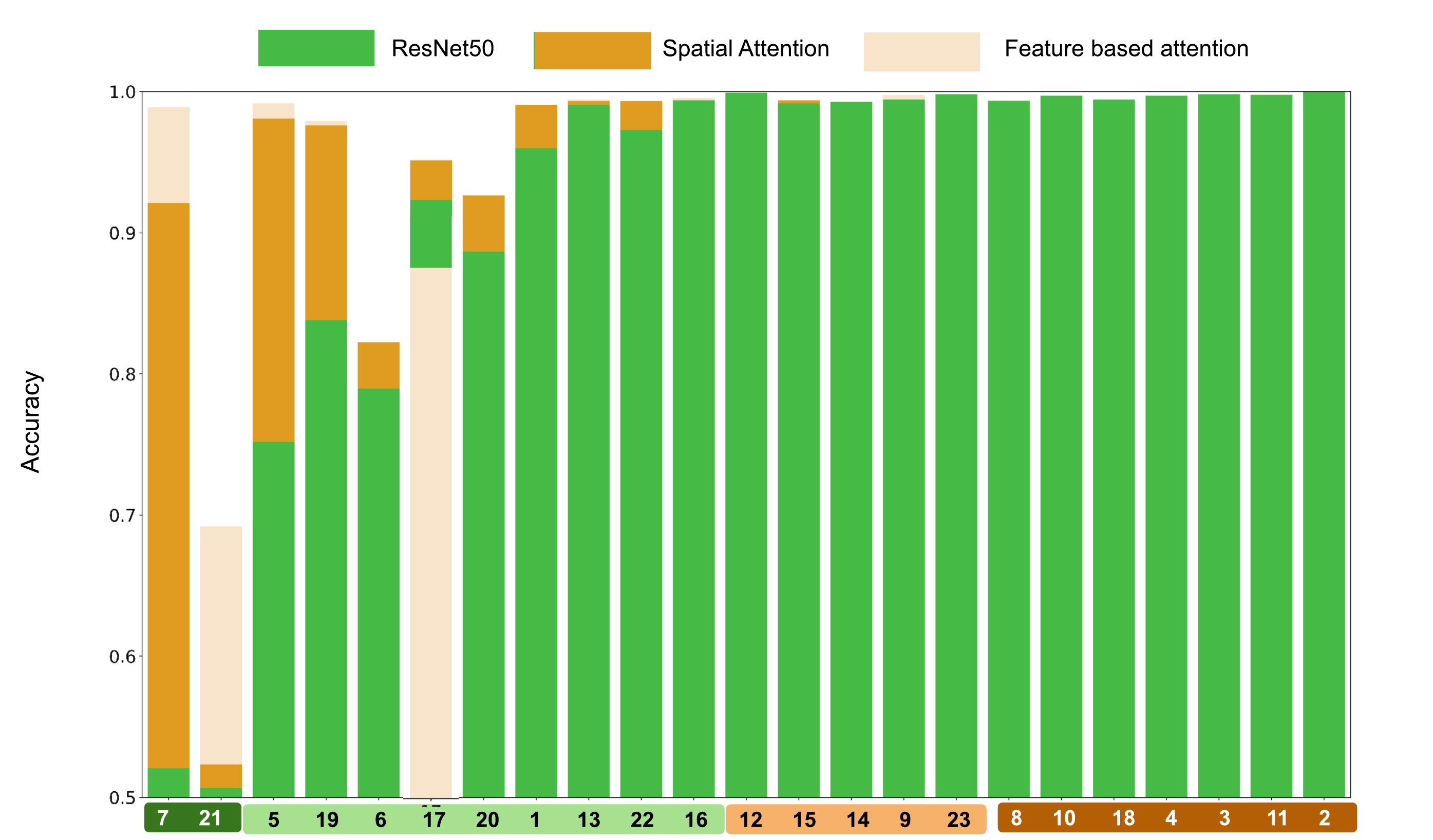}
 \caption{Test accuracies for 50-layer ResNets with spatial attention (orange), feature-based attention (tan), or no attention (green). Each bar depicts performance after training from scratch on 10k samples. }
 \label{fig:fbsa}
\end{figure}

(Figure~\ref{fig:fbsa}). On the other hand, spatial attention leads to the largest improvements for $SD_2$ and $SR_1$. This improvement is pronounced when training on 500 or 1000 examples. Taken together, the differential success of spatial versus feature-based attention reveals that their varying attentional demands can explain the task sub-clusters discovered in our data-driven taxonomy.

\clearpage

\begin{figure*}[ht!]
\centering
  \includegraphics[width=1\linewidth]{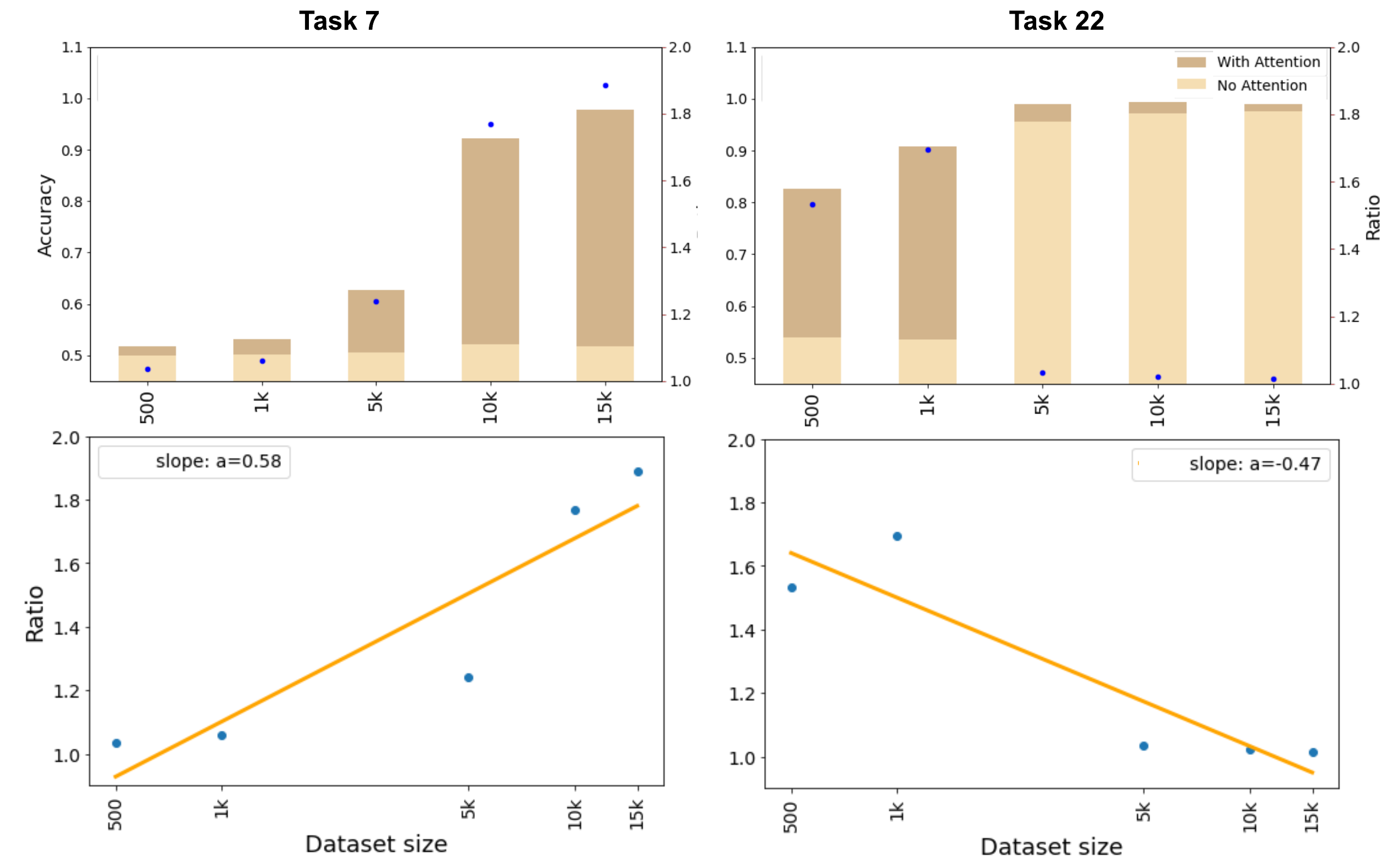}
 \caption{The benefit of attention in solving the SVRT is greatest in data-limited training regimes. The x-axis depicts the number of samples for training, and the y-axis depicts a ratio of the average performance of models with attention to models without attention. When the ratio is greater than 1, it shows that attention helps vs. hurts when lower than 1. This gives us five ratios per task and attention process corresponding to each dataset size. We performed a linear fitting procedure for these points and calculated the corresponding slope. This slope characterizes the relative benefits of attention for that particular task as the number of training examples available increases. If the benefit of attention is most evident in lower training regimes, one would expect a relatively small slope. If the benefit of attention is most evident in higher training regimes, one would expect a large slope.
 }\label{fig:slope}
\end{figure*}

To better understand how the ResNet-derived taxonomy found in Experiment 1 can be explained by the need for spatial and feature-based attention, we measured the relative improvement of each form of attention over the vanilla ResNet. For each attention model and task, we calculated the ratio of the test accuracies between the model and the vanilla ResNet50. We repeated this for every training dataset size, then fit a linear model to these ratios to calculate the slope across dataset sizes (see Figure~\ref{fig:slope} for representative examples). We then repeated this procedure for all twenty-three tasks to produce two 23-dimensional vectors containing slopes for each model and every task.

\begin{figure*}[t]
\centering
  \includegraphics[width=.8\linewidth]{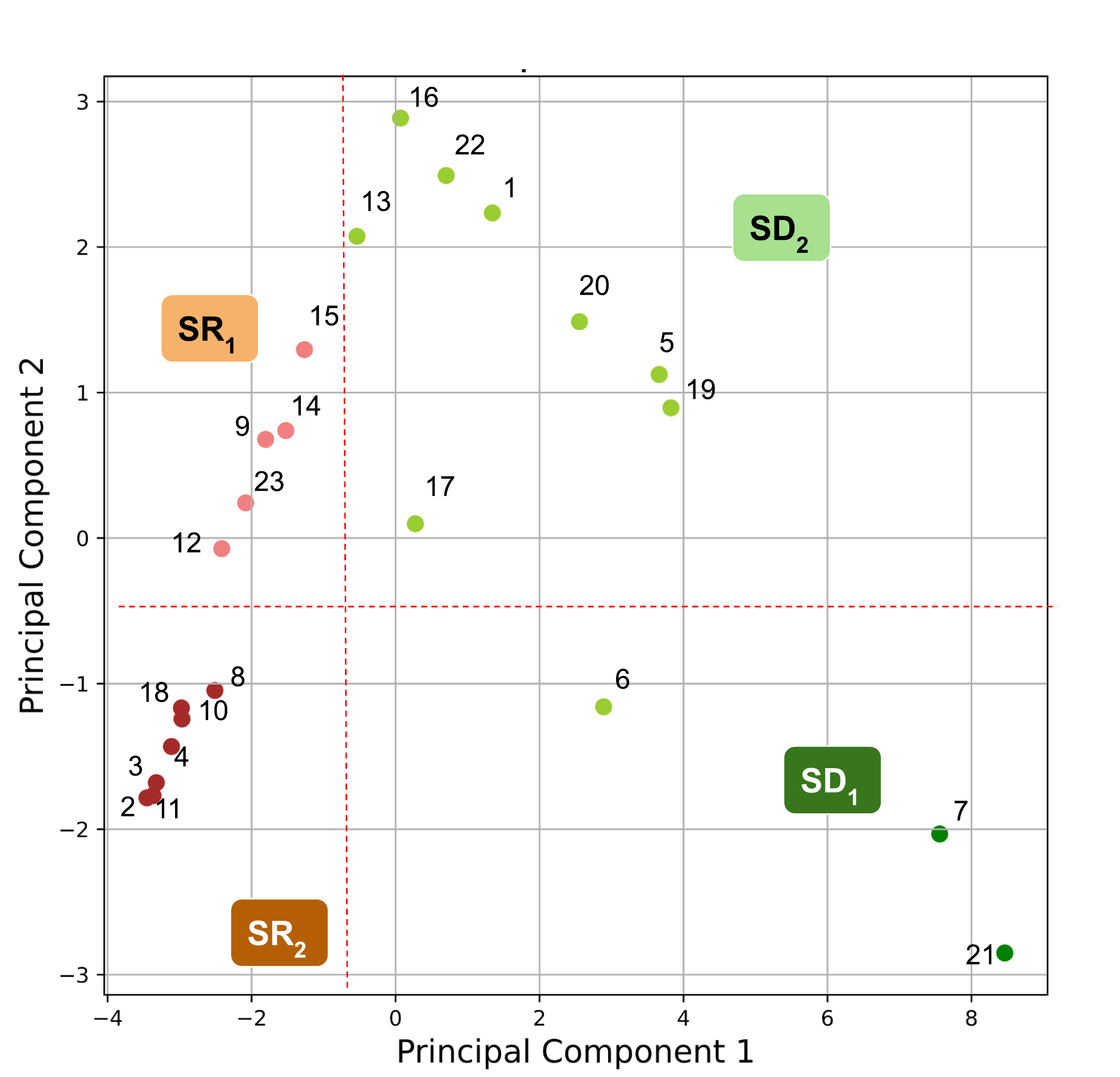}
 \caption{Principal component analysis of the twenty-three tasks using the 15-dimensional feature vectors derived from Experiment 1 representing the test accuracy obtained for each task for different dataset sizes and ResNets of varying depths (18, 50 \& 152). The dotted red line represents 4 different bins in which these tasks can be clustered.} \label{fig:pca}
\end{figure*}

We next used these slopes to understand the attentional demands of each SVRT task. We did this through a two-step procedure. First, we applied a principal component analysis (see Figure~\ref{fig:pca}) to the vanilla ResNet performance feature vectors ($N=15$) derived from Experiment 1. Second, we correlated the principal components with the slope vectors from the two attention models. We restricted our analysis to the first two principal components, which captured  $\sim 93\%$ of the variance in the vanilla ResNet's performance (Figure~\ref{fig:pca}). This analysis revealed a dissociation between the two forms of attention: feature-based attention was most correlated with the first principal component, and spatial attention with the second principal component. Additionally, along the first principal component, we found the broader dichotomy of these 23 tasks into $SD$ and $SR$ clusters, whereas the second principal component divulges the tasks which responded better with spatial attention from tasks requiring either no attention or feature-based attention (as seen in dotted red line along both the axis in Figure~\ref{fig:pca}).  The corresponding Pearson coefficient $r$ and $p$ values are given in Table~\ref{tab:rp}. 

\begin{table}[ht]
\centering
\begin{center}
    \caption{Pearson coefficient ($r$) and corresponding $p$ values obtained by correlating the slope vectors of the spatial attention and the feature-based attention modules with the two principal components of Figure~\ref{fig:pca}. See text for details.}
    \label{tab:rp}

\end{center}

\begin{tabular}{lcccc} 
\hline 
   & \multicolumn{2}{c}{$Spatial$} & \multicolumn{2}{c}{$Feature$} \\ \hline 
   & \textbf{r }      & \textbf{p }     & \textbf{r }         & \textbf{p }        \\  \hline \hline 
$PC_{1}$ & 0.466                 & 0.0249                & \textbf{0.649}            & 0.0008                  \\ \hline
$PC_{2}$ & \textbf{-0.652}       & 0.0007               & -0.491                   & 0.0174 \\ \hline
\end{tabular}

\end{table}

To summarize our results from Experiment 2,  we have found that the task clusters derived from ResNet test accuracies computed over a range of depth and training set sizes can be explained in terms of attentional demands. Here, we have shown that endowing these networks with attentional mechanisms helps them learn some of the most challenging problems with far fewer training examples. We also found that the relative improvements obtained over standard ResNets with feature-based and spatial attention are consistent with the taxonomy of visual reasoning tasks found in Experiment 1. More generally, our analysis shows how the relative need for feature vs. spatial attention seems to account for a large fraction of the variance in computational demand required for these SVRT tasks defined in Experiment 1 according to their learnability by ResNets. 

\section{Experiment 2: Feature vs. rule learning}

The learnability of individual SVRT tasks reflects two components: the complexity of the task's visual features and, separately, the complexity of the rule needed to solve the task. To what extent are our estimates of learnability driven by either of these components? We tested this question by training a new set of ResNets without attention according to the procedure laid out in Experiment 1, but with different pre-training strategies. One of the ResNets was pre-trained to learn visual statistics (but not rules) of SVRT images, and another was pre-trained on ImageNet, \citep[a popular computer vision dataset containing natural object categories;][]{deng2009imagenet}.

For pre-training on SVRT, we sampled 5,000 class-balanced images from each of the 23 tasks (5,000 $\times$ 23 = 115,000 samples in total). To ensure the networks did not learn any of the SVRT task rules, we shuffled images and binary class labels across all twenty-three problems while pre-training the network. We then trained models with binary cross-entropy to detect positive examples \textit{without discriminating tasks}. \textcolor{black}{Our assumption is that shuffling images and labels removes any semantic information between individual images and SVRT rules. However, a network with sufficient capacity can still learn the corresponding mapping between arbitrary images and class labels (even though it cannot generalize it to novel samples). To learn this arbitrary mapping, the network has to be able to encode visual features; but by construction, it cannot learn the SVRT task rule.} When training this model and the ImageNet-initialized model to solve individual SVRT tasks, we froze the weights of the convolutional layers and only fine-tuned the classification layers to solve SVRT problems.

\begin{figure*}[htbp]
\centering
    \includegraphics[width=1\linewidth]{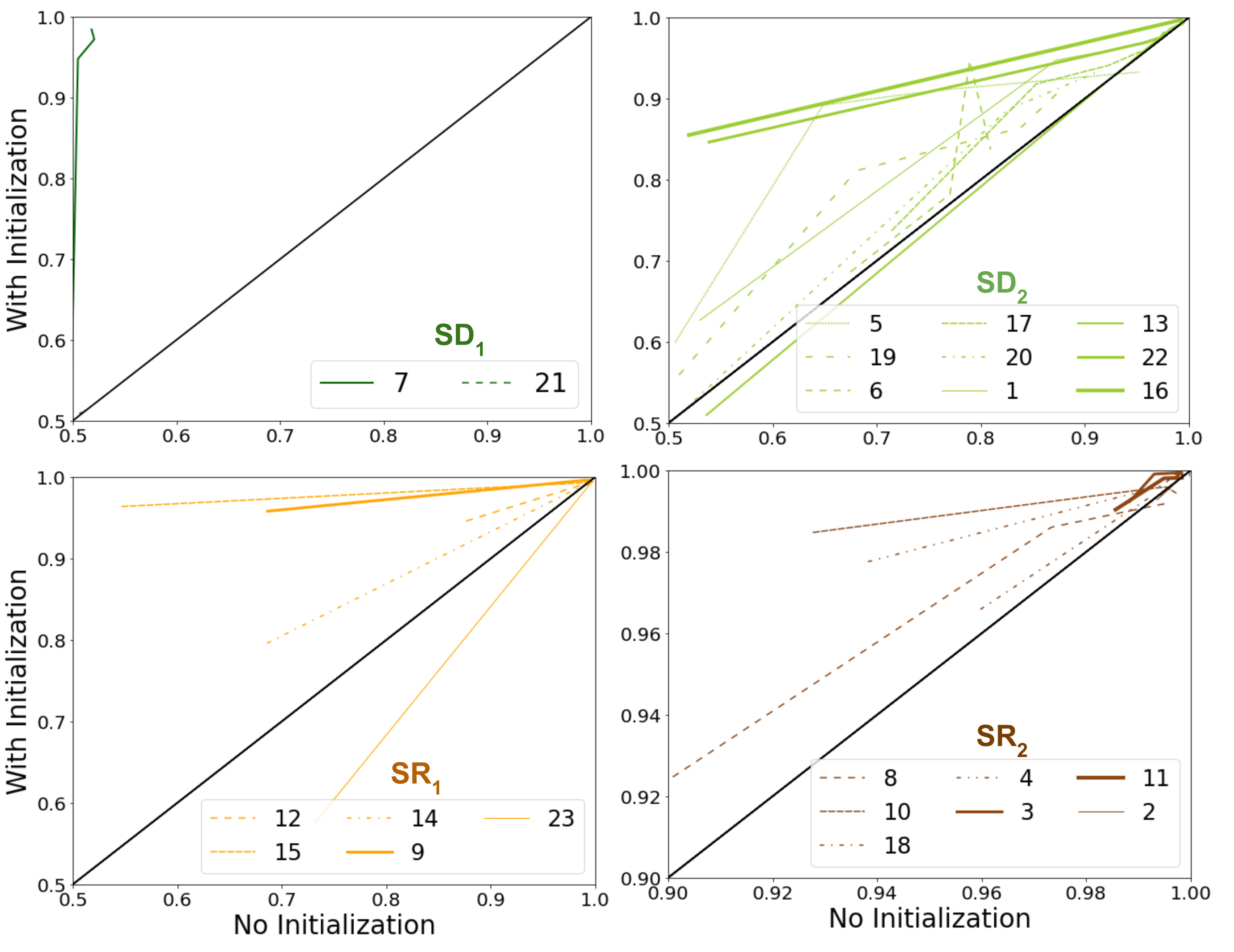}
 \caption{Test accuracies for a baseline ResNet50 trained from scratch (``No initialization'') vs. the same architecture pre-trained on an auxiliary task in order to learn visual representations that are already adapted to the SVRT stimuli for different \textcolor{black}{numbers} of training examples. The format is the same as used in Figure \ref{fig:xy_sa}. A different axis scale is used for $SR_2$ to improve visibility. \textcolor{black}{These curves are constructed by joining task accuracy for five points representing dataset sizes.}
 }\label{fig:xy_c_f}
\end{figure*}

Figure~\ref{fig:xy_c_f} shows a comparison between the different architectures in terms of their test accuracies according to the sub-clusters discovered in Experiment 1. These results first confirm that the SVRT pre-training approach works because it consistently outperforms pre-training on ImageNet (Figure~\ref{fig:xy_img}) or training from scratch. Interestingly, for the $SR_{2}$ sub-cluster, we found that the benefits of pre-training on SVRT go down very quickly as the number of training examples grows. We interpret these results as reflecting the fact that generic visual features are sufficient for the task and that the rule can be learned very quickly (somewhere around 500 and 5,000 samples). For $SR_{1}$ sub-cluster, the benefits of starting from features learned from SVRT are somewhat more \textcolor{black}{evident} in low training regimes. Still, these advantages quickly vanish as more training examples are available (the task is learned by all architectures within 5,000 training samples). 

For $SD_{1}$ while there appears to be a \textcolor{black}{noteworthy} advantage of pre-training on SVRT over ImageNet pre-training and training from scratch, the tasks never appear to be fully learned by any of the networks even with 15,000 training examples. This demonstrates the challenge of learning the rules associated with this sub-cluster beyond simply learning good visual representations. Finally, our results also show that the performance gap across all the architectures for $SD_{2}$ vs. $SD_{1}$ increases rapidly with more training examples -- demonstrating the fact that the abstract rule for $SD_{2}$ tasks are more rapidly learned than for $SD_{1}$.

\textcolor{black}{Finally}, we carried out a similar analysis with the pre-trained network as done in Experiment 2: We built test accuracy vectors for the SVRT pre-trained network trained using all five dataset sizes (.5k, 1k, 5k, 10k, 15k) and searching over a range of optimal learning rates (\textit{1e-4, 1e-5, 1e-6}). This led to a five-dimensional vector, which we  normalized by dividing each entry with the corresponding test accuracy of a baseline ResNet50 trained from scratch. Hence, the normalized vector represents the improvement (ratio larger than 1) or reduction in accuracy (ratio smaller than 1) that results from the pre-training on SVRT for that particular task and training set size. We then calculated the slope vector in $\mathcal R^{(23)}$, which we correlated with the corresponding spatial and feature-based attention vectors from Experiment 2.

\textcolor{black}{We found that task improvements due to SVRT pre-training correlated more strongly with task improvements due to spatial ($r=0.90$, $p=4e-9$) than feature-based attention ($r=0.595$, $p=0.002$)}. This suggests that the observed improvements in accuracy derived from spatial attention are more consistent with learning better feature representations compared to feature-based attention. 

To summarize, in Experiment 3, we have tried to address the question of the learnability of SVRT features vs. rules. We found that  using an auxiliary task to pre-train the networks on the SVRT stimuli in order to learn visual representations beforehand provides learning advantages to the network compared to a network trained from scratch. 

We also found a noteworthy correlation between the test accuracy vector of a network pre-trained on SVRT visual statistics and a similar network endowed with spatial attention. This suggests that spatial attention helps discover the abstract rule more so that it helps improve learning good visual representations for the task. 

\section{Conclusion}

Earlier, \citet{kim2018not} hypothesized that such straining by convolutional networks is due to their lack of attention mechanisms to allow the explicit binding of image regions to mental objects. A similar point was made by \citet{greff2020binding} in the context of the contemporary neural network failure to carve out sensory information into discrete chunks which can then be individually analyzed and compared (see also \citet{10.1007/978-3-540-75555-5_15} for a similar point). Interestingly, this prediction was recently tested using human EEG by \citet{AlamiaENEURO.0267-20.2020} who showed that indeed the brain activity recorded during SD tasks is compatible with greater attention and working memory demands than SR tasks. At the same time, that CNNs can learn SR tasks more efficiently than SD tasks does not necessarily mean that human participants can solve these tasks without attention. Indeed, \citep{logan1994spatial} has shown that SR tasks such as judging insideness require attention under some circumstances.

To assess the role of attention in visual reasoning, we used Transformer modules to endow deep CNNs with spatial and feature-based attention. The relative improvements obtained by the CNNs with the two forms of attention varied across tasks. Many tasks reflected a larger improvement for spatial attention, and a smaller number benefited from feature-based attention. Further, we found that the patterns of relative improvements accounted for much of the variance in the space of SVRT tasks derived in Experiment 1. Overall, we found that the requirement for feature-based and spatial attention accounts well for the taxonomy of visual reasoning tasks identified in Experiment 1.  \textcolor{black}{Our} computational analysis also lead to testable predictions for human experiments by suggesting tasks that either benefit from spatial attention (task \textit{22}) or from feature-based attention (task \textit{21}), tasks that benefit from either form of attention (task \textit{19}), and tasks that do not benefit from attention (task \textit{2}).

Finally, our study has focused on the computational benefits of spatial and feature-based attention for visual reasoning. Future work should consider the role of other forms of attention, including object-based attention \citep{egly1994covert} for visual reasoning.

In our second experiment, we studied the learnability of SVRT features vs. rules. We did this by pre-training the neural networks on auxiliary tasks in order to learn SVRT features before training them to learn the abstract rules associated with individual SVRT problems. Our pre-training methods led to networks that learn to solve the SVRT problems better than networks trained from scratch as well as networks that were pre-trained to perform image categorization on the ImageNet dataset. We have also found that such attention processes seem to contribute more to rule learning than to feature learning. For $SR_1$ sub-cluster we find this type of pre-training to be advantageous in lower training regimes but the benefits rapidly fade away in higher training regimes. In contrast, this pre-training does not allow the tasks from the $SD_1$ sub-cluster to be learned even with 15k samples -- suggesting that the key challenge with these tasks is not to discover good visual representations but rather to discover the rule. This suggests the need for additional mechanisms beyond those implemented in ResNets. This is also consistent with the improvements observed for these tasks with the addition of attention mechanisms. 

In summary, our study compared the computational demands of different visual reasoning tasks. While our focus has been on understanding the computational benefits of attention  and feature learning mechanisms, it is clear that additional mechanisms will be required to fully solve all SVRT tasks. These mechanisms are likely to include working memory which is known to play a role in SD tasks \citep{AlamiaENEURO.0267-20.2020}. Overall, this work illustrates the potential benefits of incorporating brain-like mechanisms in modern neural networks and provides a path forward to achieving human-level visual reasoning. 

\part*{Chapter 4}
\chapter{Role of self-attention in a cognitive architecture}
\label{chapter2b}

\startcontents[chapters]
\printmyminitoc

\epigraph{\itshape  Intelligence is not only the ability to reason; it is also the ability to find relevant material in memory and to deploy attention when needed.}{-- Daniel Kahneman}

\section{Introduction}
\label{sec:introc2}

Abstract reasoning refers to our ability to analyze information and discover rules to solve arbitrary tasks, and it is  fundamental to general intelligence in human and non-human animals \citep{gentner1997structure,lovett2017modeling}. It is considered a critical component for the development of artificial intelligence (AI) systems and has rapidly started to gain attention. A growing body of literature suggests that current neural architectures exhibit significant limitations in their ability to solve relatively simple visual cognitive tasks in comparison to humans (see ~\citet{ricci37same} for review). 
Given the vast superiority of animals over state-of-the-art AI systems, it makes sense to turn to brain sciences to find inspiration to leverage brain-like mechanisms to improve the ability of  modern deep neural networks  to solve complex visual reasoning tasks. Indeed, a recent human EEG study has shown that attention and memory processes are needed to solve same-different visual reasoning tasks~\citep{alamia2021differential}. \textcolor{black}{This interplay between attention and memory is previously discussed in \citet{buehner2006cognitive,fougnie2008relationship,cochrane2019fluid} emphasizing that a model must learn to perform attention over the memory for reasoning.}

It is thus not surprising that deep neural networks which lack attention and/or memory system fail to robustly solve visual reasoning problems that involve such same-different judgments~\citep{kim2018not}. Recent computer vision works~\citep{messina2021recurrent} including our own work in \hyperref[chapter1b]{Chapter 3} have provided further computational evidence for the benefits of attention mechanisms in solving a variety of visual reasoning tasks. Interestingly, in both aforementioned studies, a Transformer module was used to implement a form of attention known as self-attention~\citep{cheng2016long,parikh-etal-2016-decomposable}. In such a static module, attention mechanisms are deployed in parallel across an entire visual scene. By contrast, modern cognitive theories of active vision postulate that the visual system explores the environment dynamically  via sequences of attention shifts to select and route task-relevant information to memory. 
Psychophysics experiments \citep{hayhoe2000vision} on overt visual attention have shown that eye movement patterns are driven according to task-dependent routines. 

Inspired by active vision theory, we describe a dynamic attention mechanism, which we call \textit{guided attention}. Our proposed Guided Attention Module for (visual) Reasoning (GAMR) learns to shift attention dynamically, in a task-dependent manner, based on queries internally generated by an LSTM executive controller. Through extensive experiments on the two visual reasoning challenges, the Synthetic Visual Reasoning Test (SVRT) by~\citet{fleuret2011comparing} and the Abstract Reasoning Task (ART) by~\citet{Webb2021EmergentST}, we demonstrate that our neural architecture is capable of learning complex compositions of relational rules in a data-efficient manner and performs better than other state-of-the-art neural architectures for visual reasoning. Using explainability methods, we further characterize the visual strategies leveraged by the model in order to solve representative reasoning tasks. We demonstrate that our model is compositional -- in that it is able to generalize to novel tasks efficiently and learn novel visual routines by re-composing previously learned elementary operations. It also exhibit zero shot generalization ability -by translating knowledge across the tasks sharing similar abstract rules without the need of re-training.

\section{Related Work}
\label{sec:relwork}

Multiple datasets have been used to assess the visual reasoning ability of neural networks. One of the first challenges included the SVRT. Recently introduced Raven style Progressive Matrix datasets, RAVEN \citep{zhang2019raven}, PGM \citet{barrett2018measuring}, focuses on learning nearly seven unique rules and choose one of the eight choices. However, it was found that the dataset was seriously flawed as it was later found that neural architectures could solve tasks by leveraging shortcuts \citep{hu2020hierarchical,spratley2020closer} which were later removed in I-RAVEN~\citep{hu2021stratified}. Prior work~\citep{kim2018not,vaishnav2021understanding,messina2021recurrent} on SVRT studies has focused on the role of attention in solving some of these more challenging tasks. In SVRT, tasks that involve same-different (SD) judgements appear to be significantly harder for neural networks to learn compared to those involving spatial relation judgement (SR) \citep{stabinger201625,Yihe2019ProgramSP,kim2018not} (see \citet{ricci37same} for review). Motivated by neuroscience principles, \citet{vaishnav2021understanding} studied how the addition of feature-based and spatial attention mechanisms differentially affects the learnability of the tasks. These authors found that SVRT tasks could be further taxonomized according to their differential demands for these two types of attention. In another attempt to leverage a Transformer architecture to incorporate attention mechanisms for visual reasoning,  \citet{messina2021recurrent} proposed a recurrent extension of the classic Vision Transformer block (R-ViT). Spatial attention and feedback connections helped the Transformer to learn visual relations better. The authors compared the accuracy of four same-different (SVRT) tasks (tasks \textit{1,5,20,21}) to demonstrate the efficacy of their model. 

With the introduction of Transformer architecture, attention mechanisms started gaining popularity in computer vision. They can either complement \citep{bello2019attention,vaishnav2021understanding,d2021convit} or completely replace existing CNN architectures \citep{ramachandran2019stand,pmlr-v139-touvron21a,Dosovitskiy2020-iq}. Augmenting the attention networks with the convolution architectures helps them explore the best of the both worlds and train relatively faster. In contrast, stand-alone attention architecture takes time to develop similar inductive biases similar to CNN. As initially introduced by \citet{vaswani2017attention}, Transformer uses a self-attention layer, followed by residual connection and layer normalization and a linear projection layer to compute the association between input tokens. We used a similar system where instead of using a self-attention module, in GAMR, feature-based attention vector (an internally generated query) is obtained via an LSTM to guide the attention module to the location essential for the task and we thus call the model as \textit{guided-attention}. 
Since there could be more than one location where the model will attend, we then implemented a memory bank. \textcolor{black}{ A more closely aligned model with the human visual system is proposed by \citet{mnih2014recurrent} -- Recurrent Attention Model (RAM). It learns a saccadic policy over visual images and is trained using reinforcement learning to learn policies.} The Mnih et al system constitutes an example of overt attention. Conversely, GAMR constitutes an example of a covert attention system and assumes a fixed acuity.

We took inspiration for the memory bank from ESBN~\citep{Webb2021EmergentST}, where mechanisms for variable binding and indirection were introduced in architecture for visual reasoning with the help of external memory. Variable binding is the ability to bind two representations, and indirection is the mechanism involved in retrieving one representation to refer to the other. 
\textcolor{black}{While ESBN was indeed a source of inspiration, we would like to emphasize that GAMR constitutes a substantial improvement over ESBN. First and foremost, ESBN lacks attention. It requires items/objects to be passed serially one by one and hence it cannot solve SVRT or any other multi-object visual reasoning problems. In a sense, the approach taken in ESBN  is to assume an idealized frontend that uses hard attention to perfectly parse a scene into individual objects and then serially pipe them through the architecture. This is where our work makes a substantial contribution by developing an attention front-end (which is soft and not hard) to sequentially attend to relevant features and route them into memory. We tested this template-matching behavior of the ESBN architecture by training it in the presence of Gaussian noise and spatial jittering. It led to a chance-level performance. }
Here, we build on this work and describe an end-to-end trainable model that learns to individuate task-relevant scenes and store their representations in memory to allow the judging of complex relations between objects. Finally, our relational mechanism is inspired by the work in \citet{santoro2017simple} that introduced a plug-and-play model for computing relations between object-like representations in a network. 

\section{Proposed approach}
\label{sec:model}

\begin{figure*}[ht]
\centering
  \includegraphics[width=1\linewidth]{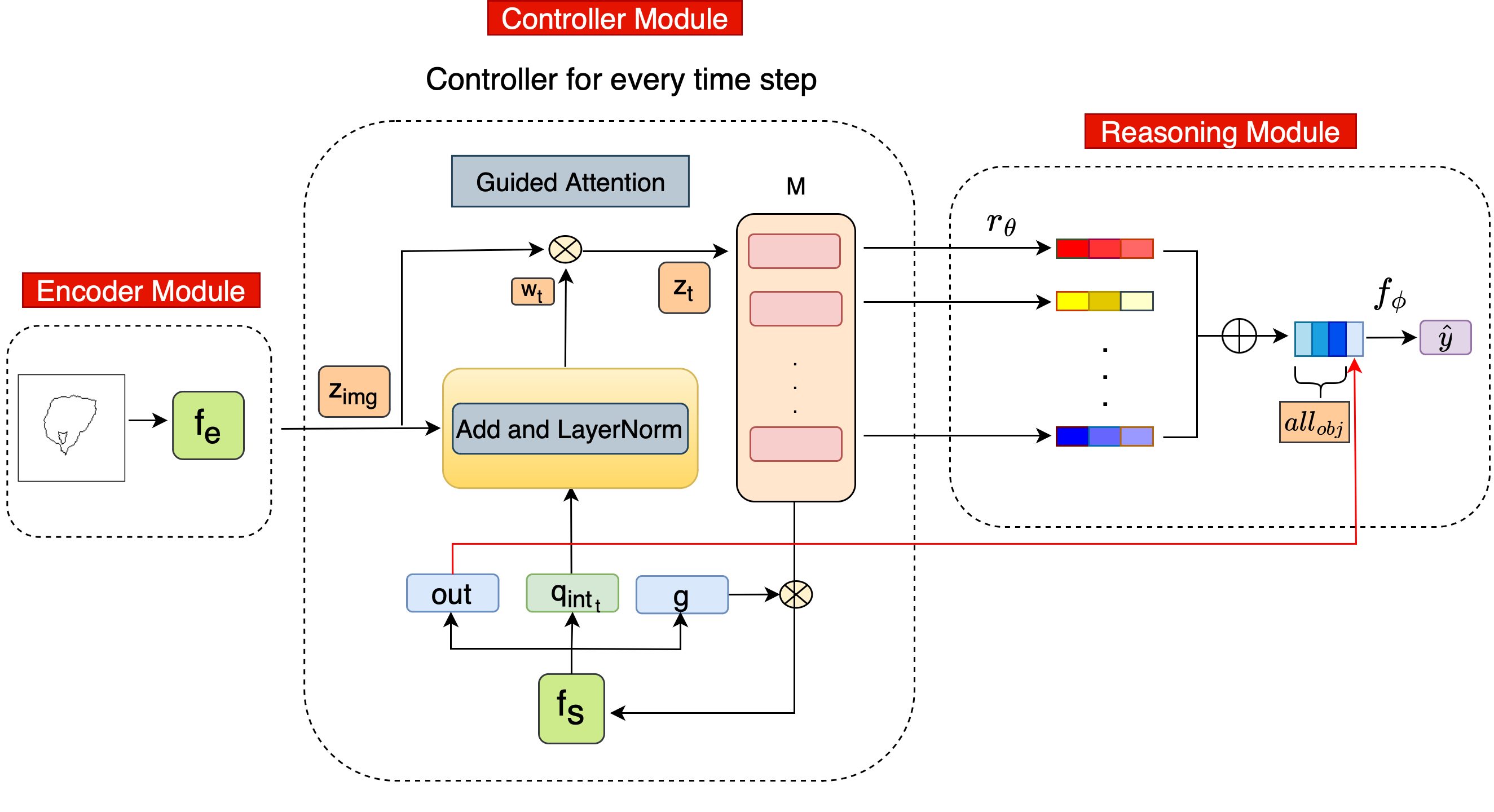}
  \caption{Our proposed \textit{GAMR} architecture is composed of three components: an \textit{encoder} module ($f_e$) builds a representation ($z_{img}$) of an image, a \textit{controller}  guides the attention module to dynamically shift attention, and selectively routes task-relevant object representations ($z_t$) to be stored in a memory bank ($M$). The recurrent controller ($f_s$) generates a query vector ($q_{int_t}$) at each time step to guide the next shift of attention based on the current fixation. After a few shifts of attention,  a \textit{reasoning} module ($r_{\theta}$) learns to identify the relationships between objects stored in memory.} 
  \label{fig:model}
\end{figure*}

\begin{figure}[htb]
\vskip 0.2in
\begin{center}
\includegraphics[width=1.\linewidth]{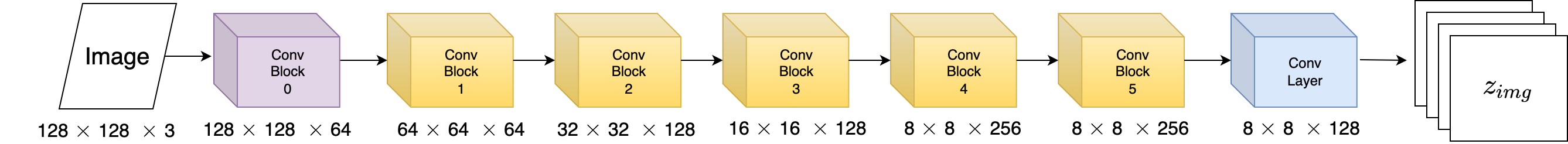} \\
\caption{Encoder module ($f_e$) used in \textit{GAMR}. It consists of four convolutional blocks to process input image of 128$\times$128 resolution}
\label{fig:encoder}
\end{center}
\vskip -0.2in
\end{figure}

Our model can be divided into three components: an encoder, a controller, and a relational module (see Fig.~\ref{fig:model} for an overview). In the \textbf{encoder module}, a low dimensional representation ($z_{img}$) for an input image ($x_{in}$) is created. It includes a feature extraction block ($f_e$) which is composed of five convolutional blocks (Figure~\ref{fig:encoder}). The output of the module is denoted as $z_{img}$ $\in \mathcal{R}^{(128,hw)}$ (with $h$ height and  $w$ width). We applied instance normalization (\textit{iNorm}) ~\citep{ulyanov2016instance} over $z_{img}$ before passing it to the controller for further processing without which the network even struggles to learn even simple relations.

\textbf{Controller module} is composed of two blocks: a recurrent neural network ($f_s$) which generates an internal query to guide the attention spotlight over the task relevant features. These features are selected with the help of the second block, i.e., guided attention (\textit{GA}) and are sent to the memory bank (\textit{M}).

After $z_{img}$ is built, a \textit{guided-attention} block is used to extracts the relevant visual information from the input image in a top-down manner at each time step ($t$). This block is responsible for generating context vector ($z_t$) to be stored in the memory bank (\textit{M}) along with all the previous context vectors. They are subsequently accessed again later by a reasoning module. This memory bank is inspired by the differentiable memory used in \citet{Webb2021EmergentST}.  

In the guided attention block, an attention vector ($w_t$ $\in \mathcal{R}^{128}$) is obtained by normalizing the addition of encoded feature ($z_{img}$) with internally  generated query ($q_{int_t}$) fetched by $f_s$. This normalized attention vector is used to re-weight the features at every spatial location of $z_{img}$ to generate the context vector $z_t$ $\in \mathcal{R}^{128}$. 
The recurrent controller ($f_s$) uses a Long Short-Term Memory (LSTM) to provide a query vector ($q_{int_t}$ $\in \mathcal{R}^{128}$) in response to a task-specific goal in order to guide attention for the current time step $t$. $f_s$ also independently generates a gate vector ($g$ $\in \mathcal{R}^{128}$) and output vector ($out$ $\in \mathcal{R}^{512}$) with the help of linear layers. 
The gate ($g$) is later used to shift attention to the next task-relevant feature based on the features previously stored in $M$. On the other hand, the decision layer uses the output vector ($out$) to produce the system classification output.

The \textbf{relational module} is where the reasoning takes place over the context vector ($z_t$) stored in the memory bank ($M$). This module is composed of a \textcolor{black}{two layered} MLP ($r_{\theta}$) which produces a relational vector ($all_{obj}$) \textcolor{black}{similar to the relational network~\citep{santoro2017simple}}. As we will show in section \ref{subsec:learningcomp}, $r_{\theta}$ learns elementary operations associated with basic relational judgments between context vectors ($z_t$) stored in the memory ($M$). It is concatenated with the output ($out$) of the controller ($f_s$) at the last time step ($t=T$) and passed through the decision layer ($f_{\phi}$) to predict the output ($\hat{y}$) for a particular task. We have summarized the steps in Algorithm~\ref{alg:cap}.

\begin{algorithm}[H]
    \caption{Guided Attention Model for (visual) Reasoning (\textit{GAMR}). $LN$ represents layer normalization \citep{ba2016layer} ($||$) indicates the concatenation of two vectors, forming a new vector. \{, \} indicates the concatenation of a matrix and a vector, forming a matrix with one additional row. $\odot$ represents element-wise multiplication and \{. \} represents the product between a scalar and a vector. (h,w) corresponds to the height and width of the feature map obtained from the encoder ($f_e$).}
    \label{alg:cap}

\begin{algorithmic}
    \State \begin{math}{k}_{r_{t=1}} \leftarrow{} 0 \end{math} \Comment{\textcolor{blue}{\begin{math}\in \mathcal{R}^{128}\end{math}}}
    
    \State \begin{math}{h}_{t=1} \leftarrow{} 0\end{math} 
    \Comment{\textcolor{blue}{\begin{math}\in \mathcal{R}^{512}\end{math}}}
    
    \State \begin{math}{M}_{t=1} \leftarrow{} \{\}\end{math}
    \State \begin{math}z_{img} \leftarrow{}
    f_e(x_{in})\end{math} 
    \Comment{\textcolor{blue}{\begin{math}\in \mathcal{R}^{(hw,128)}\end{math}}}
        \For{\texttt{t in 1...T}} 
            \State \begin{math} out,~ g,~ q_{int_t},~ h_t \leftarrow{} f_s(h_{t-1}, ~k_{r_{t-1}}) \end{math}
            \Comment{\textcolor{blue}{\begin{math}~~~~~~out\in \mathcal{R}^{512}, g\in \mathcal{R}^{128}, q_{int_t}\in \mathcal{R}^{128}\end{math}}}
            
            \State \begin{math} w_t \leftarrow{} LN (z_{img} + q_{int_t}.repeat(hw, axis=1))\end{math}
            \Comment{\textcolor{blue}{\begin{math}\in \mathcal{R}^{(hw,128)}\end{math}}}
            
            \State \begin{math} z_t \leftarrow{} (z_{img}~\odot~w_t.sum(axis=1)).sum(axis=1)\end{math}
            \Comment{\textcolor{blue}{\begin{math}\in \mathcal{R}^{128}\end{math}}}
            
            \If{t is 1}
                \State \begin{math} k_{r_{t}} \leftarrow{} 0 \end{math}
            \Else

                \State \begin{math} k_{r_{t}} \leftarrow{} g ~\odot~ M_{t-1}.sum(axis=1) \end{math}
                
            \EndIf
            \State \begin{math} M_{t} \leftarrow{} \{M_{t-1},~ z_{t}\} \end{math}
            \Comment{\textcolor{blue}{\begin{math}\in \mathcal{R}^{(t,128)}\end{math}}}
          \EndFor
    
    \State \begin{math} {all}_{obj} \leftarrow{} r_{{\theta}}(\sum_{i,j=1}^{T} (M_{v_i},~M_{v_j})) \end{math}
    
    \State \begin{math} \hat{y} \leftarrow{} f_{\phi}({all}_{obj}~||~out) \end{math}

\end{algorithmic}
\end{algorithm}

\section{Method}
\label{sec:exp}

\paragraph{Dataset}

The SVRT dataset is composed of \textit{twenty-three} different binary classification challenges, each representing either a single rule or a composition of multiple rules. A complete list of tasks with sample images from each category is shown in Appendix (Fig.~\ref{fig:exampleSD}, \ref{fig:exampleSR}). We formed four different datasets with 0.5k, 1k, 5k, and 10k training samples to train our model. We used unique sets of 4k and 40k samples for validation and test purposes. Classes are balanced for all the analyses. 

We trained the model from scratch for a maximum of 100 epochs with an early stopping criterion of 99\% accuracy on the validation set as followed in \citet{vaishnav2021understanding} using Adam \citep{kingma2014adam} optimizer and a binary cross-entropy loss. We used a hyperparameter optimization framework \textit{Optuna} \citep{optuna_2019} to get the best learning rates and weight decays for these tasks and reports the test accuracy for the models that gave the best validation scores.

\paragraph{Baselines} For the baselines in this dataset, we compared our architecture performance to a Relational Network ($RN$), a popular architecture for reasoning in VQA. The $RN$ uses the same CNN backbone as \textit{GAMR} with feature maps of dimension $\mathcal{R}^{128,hw}$ where $h=8$ and $w=8$. We consider each spatial location of the encoded feature representation as an object (i.e., $N=8\times 8=64$ object representations). We computed all pairwise combinations between all 64 representations using a shared MLP between all the possible pairs (totalling 4096 pairs). These combinations are then averaged and processed through another MLP to compute a relational feature vector ($all_{obj}$) before the final prediction layer ($f_{\phi}$). In a sense, \textit{GAMR} is a special case of an $RN$ network endowed with the ability to attend to a task-relevant subset ($N$ = 4) of these representations with the help of a controller instead of exhaustively computing all 4,096 possible relations -- thus reducing the computing and memory requirements of the architecture very significantly. 

As an additional baseline model, we used 50 layered \textit{ResNet} \citep{he2016deep} and its Transformer-based self-attention network (\textit{Attn-ResNet}) introduced in \citet{vaishnav2021understanding} \textcolor{black}{and follow the training mechanism as defined in the paper}. These have been previously evaluated on SVRT tasks \citep{Funke2021a,vaishnav2021understanding,messina2021solving,messina2021recurrent}. \textit{Attn-ResNet} serves as a powerful baseline because of more free parameters and a self-attention module to compare the proposed active attention component of \textit{GAMR}. In our proposed method, the controller shifts attention head sequentially to individual task-relevant features against a standard self-attention module -- where all task-relevant features are attended to simultaneously. We also evaluated memory based architecture, ESBN~\citep{Webb2021EmergentST} in which we used a similar encoder to that of \textit{GAMR} and pass the images in sequential order with each shape as a single stimulus and the number of time steps as the number of shapes present in the SVRT task. In order to train these models we used images of dimension $128 \times 128$ for architectures such as \textit{RN, ESBN, GAMR} and $256 \times 256$ for \textit{ResNet, Attn-ResNet} (to be consistent with configuration in \citet{vaishnav2021understanding}).  ResNet-50 ($ResNet$) has 23M parameters, Relation Network ($RN$) has 5.1M parameters, ResNet-50 with attention (\textit{Attn-ResNet}) has 24M parameters and \textit{GAMR} \& \textit{ESBN} both have 6.6M parameters. 
\section{Benchmarking the system}
\label{sec:result}

\begin{figure*}[ht]
\vspace{-.3cm}
\begin{center}
\includegraphics[width=.47\textwidth]{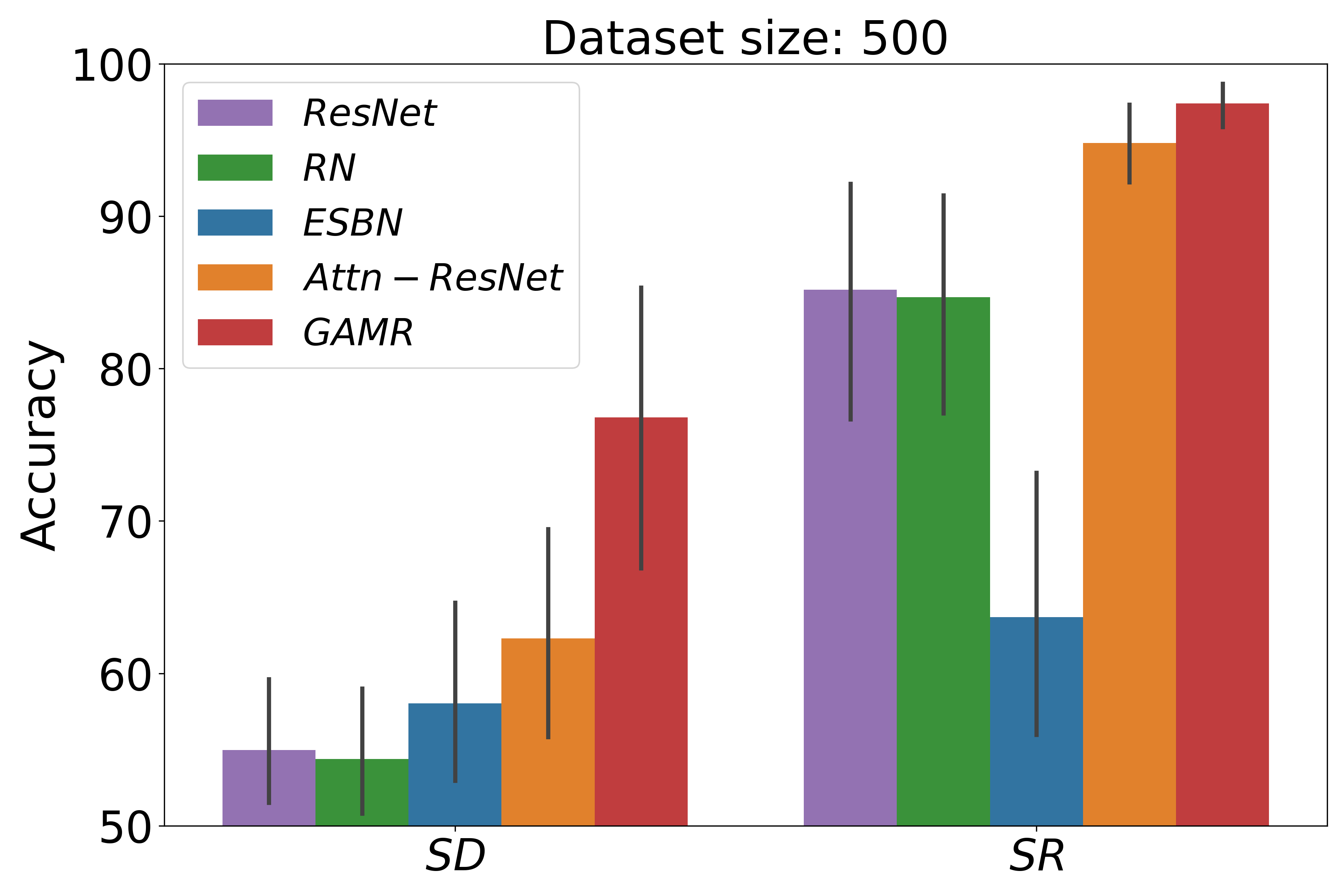} 
\includegraphics[width=.47\textwidth]{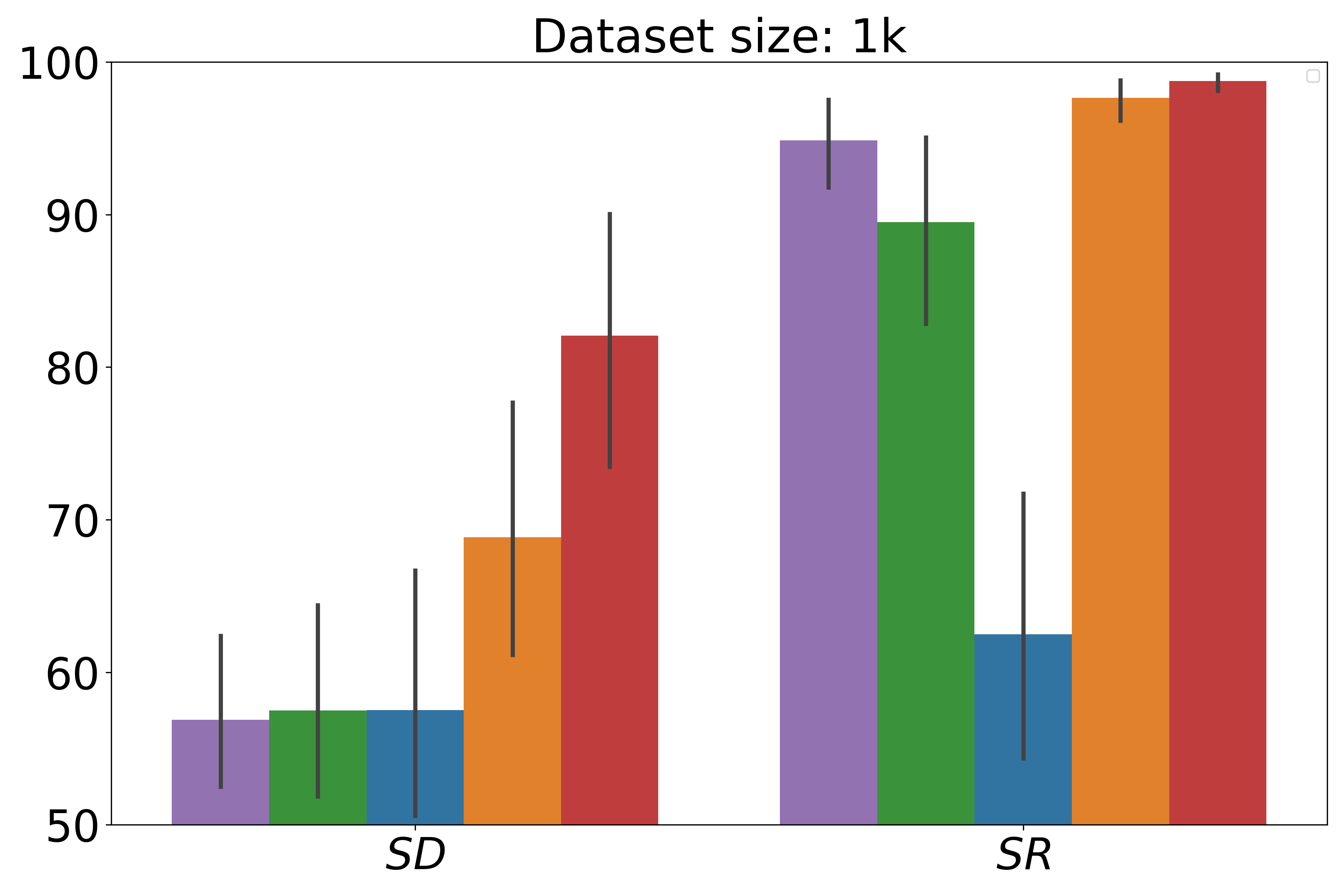} \\
\includegraphics[width=.47\textwidth]{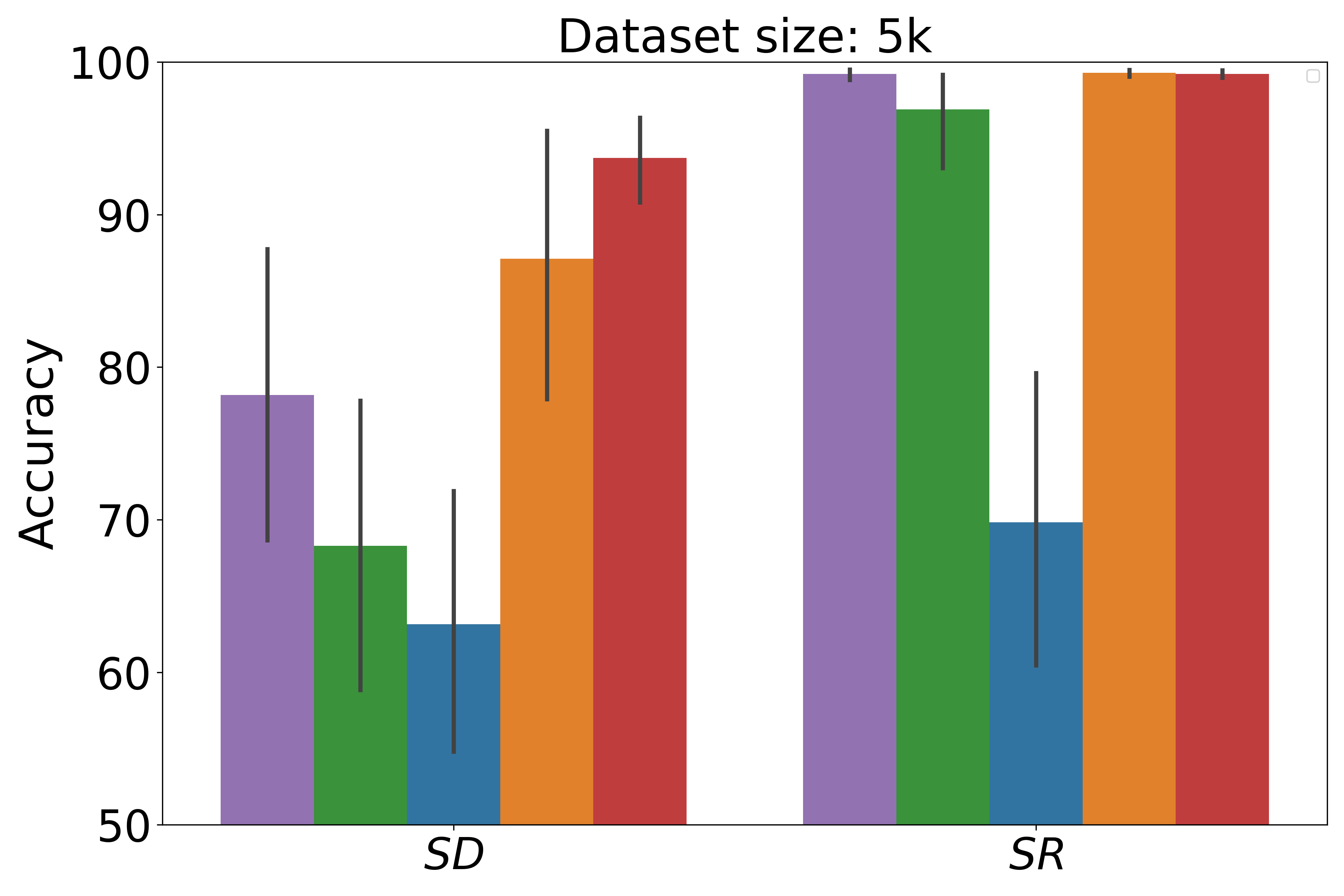} 
\includegraphics[width=.47\textwidth]
{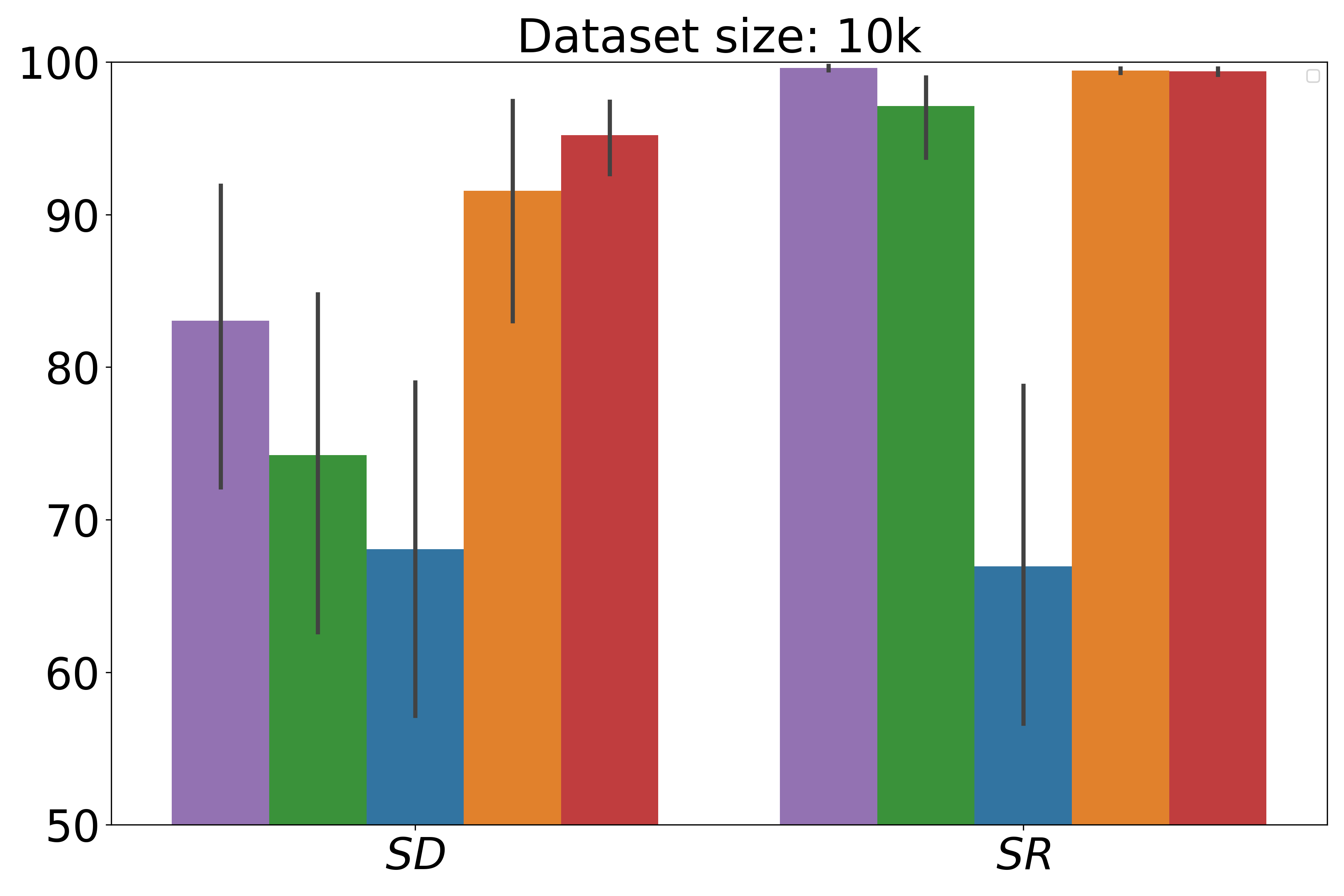} 
\caption{Bar plot analysis for the SVRT tasks grouped in same-different ($SD$) and spatially related ($SR$) tasks. We compared the accuracies of five baseline architectures with \textit{GAMR}. We trained these models with .5k, 1k, 5k and 10k samples.}
\label{fig:box}
\end{center}
\end{figure*}

All twenty-three tasks in the SVRT dataset can be broadly divided into two categories: same-different (SD) and spatial relations (SR), based on the relations involved in the tasks. Same-different tasks (\textit{1, 5, 6, 7, 13, 16, 17, 19, 20, 21, 22}) have been found to be harder for neural networks \citep{Ellis2015UnsupervisedLB,kim2018not,stabinger201625,stabinger2021evaluating,Puebla2021.04.06.438551,messina2021recurrent,vaishnav2021understanding} compared to spatial relations tasks (\textit{2, 3, 4, 8, 9, 10, 11, 12, 14, 15, 18, 23}). 

We analyzed an array of architectures and found that, on average, \textit{GAMR} achieves at least 15\% better test accuracy score on $SD$ tasks for 500 samples. In contrast, for $SR$ tasks, the average accuracy has already reached perfection. We find a similar trend for other architectures when trained with different dataset sizes. Overall, RN (GAMR minus attention) and ESBN struggled to solve SVRT tasks even with 10k training samples, pointing towards the lack of an essential component, such as attention. On the other hand, Attn-ResNet architecture demonstrated the second-best performance, which shows its importance in visual reasoning. Results are summarized in Figure~\ref{fig:box}.

\section{Learning Compositionality}
\label{subsec:learningcomp}

\begin{figure}[ht]
    \begin{center}
      \includegraphics[width=.7\linewidth]{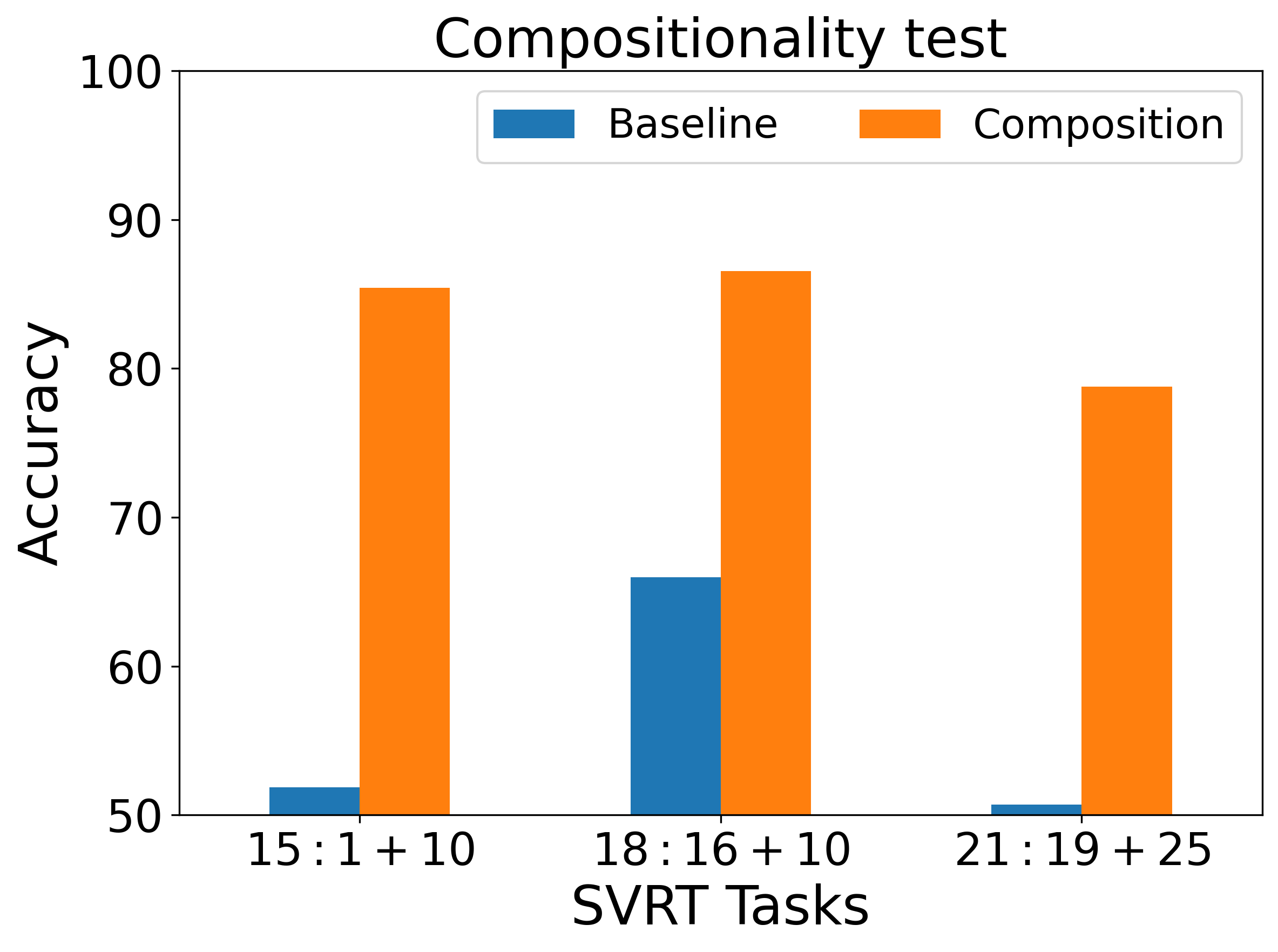}
      \caption{\textbf{Compositionality test}: We train the model with tasks containing specific rules (e.g., task \textit{1} representing same-different discrimination and  task \textit{10} involving identification if the four shapes form a square or not). We show that with its ability to compose already learned rules, \textit{GAMR} can quickly learn with 10 samples per class to adapt to a novel scenario (e.g., task \textit{15} where the rule is to identify if the four shapes forming a square are identical or not).}  
      \label{fig:comp}
    \end{center}
  
\end{figure}

Compositionality is the capacity to understand novel combinations from previously known components. While the human brain learns compositionally, deep learning models work on the learning Single task with a Single Model principle. Below, we provide evidence that \textit{GAMR} is capable of harnessing compositionality. We looked for triplets of tasks $(x,y,z)$ such that $z$ would be a composition of tasks $x$ and $y$. We systematically looked for all such available triplets in the SVRT dataset and found three triplets: (\textit{15, 1, 10}), (\textit{18, 16, 10}) and (\textit{21, 19, 25}). We study the ability of the network to learn to compose a new relation with very few training samples, given that it had previously learned the individual rules. We first trained the model with tasks $x$ and $y$ so that the rules are learned with the help of the reasoning module $r_{\theta}$ \textcolor{black}{which is a two-layered MLP}. We expect that the first layer learns the elementary operation over the context vectors stored in the memory block ($M$), while the second layer learns to combine those operations for the tasks $z$. We freeze the model after training with tasks $x$, $y$ and only fine-tune: (i) a layer to learn to combine elementary operations and (ii) a decision layer ($f_{\phi}$) on tasks $z$ with ten samples per category and 100 epochs in total. Results are shown in Figure~\ref{fig:comp}. \textcolor{black}{As our baseline, we trained the model from scratch on task $z$ from a triplet of tasks (\textit{x,y,z}) to show that the model is exploring indeed compositionality. We also ran an additional control experiment for compositionality choosing the random pair of tasks (\textit{x=5, y=17}) such that the rules are not included in tasks (z) \textit{15, 18}, and \textit{21}. When we evaluated the network in this setup, we found the performance of the network to be at the chance level -- aligning with the claim.}

We selected group corresponding to each tasks (\textit{15, 18, 21}) used for composition. Task \textit{15} has four shapes forming a square and are identical. It can be composed of task \textit{1}, helping to identify the same shapes and task \textit{10}, which helps to learn if the four shapes are forming a square. In task \textit{18}, a rule is needed to be learned related to symmetry along the perpendicular bisector of the image. It can be taken as a composition of task \textit{16} which requires learning mirror reflection of the image along the perpendicular bisector of the image and task \textit{10} in which symmetry could be discovered in between 4 shapes (forming a square). At last, we took task \textit{21}, which involves both scaling and rotation between two shapes in an image. As its compositional elements, we designed a variant where there is only rotation and no scaling and represented it with \textit{25} and combined it with another counterpart of \textit{21} where there is scaling and no rotation, i.e., task \textit{19}.

\section{Zero-shot generalization}
\label{sec:zeroshot}

\begin{table}[htbp]
\centering
    \begin{tabular}{cccccc}
\hline
Training            & Test & \multicolumn{3}{c}{Test Accuracy} \\\cline{3-5} 
Task            & Task & GAMR & Attn-ResNet & ResNet \\ \hline
\midrule
\multirow{3}{*}{1}  & 5    & 72.07          & 53.03 &  \textbf{73.04} \\
                    & 15   & \textbf{92.53} & 92.07 &   78.87  \\
                    & 22   & \textbf{84.91} & 80.10 &   67.15  \\\hline
\multirow{3}{*}{5}  & 1    & \textbf{92.64} & 85.73 &   92.28    \\
                    & 15   & \textbf{84.36} & 62.69 &   49.95    \\
                    & 22   & \textbf{76.47} & 55.69 &   50.19    \\\hline
7                   & 22   & \textbf{83.80} & 79.11 &   50.37   \\\hline
21                  & 15   & \textbf{90.53} & 50.00 &   49.76   \\\hline
23                  & 8    & \textbf{85.84} & 58.90 &   59.25   \\ 
\bottomrule
\end{tabular}

\caption{Test accuracy to show if the model learns the correct rules when we train it with a task and test on a different set of SVRT tasks with \textit{GAMR}, Attention with ResNet50 (\textit{Attn-ResNet}) and ResNet-50 (\textit{ResNet}).}
\label{tab:transfer}
\end{table}

We hypothesize that if a model has learned the abstract rule underlying a given task, it should be able to re-use its knowledge of this task on other novel tasks which share a similar rule. To verify that \textit{GAMR} is indeed able to generalize across tasks that share similar rules, we searched for pairs of tasks in SVRT which were composed of at least one common elementary relation \citep{vaishnav2021understanding} between them. For example, in pair (\textit{1}, \textit{22}), task \textit{1} involves the identification of \underline{\textit{two}} similar shapes in category 1 and task \textit{22} involves the identification of \underline{\textit{three}} similar shapes in category 1. In the selected pair, the category that judges the similar rule should belong to the same class (let us say category 1 in the above example) so that we test for the right learnability. We systematically identified a set $x$ of tasks \textit{1, 5, 7, 21, 23} representing elementary relations such as identifying same-different (\textit{1, 5}), grouping (\textit{7}), learning transformation like scaling and rotation (\textit{21}) and learning insideness (\textit{23}). Then we paired them with other tasks sharing similar relations. 
These pairs are task \textit{1} with each of \textit{5, 15 and 22}, task \textit{5} with each of \textit{1, 15 and 22}. Similarly other pairs of tasks are (\textit{7, 22}), (\textit{21, 15}) and (\textit{23, 8}). 
We separately trained the model on the set $x$ and tested the same model on their respective pairs without finetuning further with any samples from the test set (zero-shot classification). 
We observed that \textit{GAMR} could easily generalize from one task to another without re-training. On the contrary, a chance level performance by \textit{ResNet} shows the network's shortcut learning and rote memorization of task-dependent features. In comparison, \textit{GAMR} exhibits far greater abstraction abilities -- demonstrating an ability to comprehend rules in unseen tasks without any training at all. We further explored the strategies learned by \textit{GAMR} using attribution methods for all the tasks. These attribution methods confirm that \textit{GAMR} does indeed use a similar visual routine between the original task for which it was trained and the new task for which it was never trained. Table~\ref{tab:transfer} summarizes these results.

\section{Ablation Study}
\label{sec:abl}

\paragraph{Benchmarking guided attention}
\label{sec:ga}

\begin{figure}[ht]
  \begin{center}
  \includegraphics[width=.7\linewidth]{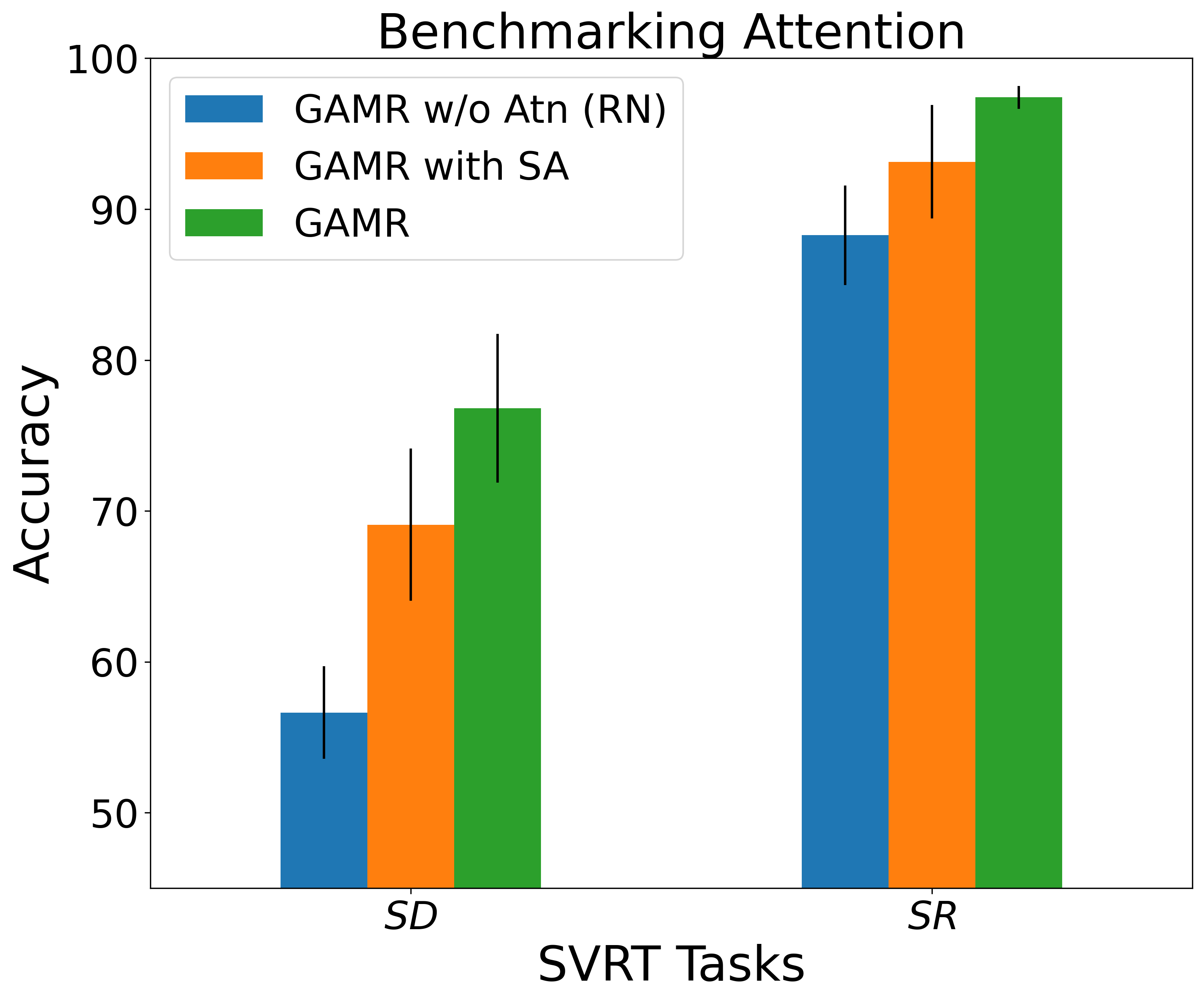}
  \caption{We compared the average accuracy over two sub-clusters of SVRT obtained by \textit{GAMR} with its variant when we replaced the guided-attention module with the self-attention (\textit{GAMR-SA}) and when we completely gave away attention and made it a relational reasoning architecture (\textit{GAMR w/o Atn (RN)}).}  
  \label{fig:gam}
  \end{center}
  
\end{figure}

We evaluated our guided-attention module (\textit{GAMR}) and compared it with alternative systems with comparable base-architecture but endowed with self-attention (\textit{With-SA}) or no attention and/or memory (\textit{GAMR w/o Atn (RN)})  over 23 SVRT tasks \textcolor{black}{for the same number of time steps. In \textit{GAMR with-SA}, we add a self-attention layer in the guided attention module and all three input vectors to the attention module are the same ($z_{img}$). Our intuition is that at each time step, the self-attention mechanism should provide the mechanism to learn to attend to different objects in a scene.} As a side note, \textit{GAMR with-SA} turns out to be similar to ARNe~\citep{hahne2019attention} used for solving Raven's tasks. We found that, on average, our Guided Attention Model's relative performance is 11.1\% better than its SA counterpart and 35.6\% than a comparable system lacking attention (or memory) for $SD$ tasks; similarly, relative improvements for $SR$ tasks are 4.5\% and 10.4\% higher. It shows that \textit{GAMR} is  efficient as it yields a higher performance for the same number (1k) of training samples. Results are shown in Figure~\ref{fig:gam}. 

\begin{figure}[htbp]
    \begin{center}
      \includegraphics[width=1\linewidth]{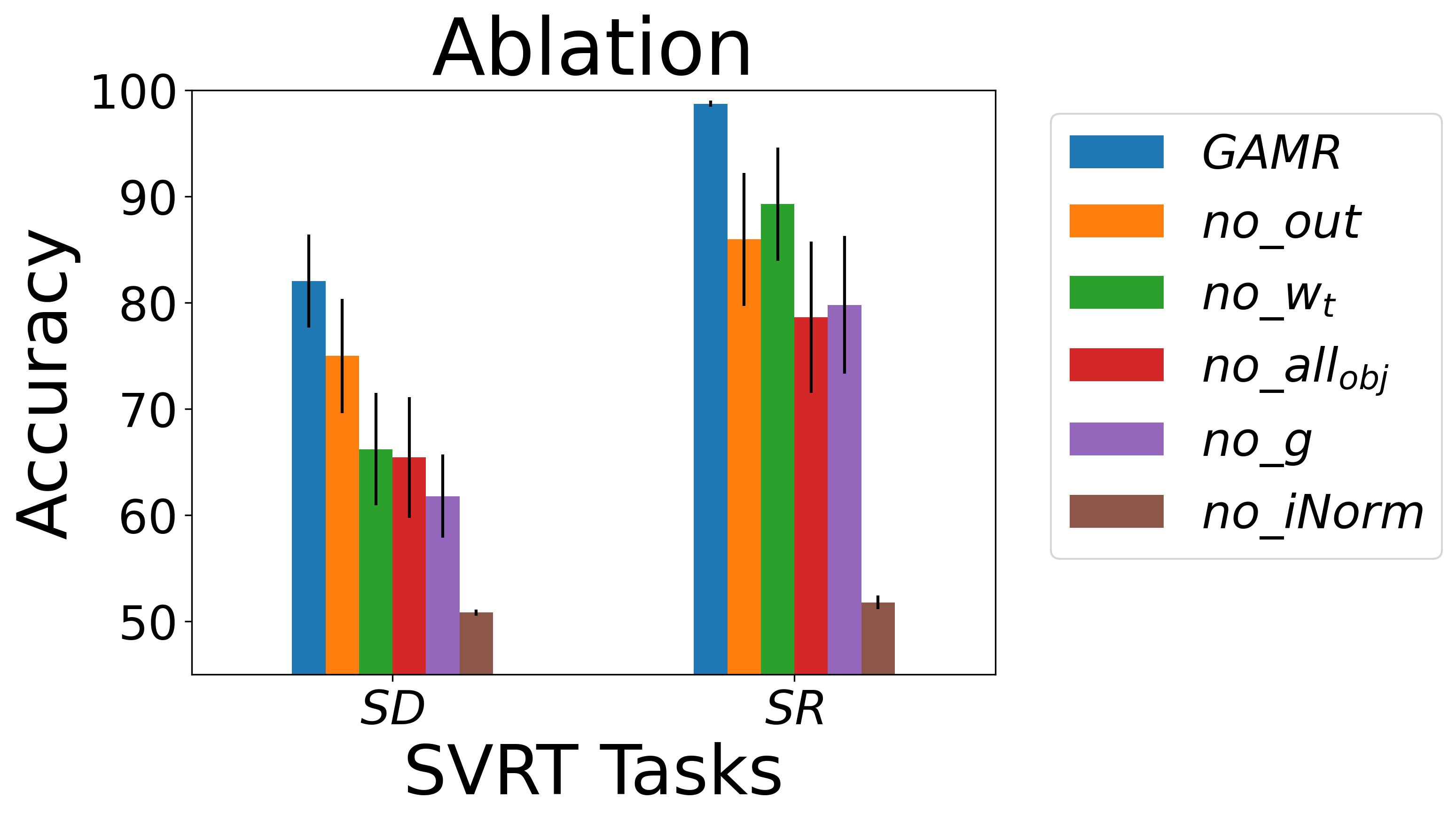} 
      
      \caption{
      \textbf{Ablation studies}: We pruned separate parts of the model, one at a time: controller output ($out$), attention vector ($w_t$), relational vector ($all_{obj}$), feature channel gain factor ($g$) and instance normalization ($iNorm$) and the bar plot show the variation in performance on SD and SR tasks when trained with 1k samples.
      }  
      \label{fig:abl}
    \end{center}
  
\end{figure}

\begin{figure}[htbp]
    \begin{center}
      \includegraphics[width=1\linewidth]{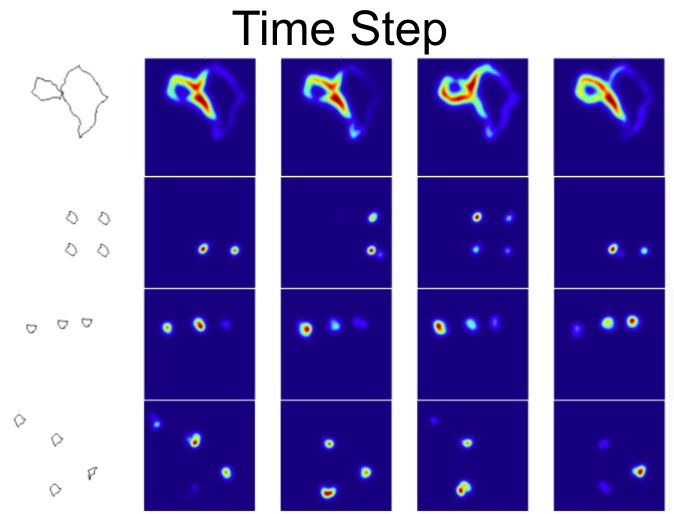} 
      
      \caption{
      \textbf{Time steps visualization}: Figure showing shift of attention with each time step in a task-dependent manner. In the first row, the task is to answer if the two shapes are touching each other from the outside. At each time step, the network explores the area where the shapes are touching each other. In other rows, attribution maps show the shifts over different shapes in an image. The controller module for the task in respective rows shifts attention across different shapes at each time step.
      }  
      \label{fig:ablts}
    \end{center}  
\end{figure}

\begin{figure}[ht]
\centering
  \includegraphics[width=.6\linewidth]{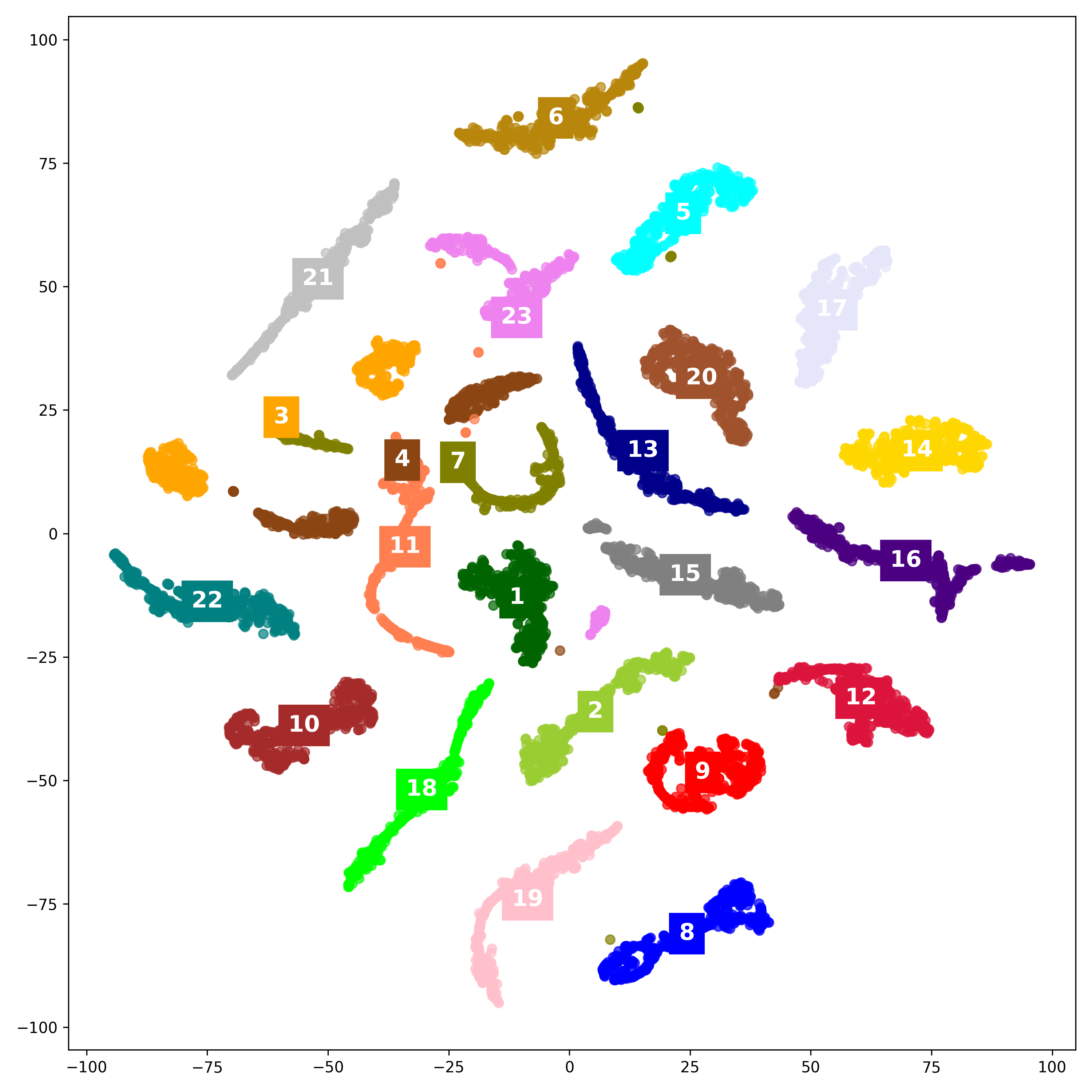}
  \caption{\textbf{Abstract variable:} t-SNE plot of the output vector (\textit{out}) obtained from the controller ($f_e$) for all 23 SVRT tasks independently. Each cluster can be clearly identified from other clusters representing different relations learned. Tasks are represented as labels with the same colored box around them placed at the mean location of the cluster.}  
  \label{fig:abstract}
\end{figure}

\textit{GAMR} is a complex model with several component, so we now proceed to study what role different components of the proposed architecture play in it's ability to learn reasoning tasks.  
We studied the effect of these components on SD and SR categories. Our lesioning study revealed that \textit{iNorm} plays a vital role in the model's reasoning and generalization capability even for learning simple rules of $SR$ tasks. Normalizing every sample individually helps the model learn the abstract rules involved in task.  We also found that for $SD$ tasks, excluding vector $out$ from the decision-making process is detrimental. The t-SNE plot 
shows that it encodes the independent abstract representation for the SVRT tasks (Figure~\ref{fig:abstract}). We systematically ran an ablation study to show that each of these components are essential to make it an efficient model. In $no\_all_{obj}$, the model takes a decision based on the final outcome ($out$) of the recurrent module ($f_s$); for $no\_w_{t}$, output of the attention block is used to obtain $z_t$ instead after projecting it on $z_{img}$; for $no\_g$, equal weighing is applied to the feature space of the context vectors stored in the memory block. We have summarized the results in Figure~\ref{fig:abl}. We also plot the attribution maps of the model in Figure~\ref{fig:ablts} at each time step and show the way in which the model attends to task-dependent features while learning the rule.

\section{Additional Experiment}
\label{sec:art}

\begin{figure*}[htbp]
\centering
  \includegraphics[width=1\linewidth]{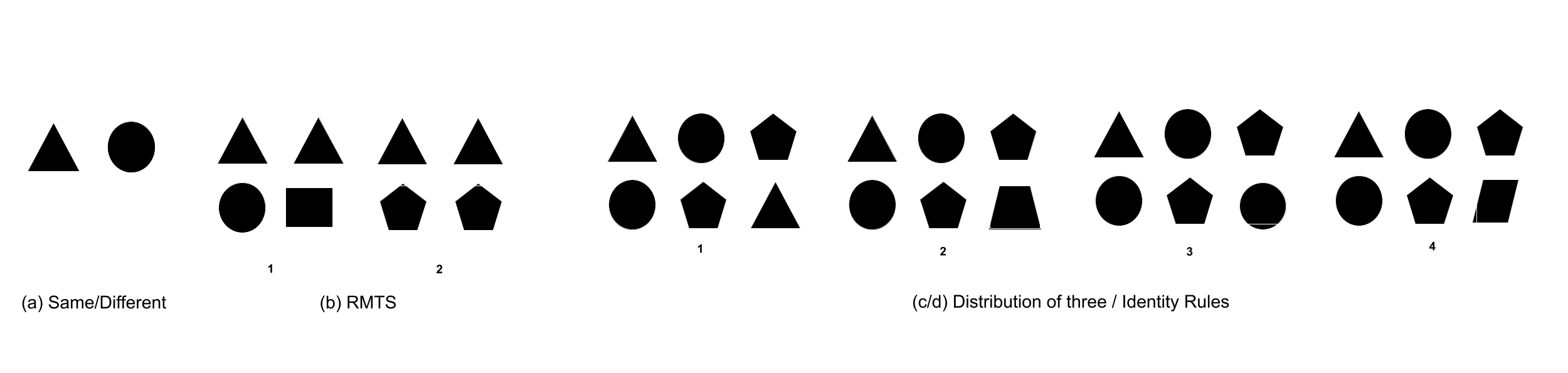}
  \caption{\textbf{ART for \textit{GAMR}:} (a) Same/different discrimination task. (b) Relational match-to-sample task (answer is 2). (c) Distribution-of-three task (answer is 1). (d) Identity rules task (ABA pattern, answer is 3).} 
  \label{fig:esbnGAMR}
\end{figure*}

\paragraph{Dataset} \citet{Webb2021EmergentST}  proposed four visual reasoning tasks (Figure \ref{fig:esbnGAMR}), that we will henceforth refer to as the \textit{Abstract Reasoning Task} (ART):  (1) a same-different (\textit{SD}) discrimination task, (2) a relation match to sample task (\textit{RMTS}), (3) a distribution of three tasks (\textit{Dist3}) and (4) an identity rule task (\textit{ID}). These four tasks utilize shapes from a set of 100 unique Unicode character images \footnote{\url{https://github.com/taylorwwebb/emergent_symbols}}. They are divided into training and test sets into four generalization regimes using different holdout character sets (m = 0, 50, 85, and 95) from 100 characters. We have described training and test samples and different hyperparameters for all four tasks in section~\ref{supp:hyper}.

\begin{figure*}[htbp]
\begin{center}
\includegraphics[width=.95\linewidth]{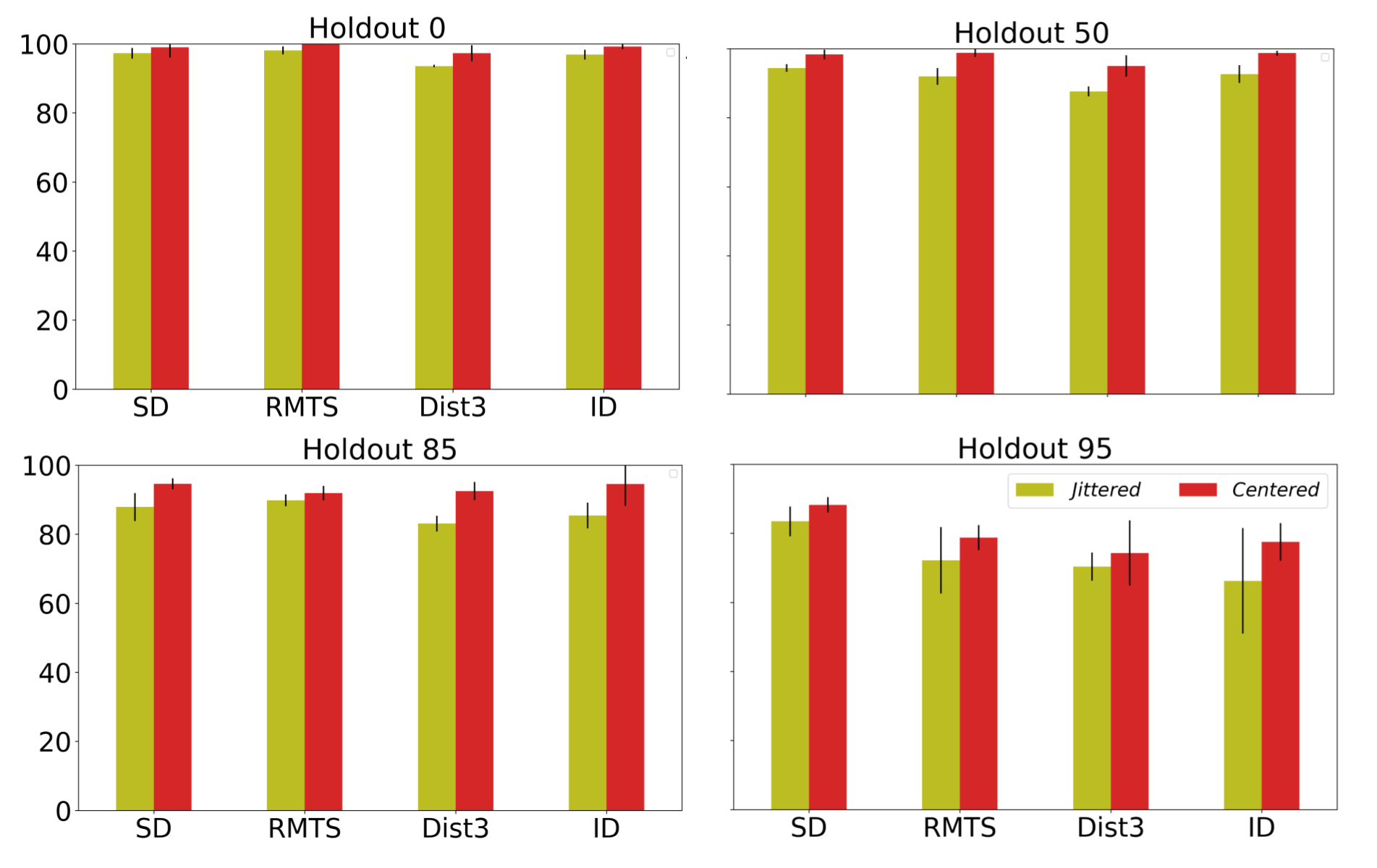} 

\caption{Test accuracy on ART with different holdout sets when the images are $centered$ and compare the accuracy when shapes are $jittered$ in every image. We find that unlike other baselines experiencing a huge drop in performance when shapes are jittered, GAMR is stable. We plot the average accuracy over ten runs on the dataset. $x$ axis corresponds to the four types of tasks, and $y$ represents the average accuracy score. These tasks are as follows: (a) same-different (SD) discrimination task, (b) Relation match to sample task (RMTS); (c) Distribution of three tasks (Dist3); and (d) Identity rule task (ID).}
\label{fig:esbn_result}
\end{center}
\end{figure*}

\paragraph{Baseline models}
As a baseline, we chose the ESBN \citep{Webb2021EmergentST} along with the two other prevalent reasoning architectures, the Transformer \citep{vaswani2017attention} and Relation Network (RN) \citep{santoro2017simple}. These three share a similar encoder backbone as in \textit{GAMR}. \textcolor{black}{We also ran an additional baseline with \textit{ResNet50} to verify if in a multi-object scenario \textit{GAMR} exploring some biases (like visual entropy) which is otherwise not present when segmented images are passed.} In order to make our baselines stronger, we evaluated these models in their natural order, i.e., by passing a single image at a time. We added a random translation (jittering) for the shapes in the area of $\pm 5$ pixels around the center to prevent these architectures from performing template matching. \textcolor{black}{This jittering increases the complexity of the tasks and hence we have to increase the number of time steps from 4 to 6. This caused the differences in results when compared to the implementation in \citet{Webb2021EmergentST}.}  For \textit{GAMR} and \textit{ResNet50}, we present task-relevant images together as a single stimulus Figure~\ref{fig:esbnGAMR} while jittering each shape. We have also added ART results where each image is centered and put together in a single stimulus in Figure~\ref{fig:esbn_result}. In order to make our architecture choose one option from multiple stimuli (\textit{RMTS}: 2, \textit{Dist3} and \textit{ID}: 4), we concatenate the relational vector ($all_{obj}$) for every stimulus and pass them to a linear layer for final decision.

\begin{figure}[htbp]
\begin{center}
\includegraphics[width=1\linewidth]{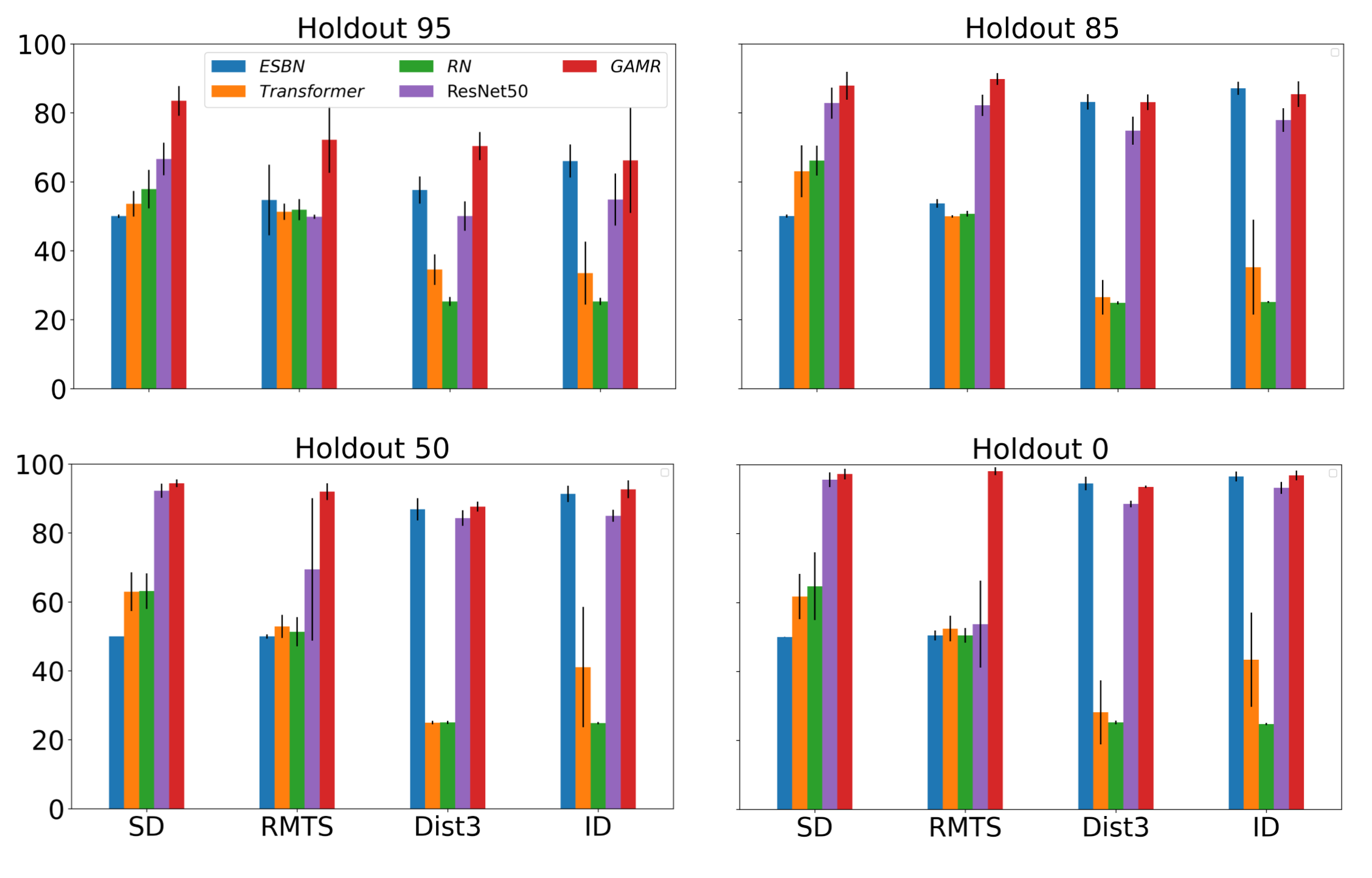} 
\caption{\textbf{ART}: Comparing the average performance of \textit{GAMR} with other baselines over 10 runs for different holdout values (m = 0, 50, 85, 95).  These models are evaluated on four types of tasks, i.e., Same-Different (SD), Relation match to sample (RMTS), Distribution of 3 (Dist3) and Identity rules (ID).}
\label{fig:esbbnresultall}
\end{center}
\end{figure}

\paragraph{Results}
We found a near-chance level (50\%) accuracy for all the baseline models and in all the four generalization regimes for the \textit{SD} and \textit{RMTS} tasks (Figure~\ref{fig:esbbnresultall}) which otherwise performed with perfection when images were centered and passed through the same models.
However, our proposed architecture is robust to handle this jittering, as shown in Figure~\ref{fig:esbn_result} where we compare its performance when images are not jittered. For the other two tasks, \textit{Dist3} and \textit{ID}, baseline models performed better than the chance level (25\%). \textit{ESBN} showed an increasing trend in accuracy for progressively easier generalization conditions approaching 0 holdouts. This points toward the fact that the first three shapes in both tasks allow \textit{ESBN} to consider a translation factor while comparing the next three shapes, letting it choose the correct option appropriately. \textit{RN} and \textit{Transformer} consistently struggled to generalize. \textit{ESBN} (memory-based model) performance on SD tasks in both the visual reasoning datasets show that attention is needed for reasoning. 

\section{Hyperparameters}

\label{supp:hyper}
\begin{table}[htbp]
\caption{\textbf{ART}: Number of training and test samples used for four different types of tasks.}
\label{tab:artsamples}
\begin{center}
\begin{tabular}{cccccc}
\\\hline
\textbf{Tasks}      &          & m=0    & m=50   & m=85   & m=95   \\ \hline
\multirow{2}{*}{SD} & Training & 18,810 & 4,900  & 420    & 40     \\
                    & Test     & 990    & 4,900  & 10,000 & 10,000 \\ \hline
RMTS                & Training & 10,000 & 10,000 & 10,000 & 480    \\
Dist3               & Training & 10,000 & 10,000 & 10,000 & 360    \\
ID                  & Training & 10,000 & 10,000 & 10,000 & 8,640  \\
                    & Test     & 10,000 & 10,000 & 10,000 & 10,000 \\ \hline
\end{tabular}
\end{center}
\end{table}

\paragraph{Holdout set} For example, holdout \textit{0} represents a generalization regime in which the test sets contain the same characters as those used during training. At the other extreme, in holdout \textit{95}, the training set contains a minimal number of characters, most of which are actually used for tests. Hence, it is necessary to learn the abstract rule in order to generalize to characters in this regime.  

\begin{table}[H]
\caption{\textbf{ART}: For four different tasks number of epochs and learning rates (LR) used to train different architectures.}
\label{tab:arttraining}
\vskip 0.15in
\begin{center}
\resizebox{\textwidth}{!}{%
\begin{tabular}{ccccccccc}
\hline
\textbf{Tasks} & \multicolumn{2}{c}{m=0}                      & \multicolumn{2}{c}{m=50}                     & \multicolumn{2}{c}{m=85}                     & \multicolumn{2}{c}{m=95} \\ \cline{2-9} 
        & \multicolumn{8}{c}{GAMR}                                                                                                                                     \\ \cline{2-9} 
        & Epoch & \multicolumn{1}{c|}{LR} & Epoch & \multicolumn{1}{c|}{LR} & Epoch & \multicolumn{1}{c|}{LR} & Epoch  & LR \\ \hline 
SD      & 50    & \multicolumn{1}{c|}{0.0001}        & 50    & \multicolumn{1}{c|}{0.0005}        & 100   & \multicolumn{1}{c|}{0.0005}        & 200    & 0.001         \\
RMTS    & 50    & \multicolumn{1}{c|}{0.00005}        & 50    & \multicolumn{1}{c|}{0.0001}        & 50    & \multicolumn{1}{c|}{0.0005}        & 300    & 0.0005        \\
Dist3   & 50    & \multicolumn{1}{c|}{0.00005}       & 50    & \multicolumn{1}{c|}{0.0001}        & 50    & \multicolumn{1}{c|}{0.00005}       & 300    & 0.0005        \\
ID      & 50    & \multicolumn{1}{c|}{0.00005}       & 50    & \multicolumn{1}{c|}{0.00005}       & 50    & \multicolumn{1}{c|}{0.0005}        & 100    & 0.0005        \\ \cline{2-9} 
        & \multicolumn{8}{c}{Other baselines}                                                                                                                           \\ \cline{2-9} 
SD      & 50    & \multicolumn{1}{c|}{0.0005}        & 50    & \multicolumn{1}{c|}{0.0005}        & 100   & \multicolumn{1}{c|}{0.0005}        & 200    & 0.0005        \\
RMTS    & 50    & \multicolumn{1}{c|}{0.0005}        & 50    & \multicolumn{1}{c|}{0.0005}        & 50    & \multicolumn{1}{c|}{0.0005}        & 300    & 0.0005        \\
Dist3   & 50    & \multicolumn{1}{c|}{0.0005}        & 50    & \multicolumn{1}{c|}{0.0005}        & 50    & \multicolumn{1}{c|}{0.0005}        & 300    & 0.0005        \\
ID      & 50    & \multicolumn{1}{c|}{0.0005}        & 50    & \multicolumn{1}{c|}{0.0005}        & 50    & \multicolumn{1}{c|}{0.0005}        & 100    & 0.0005        \\ \hline
\end{tabular}}
\end{center}
\end{table}

\section{Conclusion and limitations}
\label{sec:conc}
In this paper, we described a novel Guided Attention Module for (visual) Reasoning (\textit{GAMR}) to bridge the gap between the reasoning abilities of humans and machines. 
Inspired by the cognitive science literature, our module learns to dynamically allocate attention to task-relevant image locations and store relevant information in memory. Our proposed guided-attention mechanism is shown to outperform the self-attention mechanisms commonly used in vision transformers. Our ablation study demonstrated that an interplay between attention and memory was critical to achieving robust abstract visual reasoning. 
Furthermore, we demonstrated that the resulting systems are capable of solving novel tasks efficiently -- by simply rearranging the elemental processing steps to learn the rules without involving any training. 
We demonstrated GAMR's versatility, robustness, and ability to generalize compositionality through an array of experiments. We achieved state-of-the-art accuracy for the two main visual reasoning challenges in the process.  One limitation of the current approach is that it currently only deals with a fixed number of time steps. Training the model with four time steps was sufficient to solve all SVRT tasks efficiently. However, a more flexible approach is needed to allow the model to  automatically allocate a number of time steps according to the computational demand of the task. GAMR is also limited to covert attention unlike biological system where both covert and overt attention are demonstrated.

\part*{Chapter 5}
\chapter{Discussion and Future work}

\startcontents[chapters]

Attention, a concept extensively explored in cognitive science and machine learning, has made significant strides in the fields of computer vision and natural language processing. Self-attention-based architectures, prevalent in these domains, have demonstrated remarkable performance on various benchmarks. The self-attention mechanism employed by Transformer networks bears resemblance to the central executive controller attention theory proposed in cognitive psychology, as both are concerned with attentional control and cognitive processing. According to the central executive controller attention theory, attentional control relies on a central executive system that monitors and governs the flow of information within working memory. This system selectively attends to different components of the working memory representation based on task demands and information relevance.

Similarly, the self-attention mechanism in Transformer networks enables models to selectively attend to specific elements of the input sequence, guided by their relevance to the current task. This mechanism can be viewed as an abstraction of the central executive system, allowing the model to dynamically adjust its attentional focus in response to the input and its current state.

Furthermore, both the central executive system and the self-attention mechanism in Transformers involve the integration of multiple information sources, including internal representations and external cues, to facilitate attentional control and cognitive processing. In Transformers, this integration is achieved through multiple self-attention layers, enabling the model to construct an internal representation of the input sequence that captures the most pertinent features for the task at hand. Although the self-attention mechanism in Transformers and the central executive system in cognitive psychology are not identical, they share similar principles of attentional control and cognitive processing, enabling flexible and adaptive behavior to address evolving task demands and environmental cues.

Contrasting attention in Transformers with attention in the human mind reveals several distinguishing characteristics. In Transformers, attention "heads" operate independently in relation to the previous layer, indicating a lack of integration and coherence when compared to the integrated attention observed in the human mind. Transformers exhibit offline processing during learning, focusing solely on the present configuration, while attention in the human mind is influenced by past states, shaping the interpretation of the present and future.

Additionally, representations in Transformers are static, whereas in the human mind, representations are dynamic. Transformers employ features for scene segmentation, while in the human mind, features serve as parametric operators that facilitate scene predictions. Moreover, attention in Transformers does not drive active perception, unlike attention in the human mind, which plays a crucial role in guiding active perception and goal-directed cognition. Besides, attention in the human mind integrates latent features into a scene graph, enhancing the coherence and organization of information, a characteristic not explicitly addressed in the context of Transformers. These comparisons highlight the distinctions between attention mechanisms in Transformers and the human mind, emphasizing the integrated, dynamic, and goal-directed nature of attention within the human cognitive system.

Furthermore, attention, as a cognitive process enabling focused concentration on relevant stimuli, plays a vital role in enhancing human reasoning ability. To deepen our understanding of the self-attention mechanism, this thesis investigates its role in cognitive and computer vision architectures, particularly within the domain of visual reasoning. By exploring the interaction between self-attention and cognitive processes, we aim to contribute to the advancement of both cognitive science and machine learning, ultimately enhancing our comprehension and utilization of attention mechanisms.

Visual reasoning is the process of analyzing the provided visual information in order to solve a task. It is considered an important part of fluid intelligence, which involves thinking and reasoning independent of learning, education, and experience. This ability has not only been shown in primates \citep{gentner2021learning} but also in bees \citep{giurfa2001concepts} and in newborn ducklings \citep{martinho2016ducklings}. On the contrary, prior studies \citep{Puebla2021.04.06.438551,kim2018not,ricci37same,messina2021solving} (including our own work) have shown that modern-day neural networks struggle to solve simple visual reasoning tasks when tested on a popular benchmark called synthetic visual reasoning test (SVRT) by \citet{fleuret2011comparing} otherwise simple for humans. We found a similar trend when we tested popular reasoning architectures like Relational Network~\citep{santoro2017simple}, Transformer~\citep{vaswani2017attention}, ESBN~\citep{Webb2021EmergentST} on Abstract Reasoning Task (ART) where the stimulus contains a simple Unicode character. As a result, visual reasoning has become an increasingly popular topic of research in recent years with the emergence of numerous fluid intelligence tests for AI algorithms, including tests for Compositional Visual Reasoning (CVR)  \citep{zerroug2022benchmark}, Ravens' (RPM) \citep{barrett2018measuring,zhang2019raven} and visual progressive matrices (V-PROM) \citep{barrett2018measuring,teney2020v} as well as an Abstract Reasoning Corpus (ARC) \citep{chollet2019measure}. 

We began this thesis by studying the computational mechanisms involved in solving the Synthetic Visual Reasoning Test (SVRT) challenge \citep{fleuret2011comparing}. This challenge consists of twenty-three binary classification tasks, each involving unique abstract relations in their formulation. Previous studies have identified two broad categories of SVRT tasks \citep{Stabinger2016,kim2018not,yihe2019program} -- tasks involving spatial-relation (\textit{SR}) judgment and tasks involving same-different (\textit{SD}) judgment. The same-different tasks are found to be harder for the neural networks compared to the spatial relation tasks \citep{Ellis2015UnsupervisedLB,kim2018not,stabinger201625,stabinger2021evaluating,Puebla2021.04.06.438551,messina2021recurrent,vaishnav2021understanding}. Consistent with this work, we proposed a novel taxonomy beyond the two primary clusters, reflecting the number of relationships used to define a particular task. A closer examination is needed to better understand the trend reflected by the neural networks in terms of accuracy and the number of relations involved in defining a particular task. An earlier study by \citet{kim2018not} has also reported that feedforward neural networks demonstrate a `straining' effect in solving tasks involving same-different relations and hypothesized that the straining effect might be because of the lack of attention. The same was also shown with a human EEG experiment by \citet{alamia2021differential} where higher activity is recorded in the lower $\beta$ band while solving the same-different judgment when compared to spatial relation judgment indicating higher demands for attention and/or working memory. To test the same, in the next chapter, we focused on understanding the role of attention in solving visual reasoning tasks.

Inspired by the two types of visual attention, we proposed a self-attention module that can be used as a \textit{feature-based} or \textit{spatial} attention to augment the features of a feedforward network (ResNet50 \citep{he2016deep}). We evaluated both types of attention-augmented neural networks on SVRT tasks and found that our proposed attentional models could solve the most challenging SVRT tasks efficiently. The relative improvements obtained by feedforward networks endowed with the two different forms of attention varied across SVRT tasks. We observed that many tasks benefited from spatial attention mechanisms, whereas a few tasks from feature-based attention and showed a significant improvement. 
Our computational analysis also leads to testable predictions for human experiments by suggesting tasks that benefit from spatial attention (task 22) or feature-based attention (task 21), tasks that benefit from either form of attention (task 19), and tasks that do not benefit from attention (task 2). While we evaluated two types of attention systems, there is a future possibility to add experiments with the third type of attention -- object-based attention \citep{duncan1984selective,egly1994shifting,vecera1994does,kramer1997object}. Object-based attention focuses on the particular object rather than its spatial location or corresponding features.

In the last part of the thesis, we proposed a novel architecture, the Guided Attention Model for (visual) Reasoning (\textit{GAMR}). We integrated both cognitive abilities humans use -- attention and memory in solving reasoning tasks. It draws inspiration from the cognitive science literature on active vision, where the spotlight of attention is routed in the visual system to gather task-relevant information. According to the theory of active vision, the visual world is explored using rapid eye movements guided by shifts of visual attention. We designed a controller akin to the mechanisms involved in the active vision framework to route the spotlight of attention and send the task-relevant representations in the memory block later used for reasoning. In GAMR, the controller is implemented with the key/query/value-based self-attention layer. Contrary to the existing method where key, query and value vectors all correspond to the same vector, the query is internally generated at each time step in our model. It helps the controller to shift the spotlight of attention. 
One of the limitations of the current approach is the fixed number of time steps. I believe that a future continuation of this work could be to incorporate a mechanism to adapt the number of time steps based on the complexity of the task. For now, we have set the number of time steps as four for all the tasks; however, a simple task might require fewer time steps to arrive at a decision with high confidence. To make the model adaptive to the situation, one possibility could be to train it with a confidence variable as a stopping criterion. 

While we have limited our analysis to synthetic visual reasoning datasets, a future possibility exists to test the models on a real-world dataset like V-PROM. It consists of images organized in a Ravens' style of reasoning with some context images and some choice images from which the correct answer is selected. Another possible direction is to think of an architecture that considers two important traits -- efficient use of data and efficient use of the computational resource. One way to design this architecture is by incorporating a read-and-write mechanism similar to a Neural Turing Machine \citep{graves2014neural}. Both these mechanisms will help the network read the already stored relations from memory and write them into the memory if they are novel. We expect such cognitive architecture to demonstrate higher-order reasoning ability, continual learning, compositionally, and meta-learnability.

We also evaluated ViT~\citep{Dosovitskiy2020-iq} -- a full self-attention architecture on SVRT tasks and found that it struggles to learn the simplest of the SVRT tasks; however, \citet{messina2021recurrent} conducted a similar study on a smaller subset of four SVRT tasks trained on 28k samples and found that a recurrent version of ViT -- an attentional network with a convolutional backbone can learn those tasks. Adding convolutions in the early layers of ViT is found to help obtain better accuracy and improve sensitivity to the optimization settings \citep{xiao2021early}. This observation motivated us to propose \textit{Conviformer} \citep{vaishnav2022conviformers} for another collaborative project on leaf-fossil classification. We propose a network to incorporate a convolutional network as the front end for a full self-attention-based vision transformer network enhancing its ability to process higher-resolution images. While bigger images hold great importance in computer vision applications like object detection, segmentation and fine-grained classification, they cannot be used with vision transformers because of the associated computational memory demand. \textit{Conviformer} improves the performance of vision transformers by incorporating local features and infusing convolutional priors in a transformer architecture. We would like to see how convolution-induced vision transformers perform on SVRT tasks. 

Concept learning is yet another exciting direction of research. One of the key features of human intelligence is the ability to quickly learn new concepts and use them to generalize to a novel scenario. A \textit{concept} can be an idea representing a class of events (e.g., walking), objects (e.g., cats), or their properties (e.g., blue color). To test the concept learning ability of neural networks in a few-shot manner, we recently introduced a novel visual reasoning dataset, Compositional Visual Reasoning (\textit{CVR})~\citep{zerroug2022benchmark}. This dataset is based on the principle of odd-one-out reasoning. In this form of reasoning task, three out of four samples follow a similar concept (rule) in their formulation, while the fourth does not. Each sample contains shapes similar to the shapes used in the SVRT challenge. It extends the variety of relations used in the formulation compared to previously defined datasets like SVRT or RPM. We have also included compositionality prior in the dataset, where some elementary relations are used to compose the several tasks. The motivation is to push the community to build a compositional and sample-efficient network.

In this thesis, we embarked on one of the pioneering endeavors to investigate self-attention through the lens of visual reasoning. Attention assumes a pivotal role in showcasing visual reasoning capabilities, and an enhanced attentional model holds promise for improved reasoning abilities. We elucidated how self-attention operations can serve as a computational model for a visual attention system, encompassing spatial and feature-based attention, while also acting as a model for active vision. Our findings indicate that self-attention is equally effective in addressing reasoning tasks as it is in tackling other challenges in the realm of vision. However, further analysis is required to unravel the underlying mechanisms within a comprehensive self-attention model, which may constrain its sample-efficient learnability for reasoning tasks. Collectively, this work exemplifies the potential advantages of incorporating self-attention mechanisms into cognitive and computer vision architectures to conquer visual reasoning tasks.

\part*{Chapter 6}
\chapter{Publications}
\startcontents[chapters]
\begin{itemize}

    \item \textbf{Mohit Vaishnav}, Thomas Serre. ``GAMR: A Guided Attention Model for (visual) Reasoning.'' \textit{International Conference on Learning Representations (ICLR)} 2023,  \href{https://openreview.net/forum?id=iLMgk2IGNyv}{https://openreview.net/forum?id=iLMgk2IGNyv}

    \item \textbf{Mohit Vaishnav}, Remi Cadene, Andrea Alamia, Drew Linsley, Rufin VanRullen, Thomas Serre; ``Understanding the Computational Demands Underlying Visual Reasoning.'' \textit{Neural Computation} 2022; 34 (5): 1075–1099. doi: \url{https://doi.org/10.1162/neco_a_01485}
    
    \item Aimen Zerroug, \textbf{Mohit Vaishnav}, Julien Colin, Sebastian Musslick, Thomas Serre. `` A Benchmark for Compositional Visual Reasoning.'' \textit{In Proceedings of the Neural Information Processing Systems Track on Datasets and Benchmarks} \url{abs/2206.05379} (2022)
    
    \item \textbf{Mohit Vaishnav}, Thomas Fel, Ivan Rodriguez, Thomas Serre. ``Conviformers: Convolutionally guided Vision Transformer.'' \textit{ArXiv} \url{abs/2208.08900} (2022) (\textit{in preparation})

\end{itemize}

\part*{Chapter 7}

\chapter{Summary in French}

\startcontents[chapters]
\color{blue}

\color{black}
L'attention est un domaine amplement discut\'e et \'etudi\'e en neurosciences, en psychologie, en sciences cognitives et en apprentissage automatique \citep{chun2011taxonomy,cho2015describing}. L'attention est le processus consistant \`a se concentrer de manière s\'elective sur un aspect discret de l'information tout en ignorant les autres informations perceptibles. Une caract\'eristique largement accept\'ee de l'attention est qu'elle facilite l'utilisation efficace des ressources informatiques disponibles.
Bien que l'attention ait \'et\'e \'etudi\'ee depuis des d\'ecennies, elle est encore loin d'\^etre un concept simple ou unifi\'e \citep{lindsay2020attention}.

La litt\'erature en sciences cognitives d\'epeint plusieurs aspects de l'attention, comme le fait qu'elle puisse \^etre concentr\'ee, focalisée sur une modalit\'e particuli\`ere, divis\'ee, \^etre s\'elective et avoir une capacit\'e finie. Cependant, la s\'electivit\'e reste son trait le plus caract\'eristique. La s\'electivit\'e est n\'ecessaire en raison de la disponibilit\'e limit\'ee des ressources. R\'ecemment, l'attention visuelle a fait l'objet d'un intérêt consid\'erable dans le domaine de l'intelligence artificielle. L'attention visuelle \citep{ahmad1991visit} est la capacit\'e \`a hi\'erarchiser les informations tout en n\'egligeant les informations non pertinentes pour contenir la surcharge de donn\'ees dans notre syst\`eme visuel. L'attention visuelle permet de r\'epondre \`a la question: \textit{quoi} regarder et \textit{où} regarder. Cela a \'et\'e largement \'etudi\'ee en psychologie et en neurosciences \citep{posner1990attention,bundesen1990theory,desimone1995neural,corbetta2002control,petersen2012attention}. Ces \'etudes ont \'et\'e une source d'inspiration pour plusieurs mod\`eles d'intelligence artificielle \citep{khosla2007bio,lindsay2018biological,vaishnav2021understanding,vaishnav2022mareo}. 

Il existe trois cat\'egories de s\'electivit\'e dans un syst\`eme d'attention visuelle: par localisation spatiale \textit{(space-based)}~\citep{posner1980orienting,posner1982neural}, par appartenance \`a un objet \textit{(object-based)}~\citep{duncan1984selective,egly1994shifting,vecera1994does,kramer1997object} et par des caract\'eristiques particuli\`eres de l'entr\'ee \textit{(feature-based)}~\citep{harms1983color,driver1989movement,kramer1991perceptual,baylis1992visual,duncan1996objects}. 

L'attention est \'egalement sollicit\'ee lors de l'ex\'ecution de tâches n\'ecessitant des signaux sensoriels multiples. En pr\'esence de tâches ou de signaux sensoriels multiples, le contrôleur ex\'ecutif central aide \`a diriger l'attention. Le contrôleur ex\'ecutif central est charg\'e de coordonner l'activit\'e avec le syst\`eme cognitif pour diriger l'attention, prendre des d\'ecisions et maintenir les objectifs de la tâche. Le contexte et l'historique sont consid\'er\'es comme utiles \`a l'ex\'ecution optimale des tâches, ce qui les rend \'etroitement li\'es \`a la m\'emoire de travail. L'attention est en outre consid\'er\'ee comme le r\'esultat du contrôleur central. Le contrôleur s\'electionne les cibles de l'attention et les transmet au syst\`eme responsable de sa mise en œuvre. Il existe une relation tripartite entre le contrôle ex\'ecutif, la m\'emoire de travail et l'attention, de telle sorte que le centre d'attention est s\'electionn\'e par le contrôleur ex\'ecutif en fonction du contenu de la m\'emoire de travail ~\citep{soto2008automatic}. Bien que tous les objets de la m\'emoire de travail puissent influencer l'attention, le contrôleur ex\'ecutif aide \`a d\'ecider lequel doit affecter le plus~\citep{olivers2011difference}. Ces \'etudes cognitives vastes et approfondies li\'ees \`a l'attention ont inspir\'e le domaine de l'IA et ont contribu\'e \`a stimuler ses performances.

L'attention a profond\'ement marqué le domaine de la vision par ordinateur et du traitement naturel du langage (NLP), qui a connu un essor des architectures bas\'ees sur la \textit{self-attention} atteignant des performances de pointe sur de nombreux benchmarks. En outre, l'attention est un processus cognitif qui permet de se concentrer sur un stimulus pertinent. Cette caract\'eristique joue un rôle essentiel dans l'enrichissement de la capacit\'e de raisonnement des humains. Pour mieux comprendre le m\'ecanisme d'auto-attention, cette th\`ese a \'etudi\'e son rôle dans les architectures cognitives et de vision par ordinateur sous l'angle du raisonnement visuel.    

\color{black}
Le raisonnement visuel est le processus d'analyse des informations visuelles à disposition afin de r\'esoudre une tâche. Il est consid\'er\'e comme une partie importante de l'intelligence fluide, qui implique de penser et de raisonner ind\'ependamment de l'apprentissage, de l'\'education et de l'exp\'erience. Cette capacit\'e a \'et\'e d\'emontr\'ee non seulement chez les primates \citep{gentner2021learning} mais aussi chez les abeilles \citep{giurfa2001concepts} et chez les canetons nouveau-n\'es \citep{martinho2016ducklings}. Au contraire, des \'etudes ant\'erieures \citep{Puebla2021.04.06.438551,kim2018not,ricci37same,messina2021solving} (y compris nos propres travaux) ont montr\'e que les r\'eseaux neuronaux modernes peinent \`a r\'esoudre des tâches de raisonnement visuel simples lorsqu'ils sont test\'es sur un rep\`ere populaire appel\'e test de raisonnement visuel synth\'etique (SVRT) par \citet{fleuret2011comparing}, autrement simple pour les humains. Nous avons constat\'e une tendance similaire lorsque nous avons test\'e des architectures de raisonnement populaires de types \textit{Relational Network}~\citep{santoro2017simple}, \textit{Transformer}~\citep{vaswani2017attention} ou encore \textit{ESBN}~\citep{Webb2021EmergentST} sur la tâche de raisonnement abstrait (ART) où le stimulus contient un simple caract\`ere Unicode. Par cons\'equent, le raisonnement visuel est devenu un sujet de recherche de plus en plus populaire ces derni\`eres ann\'ees , en particulier avec l'\'emergence de nombreux tests d'intelligence fluide pour des algorithmes d'IA;  notamment les tests de raisonnement visuel compositionnel (CVR) \citep{zerroug2022benchmark}, Ravens (RPM) \citep{barrett2018measuring,zhang2019raven} et les matrices visuelles progressives (V-PROM) \citep{barrett2018measuring,teney2020v} ainsi qud les Corpus de Raisonnement Abstrait (ARC) \citep{chollet2019measure}. 

L'objectif de la premi\`ere \'etude de cette th\`ese \' a été de mettre en lumi\`ere les m\'ecanismes computationnels qui sous-tendent le raisonnement visuel \`a l'aide du \textit{Synthetic Visual Reasoning Test} (SVRT)~\citep{fleuret2011comparing}. Ce défi comprend vingt-trois problèmes de classification binaire, comprenant une vari\'et\'e de tâches de raisonnement identique-diff\'erent et spatial.  Dans notre exp\'erience, nous avons syst\'ematiquement \'evalu\'e la capacit\'e d'une batterie de $N=15$ r\'eseaux neuronaux convolutifs profonds (\textit{ResNets}) -- variant en profondeur et entraîn\'es en utilisant diff\'erentes tailles d'ensembles d'entraînement -- \`a r\'esoudre chacun des probl\`emes SVRT. Nous avons trouv\'e une gamme de précision sur l'ensemble des vingt-trois tâches. Des r\'eseaux peu profonds ont facilement appris certaines tâches, et des ensembles d'entraînement relativement petits et certaines tâches ont \'et\'e difficilement r\'esolues avec des r\'eseaux beaucoup plus profonds et des ordres de grandeur plus \'elev\'es d'exemples d'entraînement.

Sous l'hypoth\`ese que la complexit\'e de calcul des tâches individuelles peut \^etre correctement caract\'eris\'ee par le motifs de pr\'ecision des tests de ces $N=15$ r\'eseaux neuronaux, nous avons form\'e des vecteurs de pr\'ecision \`a N dimensions pour chaque tâche et ex\'ecut\'e un algorithme de regroupement hi\'erarchique. L'analyse en r\'esultant \textcolor{black}{sugg\`ere} une taxonomie des tâches de raisonnement visuel: au-del\`a de deux clusters primaires correspondant aux jugements \textit{identique-diff\'erent} (SD) vs \textit{relation spatiale} (SR), nous avons \'egalement identifi\'e une organisation plus fine avec des sous-groupes refl\'etant la nature et le nombre de relations utilis\'ees pour composer les r\`egles d\'efinissant la tâche. Nos r\'esultats sont coh\'erents avec les travaux ant\'erieurs de \citet{kim2018not}, qui a \'et\'e le premier \`a identifier une dichotomie entre les tâches SD et SR. Nos r\'esultats prolongent \'egalement les travaux ant\'erieurs de \citep{fleuret2011comparing,kim2018not,yihe2019program} en \textcolor{black}{proposant} une taxonomie plus fine des tâches de raisonnement visuel. La pr\'ecision des r\'eseaux neuronaux se refl\`ete dans le nombre de relations utilis\'ees pour d\'efinir les r\`egles de base, ce qui est attendu, mais m\'erite un examen plus approfondi.

\color{black}
\citet{kim2018not} ont pr\'ec\'edemment sugg\'er\'e que les tâches de SD "tendent" les r\'eseaux neuronaux convolutifs. C'est-\`a-dire que, bien qu'il soit possible de trouver une architecture de r\'eseau d'une profondeur suffisante (ou un nombre d'unit\'es suffisant) qui de plus, peut r\'esoudre une version de la tâche avec une certaine configuration de stimuli (par ex, en forçant tous les stimuli \`a \^etre contenus dans une fen\^etre $\Delta H \times \Delta W$), il est aussi relativement facile de rendre la m\^eme tâche non apprenable par cette m\^eme architecture de r\'eseau \textcolor{black}{passé} un certain nombre de stimuli de configuration (par exemple, en augmentant la taille de la fen\^etre qui contient tous les stimuli). Tout se passe comme si ces r\'eseaux convolutifs \'etaient capables d'apprendre la tâche quand le nombre de stimuli reste inf\'erieur \`a leurs capacit\'es de m\'emoire, et \'echouent au-del\`a. La question de savoir si des alternatives non convolutionnelles aux réseaux de neurones convolutionnels (CNNs) test\'es ici, comme les d\'esormais populaires r\'eseaux \textit{Transformer}  \citep{Dosovitskiy2020-iq,touvron2021training,tolstikhin2021mlp}, pourraient apprendre \`a r\'esoudre plus efficacement les tâches SVRT les plus difficiles reste encore une question ouverte. Comme exp\'erience initiale, nous avons tent\'e d'entraîner et de tester un \textit{Transformer} visuel (ViT) \footnote[1]{\url{https://github.com/facebookresearch/dino}}  \citep{Dosovitskiy2020-iq} contraint d'avoir un nombre de param\`etres (21M) égal au nombre de paramètres du modèle ResNet-50 utilis\'e ici. Pour la plupart des tâches SVRT difficiles, nous n'avons pas réussi \`a obtenir avec ces transformeurs visuels des r\'esultats au-delà de ceux obtenus par les réseaux de type ResNet et ce m\^eme en utilisant 100 000 \'echantillons (le même résultat a été montré \'egalement dans \citet{messina2021recurrent}). Il convient de noter qu'un jeu de donn\'ees de 100 000 \'echantillons reste relativement petit par rapport aux standards actuels en vision, puisque ViT a été entraîné \`a partir de z\'ero.

On peut d\'emontrer que sous certaines contraintes architecturales, les perceptrons multicouches et les r\'eseaux neuronaux convolutifs, y compris les réseaux ResNets ainsi que d'autres architectures, sont des approximateurs universels. En d'autres termes, ces réseaux peuvent apprendre des correspondances arbitraires entre des images et des labels de classe. En fonction de la complexit\'e de la correspondance, on peut avoir besoin d'un nombre croissant de neurones cach\'es afin de permettre une expressivit\'e suffisante du r\'eseau ; mais si l'on dispose de suffisamment de profondeur et d'une quantit\'e suffisante d'exemples d'entraînement, les CNNs profonds peuvent apprendre des tâches de raisonnement visuel arbitraires. Bien que nous ne puissions pas spécifiquement faire d'affirmation forte pour les architectures ResNet utilis\'ees dans cette \'etude (car la preuve d'approximation universelle a été faite dans le cadre d'une seule couche sans max pooling ni batch normalization \citep{lin2018resnet}), nous avons constat\'e empiriquement que toutes les tâches SVRT pouvaient effectivement \^etre apprises pour des r\'eseaux de profondeur suffisante et \`a condition d'avoir une quantit\'e suffisante d'exemples d'entraînement.
Cependant, les CNNs profonds sont g\'en\'eralement d\'epourvus de nombreuses fonctions cognitives humaines, telles que l'attention et la m\'emoire de travail. De telles fonctions sont susceptibles de fournir un avantage considérable \`a un apprenant pour r\'esoudre certaines de ces tâches \citep{10.7551/mitpress/1187.001.0001}. Au lieu de ces fonctions cognitives, les CNNs s'appuyeraient sur leur aspect d'approximateurs universels, conduisant \`a une solution de type "force brute" qui serait moins g\'en\'erale. 
Dans ces conditions, une question ouverte est de savoir si les tâches de SVRT d\'eriv\'es de nos analyses bas\'ees sur les CNNs serait effectivement des tâches valables pour une \'etude sur des sujets humains. De plus, la pr\'ediction faite par \citet{kim2018not} \`a l'aide des CNNs et selon laquelle les tâches SD sont plus difficiles à résoudre que les tâches SR et donc qu'elles peuvent exiger des calculs suppl\'ementaires (par le biais de processus de feedback) tels que l'attention et/ou la m\'emoire de travail a \'et\'e valid\'ee exp\'erimentalement avec succ\`es par \citet{AlamiaENEURO.0267-20.2020} en utilisant des données EEG.

\color{black}
Une preuve suppl\'ementaire des avantages des m\'ecanismes de r\'etroaction pour le raisonnement visuel a \'et\'e fournie par \citet{linsley2018learning} qui a montr\'e que les tâches de traçage de contours qui peuvent \^etre r\'esolues efficacement avec une seule couche d'un CNN r\'ecurrent peuvent n\'ecessiter plusieurs ordres de grandeur d'\'etapes de traitement suppl\'ementaires dans un CNN non r\'ecurrent pour r\'esoudre la m\^eme tâche. Cela se traduit en fin de compte par une efficacit\'e d'\'echantillonnage beaucoup plus grande pour les CNN r\'ecurrents sur les tâches de segmentation d'images naturelles \citep{linsley2020recurrent}. \textcolor{black}{La tâche \'etroitement li\'ee de l'inutilit\'e a \'egalement \'et\'e \'etudi\'ee par \citet{villalobos2021neural} qui a d\'emontr\'e l'incapacit\'e des CNN \`a apprendre une solution g\'en\'erale pour cette classe de probl\`emes.}
Les approximateurs universels avec des biais inductifs minimaux tels que les perceptrons multicouches, les CNN et d'autres architectures feedforward ou \textcolor{black}{non}-attentives peuvent apprendre \`a r\'esoudre des tâches de raisonnement visuel, mais ils peuvent avoir besoin d'un tr\`es grand nombre d'exemples d'entraînement pour s'adapter correctement. Par cons\'equent, au-del\`a de la simple mesure de la pr\'ecision de r\'eseaux tr\`es profonds dans des r\'egimes de donn\'ees \'elev\'es (comme lorsque des millions d'exemples d'entraînement sont disponibles), l'\'evaluation syst\'ematique des performances de r\'eseaux neuronaux de diff\'erentes profondeurs et pour diff\'erents r\'egimes d'entraînement peut fournir des informations essentielles sur la complexit\'e de diff\'erentes tasks de raisonnement visuel.

Plus tôt, \citet{kim2018not} a \'emis l'hypoth\`ese que cette tension des r\'eseaux convolutifs est due \`a leur manque de m\'ecanismes d'attention permettant de lier explicitement les r\'egions de l'image aux objets mentaux. Une remarque similaire a \'et\'e faite par \citet{greff2020binding} dans le contexte de l'incapacit\'e des r\'eseaux neuronaux contemporains \`a d\'ecouper les informations sensorielles en morceaux distincts qui peuvent ensuite \^etre analys\'es et compar\'es individuellement (voir \'egalement \citet{10.1007/978-3-540-75555-5_15} pour un point similaire). Il est int\'eressant de noter que cette pr\'ediction a r\'ecemment \'et\'e test\'ee \`a l'aide de l'EEG humain par \citet{AlamiaENEURO.0267-20.2020} qui a montr\'e qu'en effet l'activit\'e c\'er\'ebrale enregistr\'ee pendant les tâches SD est compatible avec des demandes d'attention et de m\'emoire de travail plus importantes que les tasks SR. En m\^eme temps, le fait que les CNN puissent apprendre des tasks SR plus efficacement que des tasks SD ne signifie pas n\'ecessairement que les participants humains peuvent r\'esoudre ces tâches sans attention. En effet, \citep{logan1994spatial} a montr\'e que les tâches SR telles que le jugement de l'int\'erieur n\'ecessitent de l'attention dans certaines circonstances.

Pour \'evaluer le rôle de l'attention dans le raisonnement visuel, nous avons utilis\'e des modules \textit{Transformer} pour doter les CNNs profonds d'une attention spatiale et d'une attention bas\'ee sur les caract\'eristiques. Les am\'eliorations relatives obtenues par les CNNs avec les deux formes d'attention varient selon les tasks. De nombreuses tasks refl\'etaient une plus grande am\'elioration de l'attention spatiale, et un plus petit nombre b\'en\'eficiait de l'attention bas\'ee sur les caract\'eristiques. De plus, nous avons constat\'e que les mod\`eles d'am\'eliorations relatives expliquaient une grande partie de la variance dans l'espace des tasks SVRT d\'eriv\'ees dans l'exp\'erience 1. Dans l'ensemble, nous avons constat\'e que l'exigence d'une attention spatiale et d'une attention bas\'ee sur les caract\'eristiques explique bien la taxonomie des tasks de raisonnement visuel identifi\'ees dans l'exp\'erience 1.  Notre analyse informatique a \'egalement conduit \`a des pr\'edictions testables pour les exp\'eriences humaines en sugg\'erant des tasks qui b\'en\'eficient soit de l'attention spatiale (tâche \textit{22}) soit de l'attention bas\'ee sur les caract\'eristiques (tâche \textit{21}), des tâches qui b\'en\'eficient des deux formes d'attention (tâche \textit{19}) et des tâches qui ne b\'en\'eficient pas de l'attention (tâche\textit{2}).

\color{black}
Enfin, notre \'etude s'est concentr\'ee sur les avantages computationnels de l'attention spatiale et de l'attention bas\'ee sur les caract\'eristiques pour le raisonnement visuel. Les travaux futurs devraient consid\'erer le rôle d'autres formes d'attention, y compris l'attention bas\'ee sur l'objet \citep{egly1994covert} pour le raisonnement visuel.

Dans notre deuxi\`eme exp\'erience, nous avons \'etudi\'e la capacit\'e d'apprentissage des caract\'eristiques SVRT par rapport aux r\`egles. Pour ce faire, nous avons pr\'eform\'e les r\'eseaux neuronaux sur des tasks auxiliaires afin d'apprendre les caract\'eristiques SVRT avant de les former \`a apprendre les r\`egles abstraites associ\'ees aux probl\`emes SVRT individuels. Nos m\'ethodes de pr\'e-entraînement ont conduit \`a des r\'eseaux qui apprennent \`a r\'esoudre les probl\`emes SVRT mieux que les r\'eseaux form\'es \`a partir de z\'ero, ainsi que les r\'eseaux qui ont \'et\'e pr\'e-entraîn\'es pour effectuer la cat\'egorisation d'images sur le jeu de donn\'ees ImageNet. Nous avons \'egalement constat\'e que ces processus d'attention semblent contribuer davantage \`a l'apprentissage de r\`egles qu'\`a l'apprentissage de caract\'eristiques. Pour le sous-cluster $SR_1$, nous constatons que ce type de pr\'e-entraînement est avantageux dans les r\'egimes d'entraînement inf\'erieurs, mais que les avantages disparaissent rapidement dans les r\'egimes d'entraînement sup\'erieurs. En revanche, ce pr\'e-entraînement ne permet pas d'apprendre les tasks du sous-cluster $SD_1$, m\^eme avec 15 000 \'echantillons, ce qui sugg\`ere que le principal d\'efi de ces tasks n'est pas de d\'ecouvrir de bonnes repr\'esentations visuelles, mais plutôt de d\'ecouvrir la r\`egle. Cela sugg\`ere le besoin de m\'ecanismes suppl\'ementaires au-del\`a de ceux mis en œuvre dans les ResNets. Ceci est \'egalement coh\'erent avec les am\'eliorations observ\'ees pour ces tasks avec l'ajout de m\'ecanismes d'attention. 

En r\'esum\'e, notre \'etude a compar\'e les exigences informatiques de diff\'erentes tasks de raisonnement visuel. Bien que nous nous soyons concentr\'es sur la compr\'ehension des avantages computationnels des m\'ecanismes d'attention et d'apprentissage des caract\'eristiques, il est clair que des m\'ecanismes suppl\'ementaires seront n\'ecessaires pour r\'esoudre pleinement toutes les tasks de SVRT. Ces m\'ecanismes sont susceptibles d'inclure la m\'emoire de travail, dont on sait qu'elle joue un rôle dans les tasks de SVRT \citep{AlamiaENEURO.0267-20.2020}. Dans l'ensemble, ce travail illustre les avantages potentiels de l'incorporation de m\'ecanismes semblables \`a ceux du cerveau dans les r\'eseaux neuronaux modernes et fournit une voie \`a suivre pour atteindre un raisonnement visuel de niveau humain. 

Dans la derni\`ere partie de la th\`ese, nous avons propos\'e une nouvelle architecture, le Guided Attention Model for (visual) Reasoning (\textit{GAMR}). Nous avons int\'egr\'e les deux capacit\'es cognitives utilis\'ees par les humains - l'attention et la m\'emoire - dans la r\'esolution de tasks de raisonnement. Les th\'eories cognitives modernes de la vision active postulent que le syst\`eme visuel explore l'environnement de mani\`ere dynamique par le biais de s\'equences de changements d'attention pour s\'electionner et acheminer vers la m\'emoire les informations pertinentes pour la task \citep{ullman1984massachusetts,ullman1987visual}. Des exp\'eriences de psychophysique \citep{hayhoe2000vision} sur l'attention visuelle manifeste ont montr\'e que les sch\'emas de mouvements oculaires sont dirig\'es selon des routines d\'ependantes de la task. Le GAMR s'inspire de la litt\'erature des sciences cognitives sur la vision active, où l'attention manifeste est dirig\'ee dans le syst\`eme visuel pour recueillir des informations pertinentes pour la task. Selon la th\'eorie de la vision active, le monde visuel est explor\'e \`a l'aide de mouvements oculaires rapides guid\'es par des d\'eplacements de l'attention visuelle.

\color{black}
Nous avons conçu un contrôleur semblable aux m\'ecanismes impliqu\'es dans le cadre de la vision active pour diriger le projecteur d'attention et envoyer les repr\'esentations pertinentes pour la task dans le bloc de m\'emoire utilis\'e ensuite pour le raisonnement. Dans GAMR, le contrôleur est impl\'ement\'e avec la couche d'encodeur bas\'ee sur les transformateurs. Contrairement \`a la m\'ethode existante où l'auto-attention est suivie d'une addition et d'une normalisation de la couche et où une couche lin\'eaire est ajout\'ee \`a la fin de cette op\'eration, nous supprimons la couche d'auto-attention et la remplaçons par l'attention bas\'ee sur les caract\'eristiques obtenue avec le module contrôleur LSTM. Cela aide le contrôleur \`a d\'eplacer le point d'attention. 
 
Notre proposition de module d'attention guid\'ee pour le raisonnement (visuel) (GAMR) apprend \`a d\'eplacer l'attention dynamiquement, en fonction de la task, sur la base de requ\^etes g\'en\'er\'ees en interne par un contrôleur ex\'ecutif LSTM. Grâce \`a des exp\'eriences approfondies sur les deux principaux d\'efis de raisonnement visuel, le Synthetic Visual Reasoning Test (SVRT)~\citep{fleuret2011comparing} et le Abstract Reasoning Task (ART)~\citep{Webb2021EmergentST}, nous d\'emontrons que notre architecture neuronale est capable d'apprendre des compositions complexes de r\`egles relationnelles d'une mani\`ere efficace en termes de donn\'ees et qu'elle est plus performante que les autres architectures neuronales de pointe pour le raisonnement visuel. En utilisant des m\'ethodes d'explicabilit\'e, nous caract\'erisons davantage les strat\'egies visuelles utilis\'ees par le mod\`ele afin de r\'esoudre des tasks de raisonnement repr\'esentatives. Nous d\'emontrons que notre mod\`ele est compositionnel, c'est-\`a-dire qu'il est capable de se g\'en\'eraliser efficacement \`a de nouvelles tasks et d'apprendre de nouvelles routines visuelles en recomposant des op\'erations \'el\'ementaires apprises pr\'ec\'edemment. 

Nous nous sommes inspir\'es pour la banque de m\'emoire de \citet{Webb2021EmergentST}, où des m\'ecanismes de liaison et d'indirection de variables ont \'et\'e introduits dans l'architecture pour le raisonnement visuel \`a l'aide d'une m\'emoire externe. La liaison de variables est la capacit\'e de lier deux repr\'esentations, et l'indirection est le m\'ecanisme impliqu\'e dans la r\'ecup\'eration d'une repr\'esentation pour se r\'ef\'erer \`a l'autre. Ces auteurs introduisent \'egalement la normalisation temporelle du contexte (TCN) \citep{webb2020learning}, qui s'av\`ere b\'en\'efique pour la g\'en\'eralisation hors distribution pour les tasks de raisonnement relationnel. Cependant, le mod\`ele pr\'esente des limites importantes: Il suppose une repr\'esentation d'image centr\'ee sur l'objet, où les objets sont pr\'esent\'es individuellement dans une s\'equence. Nous ne pouvons pas \'evaluer une telle architecture dans le cadre du d\'efi SVRT, car les images de chaque task contiennent plusieurs objets qui n\'ecessitent une individualisation. Il existe \'egalement certaines relations, comme "toucher", que cette individuation ne peut repr\'esenter (\textcolor{black}{ou toute architecture centr\'ee sur l'objet}). ESBN n'a pas non plus de m\'ecanisme attentionnel et fonctionne mieux dans un sc\'enario où une attention forte au niveau du pr\'e-traitement permet de simplifier les tasks. Nous avons test\'e ce comportement de correspondance des mod\`eles de l'architecture en l'entraînant en pr\'esence d'un bruit gaussien. Il a conduit \`a une performance de niveau al\'eatoire. Ici, nous nous appuyons sur ce travail et d\'ecrivons un mod\`ele entraînable de bout en bout qui apprend \`a individualiser les sc\`enes pertinentes pour la task et \`a stocker leurs repr\'esentations en m\'emoire pour permettre de juger des relations complexes entre ces objets. Enfin, notre m\'ecanisme relationnel est inspir\'e du travail de \citet{santoro2017simple} qui a introduit un mod\`ele pr\^et \`a l'emploi pour calculer les relations entre les repr\'esentations de type objet dans un r\'eseau.

\color{black}

L'une des limites de l'approche actuelle est le nombre fixe de pas de temps. Je pense qu'une future continuation de ce travail pourrait \^etre d'incorporer un m\'ecanisme pour adapter le nombre de pas de temps en fonction de la complexit\'e de la task. Pour l'instant, nous avons fix\'e le nombre de pas de temps \`a quatre pour toutes les tasks ; cependant, une task simple pourrait n\'ecessiter moins de pas de temps pour arriver \`a une d\'ecision avec une grande confiance. Pour que le mod\`ele s'adapte \`a la situation, une possibilit\'e pourrait \^etre de l'entraîner avec une variable de confiance comme crit\`ere d'arr\^et. 

Bien que nous ayons limit\'e notre analyse \`a des ensembles de donn\'ees synth\'etiques de raisonnement visuel, il serait possible \`a l'avenir de tester les mod\`eles sur un ensemble de donn\'ees du monde r\'eel comme V-PROM. Il s'agit d'images organis\'ees dans le style de raisonnement des Ravens, avec des images de contexte et des images de choix parmi lesquelles la bonne r\'eponse est s\'electionn\'ee. Une autre direction possible est de penser \`a une architecture qui prend en compte deux caract\'eristiques importantes: l'utilisation efficace des donn\'ees et l'utilisation efficace des ressources informatiques. Une façon de concevoir cette architecture est d'incorporer un m\'ecanisme de lecture et d'\'ecriture similaire \`a une machine de Turing neuronale. Ces deux m\'ecanismes aideront le r\'eseau \`a lire les relations d\'ej\`a stock\'ees en m\'emoire et \`a les \'ecrire dans la m\'emoire si elles sont nouvelles. Nous nous attendons \`a ce qu'une telle architecture cognitive d\'emontre une capacit\'e de raisonnement d'ordre sup\'erieur, d'apprentissage continu, de composition et de m\'eta-apprentissage.

Nous avons \'egalement \'evalu\'e ViT~\citep{Dosovitskiy2020-iq} -- Nous avons \'egalement \'evalu\'e ViT~citep{Dosovitskiy2020-iq} - une architecture d'auto-attention compl\`ete sur des tasks SVRT et avons constat\'e qu'elle avait du mal \`a apprendre les tasks SVRT les plus simples ; cependant, \citet{messina2021recurrent} a men\'e une \'etude similaire sur un sous-ensemble plus petit de quatre tasks SVRT entra\^in\'ees sur 28k \'echantillons et a constat\'e qu'une version r\'ecurrente de ViT - un r\'eseau attentionnel avec une colonne vert\'ebrale convolutive peut apprendre ces tasks. L'ajout de convolutions dans les premi\`eres couches de ViT permet d'obtenir une meilleure pr\'ecision et am\'eliore la sensibilit\'e aux param\`etres d'optimisation \citep{xiao2021early}. Cette observation nous a motiv\'e \`a proposer \textit{Conviformer} \citep{vaishnav2022conviformers} pour un autre projet collaboratif sur la classification des feuilles-fossiles. Nous proposons un r\'eseau pour incorporer un r\'eseau convolutif comme frontal d'un r\'eseau de transformation de la vision bas\'e sur l'auto-attention compl\`ete, am\'eliorant ainsi sa capacit\'e \`a traiter des images \`a plus haute r\'esolution. Bien que les images de plus grande taille rev\^etent une grande importance dans les applications de vision par ordinateur telles que la d\'etection d'objets, la segmentation et la classification \`a grain fin, elles ne peuvent pas \^etre utilis\'ees avec les transformateurs de vision en raison de la demande de m\'emoire de calcul associ\'ee. \textit{Conviformer} am\'eliore la performance des transformateurs de vision en incorporant des caract\'eristiques locales et en infusant des prieurs convolutifs dans une architecture de transformateur. Nous aimerions voir comment les transformateurs de vision induits par convolution se comportent dans les tasks SVRT. 

L'apprentissage de concepts est une autre direction de recherche passionnante. L'une des principales caract\`eristiques de l'intelligence humaine est la capacit\'e d'apprendre rapidement de nouveaux concepts et de les utiliser pour g\'en\'eraliser \`a un nouveau sc\'enario. Un \textit{concept} peut \^etre une id\'ee repr\'esentant une classe d'\'ev\'enements (par exemple, la marche), des objets (par exemple, les chats), ou leurs propri\'et\'es (par exemple, la couleur bleue). Pour tester la capacit\'e d'apprentissage de concepts des r\'eseaux neuronaux en quelques coups, nous avons r\'ecemment introduit un nouveau jeu de donn\'ees de raisonnement visuel, Compositional Visual Reasoning (\textit{CVR})~\citep{zerroug2022benchmark}. Ce jeu de donn\'ees est bas\'e sur le principe du raisonnement odd-one-out. Dans cette forme de task de raisonnement, trois \'echantillons sur quatre suivent un concept similaire (r\`egle) dans leur formulation, tandis que le quatri\`eme ne le fait pas. Chaque \'echantillon contient des formes similaires aux formes utilis\'ees dans le d\'efi SVRT. Cela permet d'\'etendre la vari\'et\'e des relations utilis\'ees dans la formulation par rapport aux jeux de donn\'ees pr\'ec\'edemment d\'efinis comme SVRT ou RPM. Nous avons \'egalement inclus la compositionnalit\'e pr\'ealable dans le jeu de donn\'ees, où certaines relations \'el\'ementaires sont utilis\'ees pour composer les diff\'erentes tasks. La motivation est de pousser la communaut\'e \`a construire un r\'eseau compositionnel et efficace en termes d'\'echantillons.

Dans cette th\`ese, nous avons fait l'une des toutes premi\`eres tentatives d'explorer l'auto-attention du point de vue du raisonnement visuel. L'attention joue un rôle crucial dans la d\'emonstration des capacit\'es de raisonnement visuel, et on s'attend \`a ce qu'un meilleur mod\`ele attentionnel soit meilleur pour le raisonnement. Nous avons montr\'e comment les op\'erations d'auto-attention pouvaient \^etre utilis\'ees comme mod\`ele computationnel d'un syst\`eme d'attention visuelle repr\'esentant l'attention spatiale et bas\'ee sur les caract\'eristiques, ainsi que comme mod\`ele de vision active. Bien que nous ayons constat\'e que l'auto-attention est aussi efficace dans la r\'esolution de tasks de raisonnement que dans d'autres d\'efis li\'es \`a la vision, il est n\'ecessaire d'effectuer des analyses suppl\'ementaires pour comprendre les m\'ecanismes fondamentaux d'un mod\`ele complet d'auto-attention qui limitent son apprentissage efficace par \'echantillonnage pour les tasks de raisonnement. Dans l'ensemble, ce travail d\'emontre les avantages potentiels de l'ajout de m\'ecanismes d'auto-attention dans l'architecture cognitive et de vision par ordinateur pour r\'esoudre les tasks de raisonnement visuel.

\part*{Appendix}
\addcontentsline{toc}{part}{Appendix}
\appendix
\renewcommand\chaptername{Appendix}
\chapter{Synthetic Visual Reasoning Task}

\renewcommand{\thefigure}{A\arabic{figure}}
\renewcommand{\thesection}{A\arabic{section}}
\setcounter{figure}{0} 
\setcounter{section}{0} 

\startcontents[chapters]

\begin{figure*}[t]
\centering
  \includegraphics[width=1\linewidth]{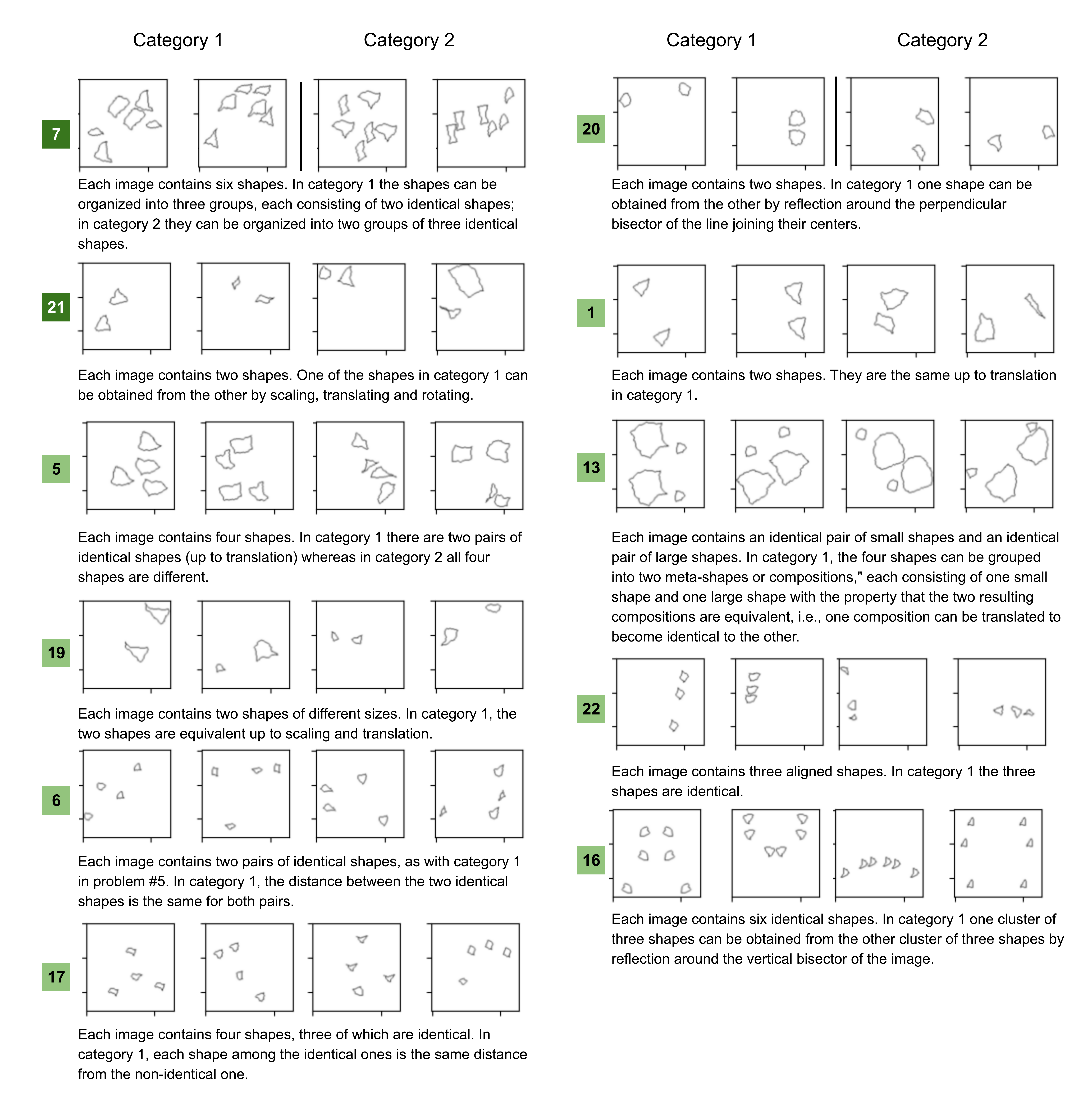}
 \caption{Sample images for Same Different (SD) tasks }\label{fig:exampleSD}
\end{figure*}

\begin{figure*}[t]
\centering
  \includegraphics[width=1\linewidth]{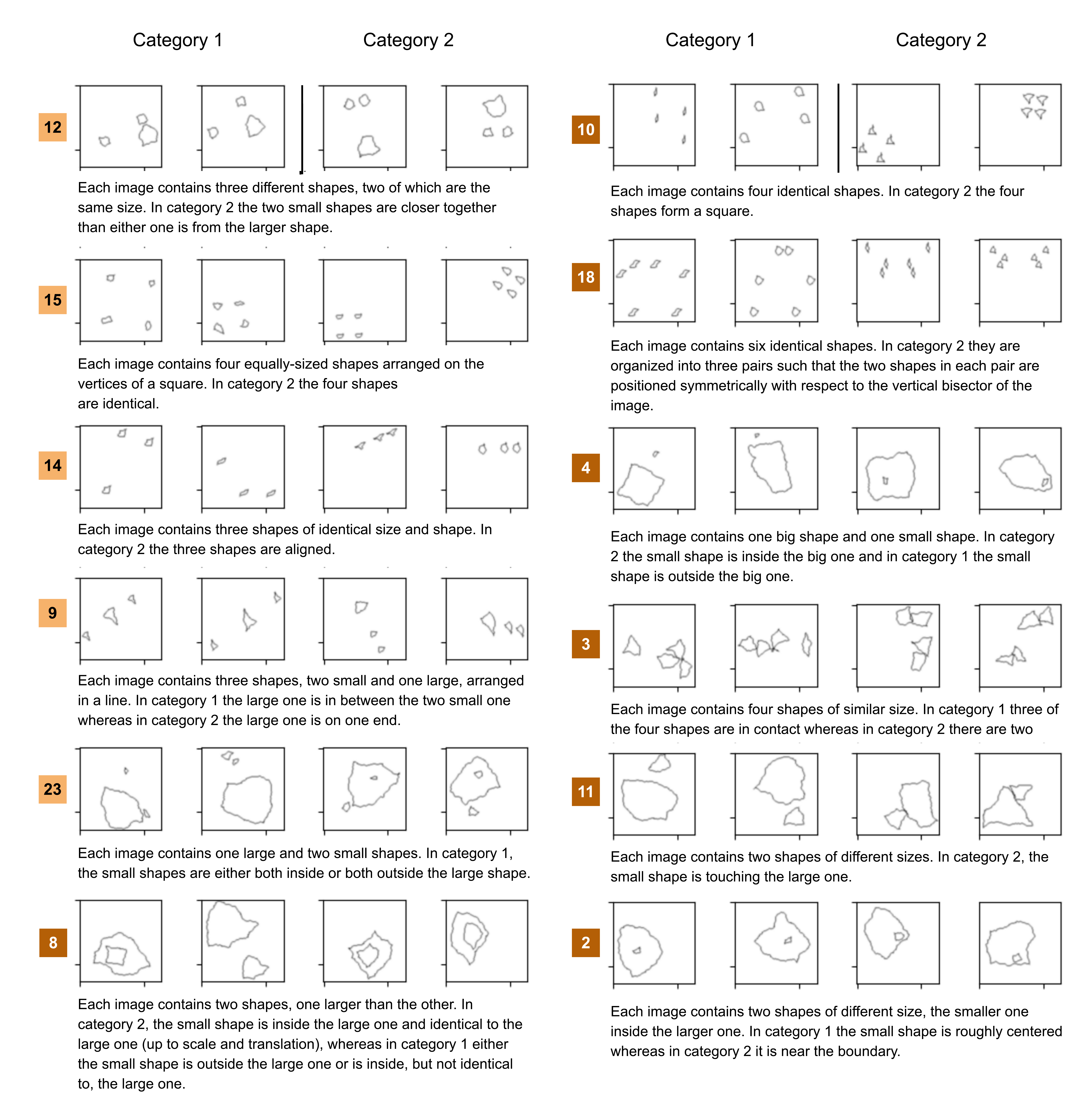}
 \caption{Sample images for Spatial Relation (SR) tasks }\label{fig:exampleSR}
\end{figure*}

\begin{table}[t]
\centering
\caption{Each cell represents attempts participants took to solve seven consecutive correct categorizations. Here, row and column represents $task~ number$ and $participant ~ number$. Entries containing "X" indicate that the participant failed to solve the problem, and those cells are not included in the marginal means. \citep{fleuret2011comparing}}
\vspace{5mm}

\small
\begin{adjustbox}{width=\textwidth}
\begin{tabular}{|c|llllllllllllllllllll|c|c|}
\hline 
\multicolumn{1}{|l|}{} & \multicolumn{20}{c}{\textbf{Participant No.}}  & \multicolumn{1}{|l|}{}   & \multicolumn{1}{|l|}{}  \\ \cline{2-21}
\textbf{Task No.}    & \textbf{1}             & \textbf{2}                                 & \textbf{3}                                 & \textbf{4}                                 & \textbf{5}                                 & \textbf{6}                                 & \textbf{7}             & \textbf{8}                                 & \textbf{9}                                 & \textbf{10}                                & \textbf{11}                                & \textbf{12}                                & \textbf{13}                                & \textbf{14}                                & \textbf{15}                                & \textbf{16}                                & \textbf{17}                                & \textbf{18}                                & \textbf{19}                                & \textbf{20}                                & \multicolumn{1}{|l|}{\multirow{-2}{*}{\textbf{Mean}}} & \multicolumn{1}{|l|}{\multirow{-2}{*}{\textbf{Fail}}} \\ \hline
\textbf{1}           & {\color{blue} 1}  & {\color{blue} 12}                     & {\color{blue} 1}                      & {\color{blue} 2}                      & {\color{blue} 8}                      & {\color{blue} 8}                      & {\color{blue} 1}  & {\color{blue} 1}                      & {\color{red} X} & {\color{blue} 1}                      & {\color{blue} 14}                     & {\color{blue} 1}                      & {\color{blue} 4}                      & {\color{blue} 1}                      & {\color{blue} 1}                      & {\color{blue} 1}                      & {\color{blue} 2}                      & {\color{blue} 1}                      & {\color{blue} 1}                      & {\color{blue} 1}                      & 3.26                                                & 1                                                   \\
\textbf{2}           & {\color{blue} 3}  & {\color{blue} 1}                      & {\color{blue} 2}                      & {\color{blue} 2}                      & {\color{blue} 10}                     & {\color{blue} 19}                     & {\color{blue} 4}  & {\color{blue} 4}                      & {\color{blue} 14}                     & {\color{blue} 3}                      & {\color{blue} 2}                      & {\color{blue} 3}                      & {\color{blue} 21}                     & {\color{blue} 1}                      & {\color{blue} 1}                      & {\color{blue} 5}                      & {\color{blue} 3}                      & {\color{blue} 2}                      & {\color{blue} 22}                     & {\color{blue} 9}                      & 6.55                                                & 0                                                   \\
\textbf{3}           & {\color{blue} 7}  & {\color{blue} 1}                      & {\color{blue} 3}                      & {\color{blue} 1}                      & {\color{blue} 4}                      & {\color{blue} 3}                      & {\color{blue} 1}  & {\color{blue} 1}                      & {\color{blue} 7}                      & {\color{blue} 1}                      & {\color{blue} 6}                      & {\color{blue} 1}                      & {\color{blue} 1}                      & {\color{blue} 1}                      & {\color{blue} 4}                      & {\color{blue} 1}                      & {\color{blue} 1}                      & {\color{blue} 1}                      & {\color{blue} 4}                      & {\color{blue} 2}                      & 2.55                                                & 0                                                   \\
\textbf{4}           & {\color{blue} 1}  & {\color{blue} 6}                      & {\color{blue} 7}                      & {\color{blue} 1}                      & {\color{blue} 1}                      & {\color{blue} 3}                      & {\color{blue} 1}  & {\color{blue} 1}                      & {\color{blue} 1}                      & {\color{blue} 1}                      & {\color{blue} 3}                      & {\color{blue} 1}                      & {\color{blue} 1}                      & {\color{blue} 2}                      & {\color{blue} 1}                      & {\color{blue} 1}                      & {\color{blue} 7}                      & {\color{blue} 5}                      & {\color{blue} 7}                      & {\color{blue} 1}                      & 2.6                                                 & 0                                                   \\
\textbf{5}           & {\color{blue} 7}  & {\color{red} X} & {\color{blue} 1}                      & {\color{blue} 21}                     & {\color{blue} 8}                      & {\color{blue} 3}                      & {\color{blue} 1}  & {\color{blue} 5}                      & {\color{red} X} & {\color{blue} 1}                      & {\color{red} X} & {\color{blue} 9}                      & {\color{blue} 13}                     & {\color{blue} 1}                      & {\color{blue} 6}                      & {\color{blue} 2}                      & {\color{red} X} & {\color{blue} 8}                      & {\color{blue} 1}                      & {\color{blue} 7}                      & 5.88                                                & 4                                                   \\
\textbf{6}           & {\color{red} X}  & {\color{blue} 20}                     & {\color{red} X} & {\color{red} X} & {\color{blue} 27}                     & {\color{blue} 25}                     & {\color{blue} 12} & {\color{blue} 26}                     & {\color{red} X} & {\color{red} X} & {\color{blue} 3}                      & {\color{red} X} & {\color{red} X} & {\color{red} X} & {\color{blue} 4}                      & {\color{blue} 16}                     & {\color{red} X} & {\color{red} X} & {\color{red} X} & {\color{red} X} & 16.63                                               & 12                                                  \\
\textbf{7}           & {\color{blue} 1}  & {\color{red} X} & {\color{blue} 1}                      & {\color{red} X} & {\color{blue} 13}                     & {\color{blue} 8}                      & {\color{blue} 4}  & {\color{blue} 14}                     & {\color{red} X} & {\color{blue} 3}                      & {\color{blue} 8}                      & {\color{blue} 12}                     & {\color{blue} 7}                      & {\color{red} X} & {\color{blue} 1}                      & {\color{blue} 6}                      & {\color{blue} 1}                      & {\color{blue} 1}                      & {\color{blue} 14}                     & {\color{blue} 9}                      & 6.44                                                & 4                                                   \\
\textbf{8}           & {\color{blue} 7}  & {\color{blue} 6}                      & {\color{blue} 1}                      & {\color{blue} 14}                     & {\color{blue} 4}                      & {\color{blue} 14}                     & {\color{blue} 1}  & {\color{blue} 5}                      & {\color{blue} 1}                      & {\color{blue} 4}                      & {\color{blue} 8}                      & {\color{blue} 1}                      & {\color{blue} 1}                      & {\color{blue} 1}                      & {\color{blue} 13}                     & {\color{blue} 5}                      & {\color{blue} 3}                      & {\color{blue} 7}                      & {\color{blue} 4}                      & {\color{blue} 1}                      & 5.05                                                & 0                                                   \\
\textbf{9}           & {\color{blue} 4}  & {\color{blue} 24}                     & {\color{blue} 1}                      & {\color{blue} 16}                     & {\color{blue} 3}                      & {\color{blue} 1}                      & {\color{blue} 1}  & {\color{blue} 13}                     & {\color{red} X} & {\color{red} X} & {\color{blue} 4}                      & {\color{blue} 6}                      & {\color{red} X} & {\color{blue} 2}                      & {\color{blue} 7}                      & {\color{blue} 1}                      & {\color{blue} 3}                      & {\color{blue} 1}                      & {\color{blue} 5}                      & {\color{blue} 1}                      & 5.47                                                & 3                                                   \\
\textbf{10}          & {\color{blue} 1}  & {\color{blue} 8}                      & {\color{blue} 2}                      & {\color{blue} 2}                      & {\color{blue} 4}                      & {\color{blue} 1}                      & {\color{blue} 3}  & {\color{blue} 5}                      & {\color{red} X} & {\color{blue} 4}                      & {\color{blue} 1}                      & {\color{blue} 2}                      & {\color{blue} 16}                     & {\color{blue} 4}                      & {\color{blue} 4}                      & {\color{blue} 2}                      & {\color{blue} 1}                      & {\color{blue} 1}                      & {\color{blue} 4}                      & {\color{blue} 3}                      & 3.58                                                & 1                                                   \\
\textbf{11}          & {\color{blue} 4}  & {\color{blue} 2}                      & {\color{blue} 3}                      & {\color{blue} 1}                      & {\color{blue} 3}                      & {\color{blue} 1}                      & {\color{blue} 4}  & {\color{blue} 8}                      & {\color{blue} 1}                      & {\color{blue} 2}                      & {\color{blue} 1}                      & {\color{blue} 1}                      & {\color{blue} 1}                      & {\color{blue} 1}                      & {\color{blue} 1}                      & {\color{blue} 5}                      & {\color{blue} 2}                      & {\color{blue} 1}                      & {\color{blue} 1}                      & {\color{blue} 1}                      & 2.2                                                 & 0                                                   \\
\textbf{12}          & {\color{blue} 1}  & {\color{blue} 2}                      & {\color{blue} 8}                      & {\color{blue} 1}                      & {\color{blue} 9}                      & {\color{blue} 4}                      & {\color{blue} 8}  & {\color{blue} 4}                      & {\color{blue} 1}                      & {\color{blue} 7}                      & {\color{blue} 25}                     & {\color{blue} 2}                      & {\color{blue} 5}                      & {\color{blue} 2}                      & {\color{red} X} & {\color{blue} 2}                      & {\color{blue} 5}                      & {\color{red} X} & {\color{blue} 4}                      & {\color{blue} 1}                      & 5.06                                                & 2                                                   \\
\textbf{13}          & {\color{blue} 1}  & {\color{blue} 20}                     & {\color{blue} 5}                      & {\color{blue} 14}                     & {\color{red} X} & {\color{blue} 3}                      & {\color{blue} 1}  & {\color{blue} 13}                     & {\color{blue} 7}                      & {\color{blue} 10}                     & {\color{blue} 1}                      & {\color{blue} 13}                     & {\color{blue} 9}                      & {\color{blue} 5}                      & {\color{red} X} & {\color{blue} 3}                      & {\color{blue} 3}                      & {\color{blue} 2}                      & {\color{red} X} & {\color{blue} 1}                      & 6.53                                                & 3                                                   \\
\textbf{14}          & {\color{blue} 4}  & {\color{blue} 4}                      & {\color{blue} 1}                      & {\color{blue} 1}                      & {\color{blue} 3}                      & {\color{blue} 10}                     & {\color{blue} 2}  & {\color{red} X} & {\color{blue} 12}                     & {\color{blue} 14}                     & {\color{blue} 1}                      & {\color{blue} 19}                     & {\color{blue} 1}                      & {\color{blue} 3}                      & {\color{blue} 1}                      & {\color{blue} 1}                      & {\color{blue} 4}                      & {\color{blue} 8}                      & {\color{blue} 1}                      & {\color{blue} 2}                      & 4.84                                                & 1                                                   \\
\textbf{15}          & {\color{blue} 1}  & {\color{red} X} & {\color{blue} 1}                      & {\color{blue} 2}                      & {\color{blue} 2}                      & {\color{blue} 1}                      & {\color{blue} 1}  & {\color{blue} 1}                      & {\color{red} X} & {\color{blue} 5}                      & {\color{blue} 1}                      & {\color{blue} 2}                      & {\color{blue} 4}                      & {\color{blue} 1}                      & {\color{blue} 1}                      & {\color{blue} 18}                     & {\color{blue} 10}                     & {\color{blue} 3}                      & {\color{blue} 2}                      & {\color{blue} 1}                      & 3.17                                                & 2                                                   \\
\textbf{16}          & {\color{blue} 12} & {\color{blue} 18}                     & {\color{blue} 7}                      & {\color{red} X} & {\color{red} X}                      & {\color{blue} 2}                      & {\color{blue} 2}  & {\color{blue} 14}                     & {\color{red} X} & {\color{red} X} & {\color{blue} 28}                     & {\color{blue} 9}                      & {\color{blue} 13}                     & {\color{red} X} & {\color{blue} 22}                     & {\color{blue} 10}                     & {\color{red} X} & {\color{red} X} & {\color{red} X} & {\color{red} X} & 12.45                                               & 9                                                   \\
\textbf{17}          & {\color{blue} 14} & {\color{red} X} & {\color{blue} 6}                      & {\color{blue} 5}                      & {\color{blue} 2}                      & {\color{red} X} & {\color{blue} 21} & {\color{red} X} & {\color{red} X} & {\color{blue} 22}                     & {\color{red} X} & {\color{blue} 14}                     & {\color{red} X} & {\color{red} X} & {\color{red} X} & {\color{red} X} & {\color{blue} 13}                     & {\color{blue} 8}                      & {\color{blue} 28}                     & {\color{blue} 1}                      & 12.18                                               & 9                                                   \\
\textbf{18}          & {\color{blue} 5}  & {\color{blue} 17}                     & {\color{blue} 2}                      & {\color{red} X} & {\color{blue} 27}                     & {\color{blue} 5}                      & {\color{blue} 5}  & {\color{blue} 1}                      & {\color{red} X} & {\color{blue} 2}                      & {\color{red} X} & {\color{blue} 7}                      & {\color{blue} 19}                     & {\color{blue} 4}                      & {\color{blue} 1}                      & {\color{blue} 1}                      & {\color{blue} 5}                      & {\color{blue} 1}                      & {\color{blue} 1}                      & {\color{blue} 2}                      & 6.18                                                & 3                                                   \\
\textbf{19}          & {\color{blue} 2}  & {\color{blue} 10}                     & {\color{blue} 1}                      & {\color{blue} 11}                     & {\color{blue} 1}                      & {\color{blue} 3}                      & {\color{blue} 5}  & {\color{blue} 11}                     & {\color{blue} 8}                      & {\color{blue} 2}                      & {\color{blue} 4}                      & {\color{blue} 2}                      & {\color{blue} 17}                     & {\color{blue} 1}                      & {\color{blue} 4}                      & {\color{blue} 4}                      & {\color{blue} 1}                      & {\color{blue} 6}                      & {\color{blue} 1}                      & {\color{red} X} & 4.95                                                & 1                                                   \\
\textbf{20}          & {\color{blue} 14} & {\color{blue} 7}                      & {\color{blue} 4}                      & {\color{blue} 5}                      & {\color{blue} 1}                      & {\color{blue} 8}                      & {\color{blue} 3}  & {\color{blue} 1}                      & {\color{red} X} & {\color{blue} 18}                     & {\color{blue} 9}                      & {\color{blue} 16}                     & {\color{blue} 3}                      & {\color{blue} 1}                      & {\color{blue} 6}                      & {\color{blue} 1}                      & {\color{blue} 2}                      & {\color{blue} 1}                      & {\color{blue} 15}                     & {\color{blue} 1}                      & 6.11                                                & 1                                                   \\
\textbf{21}          & {\color{blue} 6}  & {\color{red} X} & {\color{blue} 1}                      & {\color{red} X} & {\color{blue} 1}                      & {\color{red} X} & {\color{blue} 23} & {\color{red} X} & {\color{red} X} & {\color{blue} 21}                     & {\color{blue} 28}                     & {\color{blue} 7}                      & {\color{blue} 26}                     & {\color{blue} 7}                      & {\color{blue} 15}                     & {\color{blue} 2}                      & {\color{blue} 17}                     & {\color{red} X} & {\color{blue} 16}                     & {\color{red} X} & 13.08                                               & 7                                                   \\
\textbf{22}          & {\color{blue} 1}  & {\color{blue} 9}                      & {\color{blue} 14}                     & {\color{blue} 1}                      & {\color{blue} 1}                      & {\color{blue} 4}                      & {\color{blue} 1}  & {\color{blue} 5}                      & {\color{blue} 21}                     & {\color{blue} 2}                      & {\color{blue} 1}                      & {\color{blue} 2}                      & {\color{blue} 5}                      & {\color{blue} 1}                      & {\color{blue} 6}                      & {\color{blue} 1}                      & {\color{blue} 4}                      & {\color{blue} 1}                      & {\color{blue} 1}                      & {\color{blue} 6}                      & 4.35                                                & 0                                                   \\
\textbf{23}          & {\color{blue} 1}  & {\color{blue} 1}                      & {\color{blue} 7}                      & {\color{blue} 22}                     & {\color{blue} 1}                      & {\color{blue} 1}                      & {\color{blue} 2}  & {\color{blue} 1}                      & {\color{blue} 6}                      & {\color{blue} 21}                     & {\color{blue} 2}                      & {\color{blue} 5}                      & {\color{blue} 4}                      & {\color{blue} 6}                      & {\color{blue} 4}                      & {\color{blue} 3}                      & {\color{blue} 1}                      & {\color{blue} 1}                      & {\color{blue} 6}                      & {\color{blue} 8}                      & 5.15                                                & 0                                                   \\ \hline
\textbf{Mean}        & 4.45                   & 9.33                                       & 3.59                                       & 6.78                                       & 6.33                                       & 6.05                                       & 4.65                   & 6.7                                        & 7.18                                       & 7.2                                        & 7.5                                        & 6.14                                       & 8.55                                       & 2.37                                       & 5.15                                       & 4.14                                       & 4.4                                        & 3.11                                       &   6.9                                         &                    3.05                        & \multicolumn{1}{|l}{}                                & \multicolumn{1}{l|}{}                                \\ \hline
\textbf{No of Fails} & 1                      & 5                                          & 1                                          & 5                                          & 2                                          & 2                                          & 0                      & 3                                          & 12                                         & 3                                          & 3                                          & 1                                          & 3                                          & 4                                          & 3                                          & 1                                          & 3                                          & 4                                          &       3                                     &    4                                        & \multicolumn{1}{|l}{}                                & \multicolumn{1}{l|}{}    \\ \hline                        
\end{tabular}
\end{adjustbox}

\label{table:humans}
\end{table}

\chapter[Computational Demands of Visual Reasoning]{Computational Demands of Visual Reasoning}

\renewcommand{\thefigure}{B\arabic{figure}}
\renewcommand{\thesection}{B\arabic{section}}
\setcounter{figure}{0} 
\setcounter{section}{0} 

\startcontents[chapters]

\begin{figure*}[t]
\centering
  \includegraphics[width=1\linewidth]{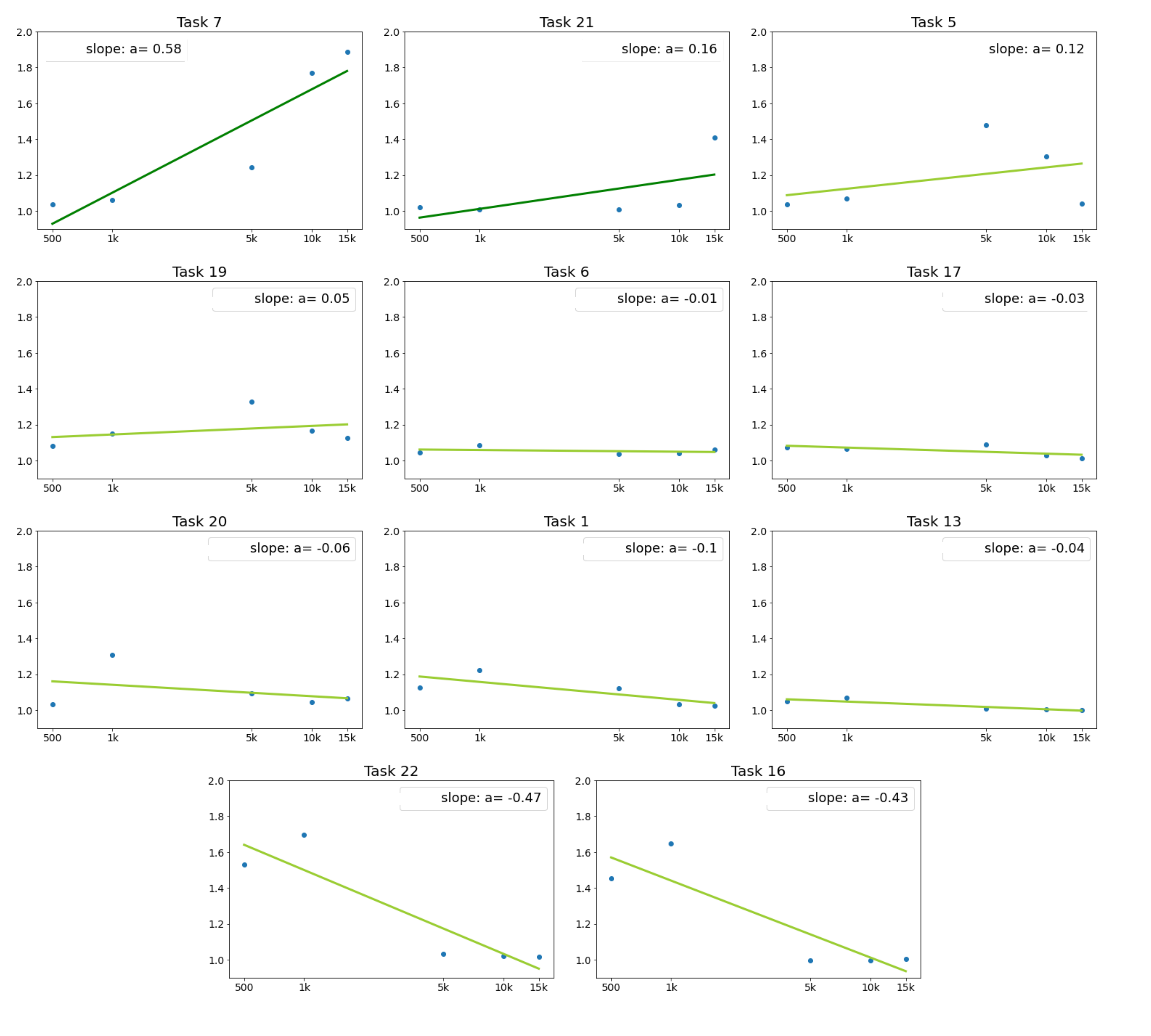}
 \caption{Slope attained by linear fitting of points obtained after taking the ratio of each of the network with spatial attention module and the test accuracy of a ResNet50 for each task and training condition for Same Different (SD) tasks }\label{fig:slope_sa_sd}
\end{figure*}

\begin{figure*}[t]
\centering
  \includegraphics[width=1\linewidth]{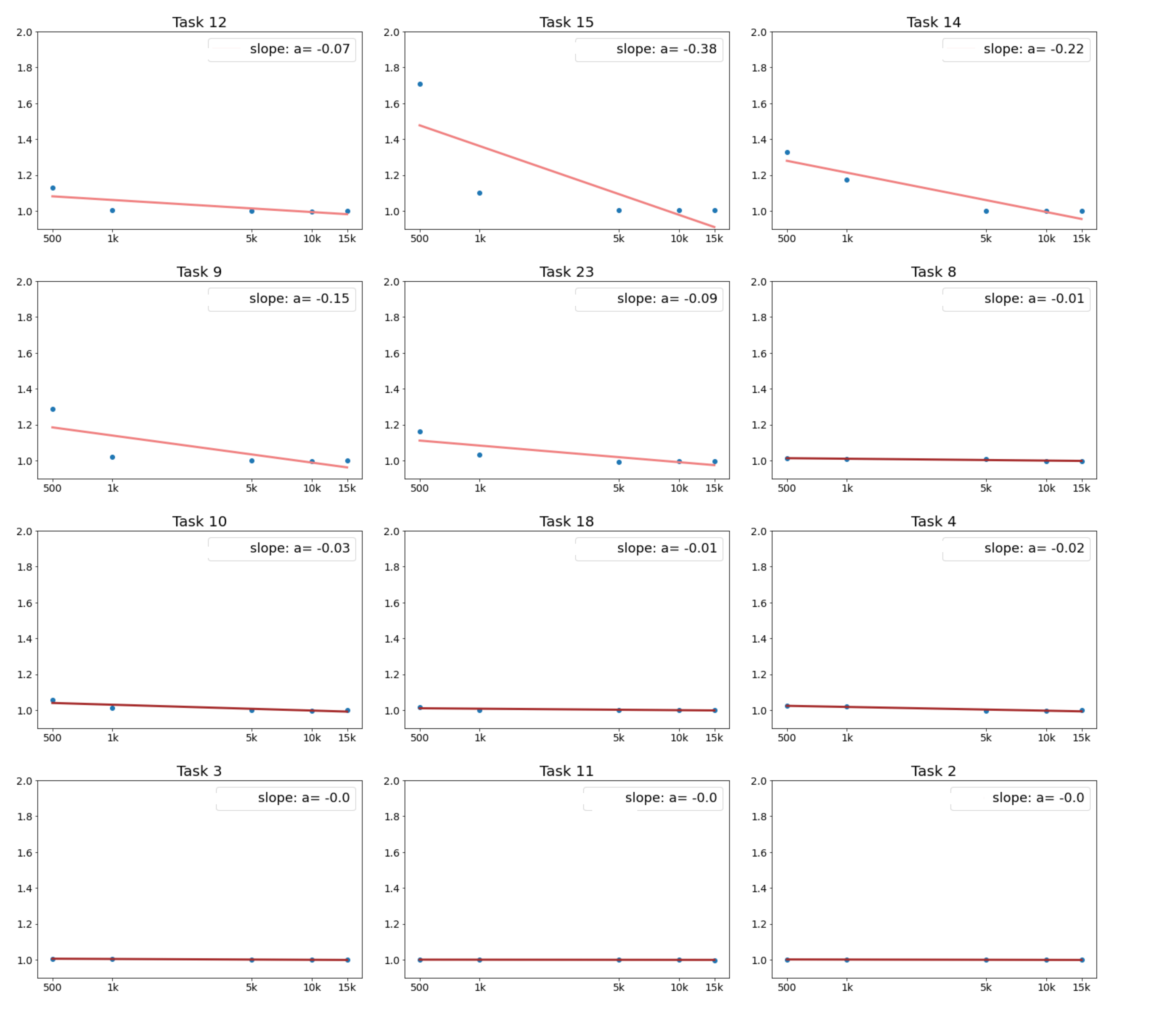}
 \caption{Slope attained by linear fitting of points obtained after taking the ratio of each of the network with spatial attention module and the test accuracy of a ResNet50 for each task and training condition for Spatial Relation (SR) tasks}\label{fig:slope_sa_sr}
\end{figure*}

\begin{figure*}[t]
\centering
  \includegraphics[width=1\linewidth]{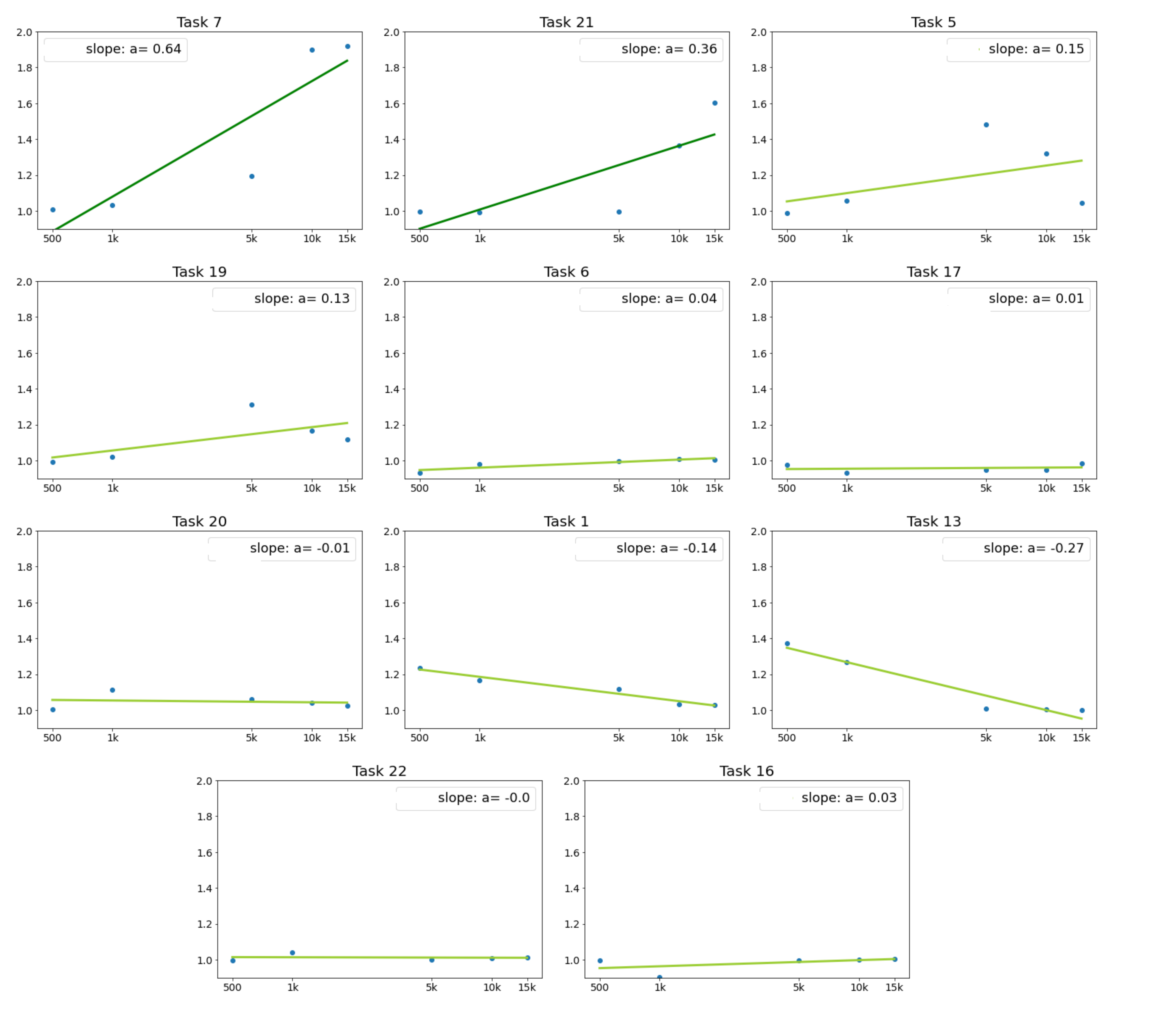}
 \caption{Slope attained by linear fitting of points obtained after taking the ratio of each of the network with feature-based attention module and the test accuracy of a ResNet50 for each task and training condition for Same Different (SD) tasks}\label{fig:slope_fba_sd}
\end{figure*}

\begin{figure*}[t]
\centering
  \includegraphics[width=1\linewidth]{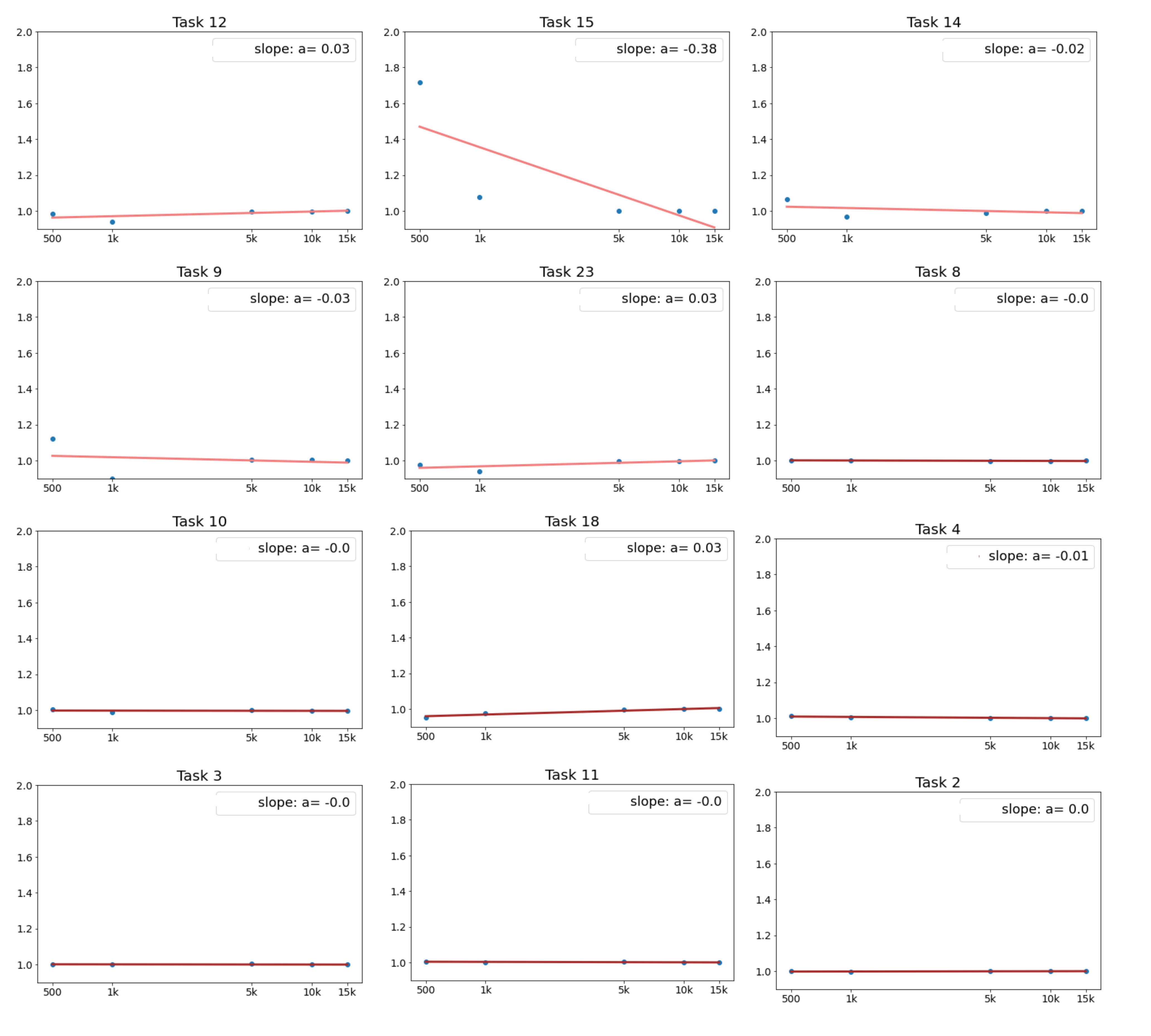}
 \caption{Slope attained by linear fitting of points obtained after taking the ratio of each of the network with feature-based attention module and the test accuracy of a ResNet50 for each task and training condition for Spatial Relation (SR) tasks }\label{fig:slope_fba_sr}
\end{figure*}

\begin{figure*}[htbp]
\centering
\includegraphics[width=1\linewidth]{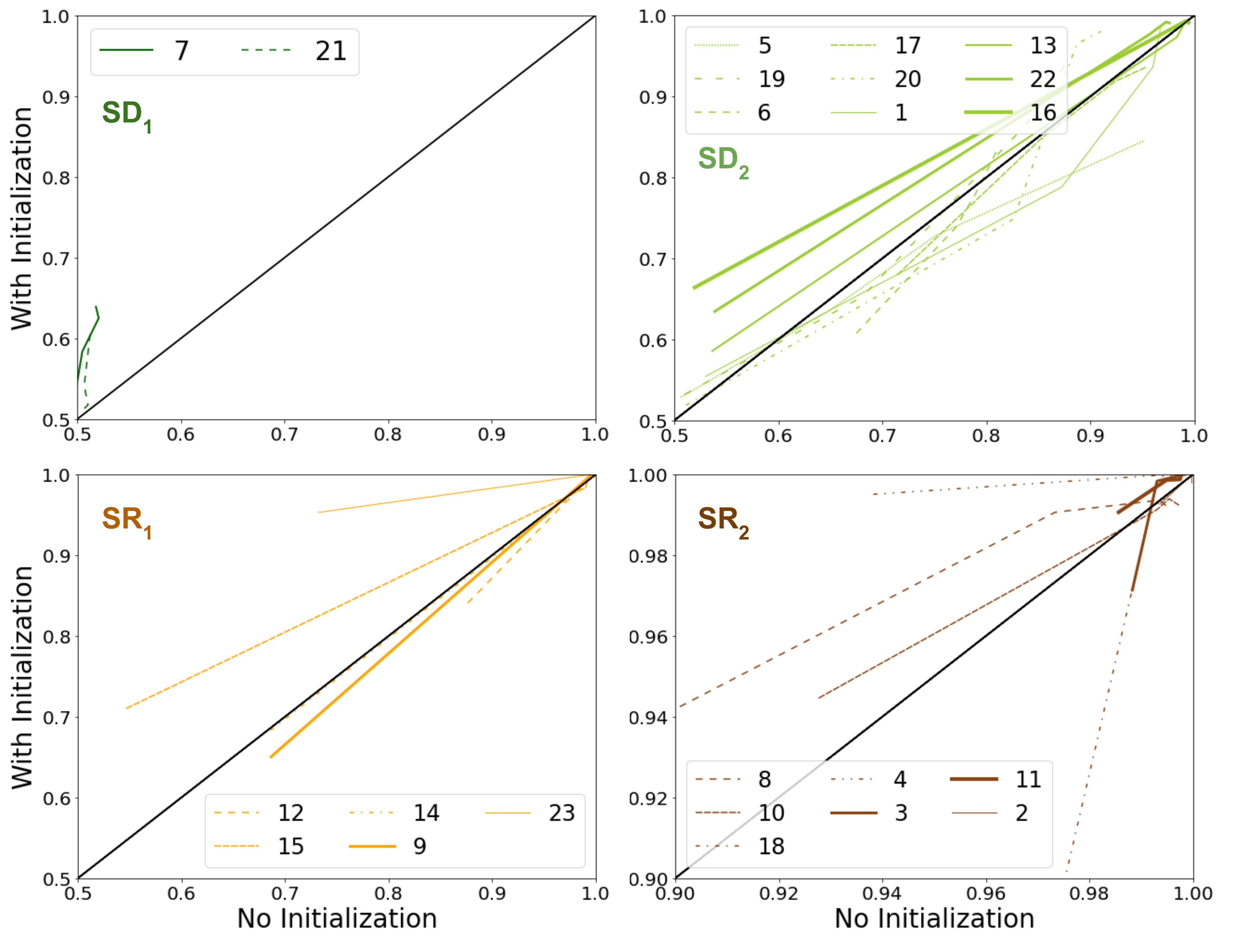} 
  
\caption{Test accuracies for a baseline ResNet50 trained from scratch (``No initialization'') vs. the same architecture pre-trained on Imagenet data for different number of training examples. Also note that a different axis scale is used for $SR_2$ to improve visibility.}\label{fig:xy_img}
\end{figure*}

\part*{References}
\addcontentsline{toc}{part}{References}
\fancyhead[LE]{\textit{REFERENCES}}
\renewcommand{\bibname}{References}
\bibliography{bib/main}

\end{document}